\documentclass{article} 
\usepackage{iclr_conference,times}

\usepackage{amsmath,amsfonts,bm}

\def\eqref#1{equation~\ref{#1}}

\def\1{\bm{1}}

\DeclareMathAlphabet{\mathsfit}{\encodingdefault}{\sfdefault}{m}{sl}
\SetMathAlphabet{\mathsfit}{bold}{\encodingdefault}{\sfdefault}{bx}{n}

\usepackage{hyperref}
\usepackage{url}
\usepackage{amsthm}
\usepackage{amssymb}
\usepackage{algorithm}
\usepackage{algpseudocode}
\usepackage{graphicx}
\usepackage{enumitem}
\usepackage{subcaption}

\usepackage{booktabs}
\usepackage[section]{placeins}

\newtheorem{theorem}{Theorem}

\newtheorem{proposition}{Proposition}
\newtheorem{definition}{Definition}
\newtheorem{lemma}{Lemma}
\newtheorem{assumption}{Assumption}
\newtheorem{corollary}{Corollary}

\newcommand{\D}{{\rm d}}

\title{Modeling AdaGrad, RMSProp, and Adam \\ with Integro-Differential Equations}

\author{
Carlos Heredia\thanks{e-mail address: carlosherediapimienta@gmail.com} \\
\textit{IAMM Research, Department of Applied Artificial Intelligence} \\
\textit{DAMM, Carrer del Rossell\'o 515, 08025 Barcelona, Catalonia, Spain}
}

\iclrfinalcopy
\begin{document}

\maketitle

\begin{abstract}
In this paper, we propose a continuous-time formulation for the AdaGrad, RMSProp, and Adam optimization algorithms by modeling them as first-order integro-differential equations. We perform numerical simulations of these equations, along with stability and convergence analyses, to demonstrate their validity as accurate approximations of the original algorithms. Our results indicate a strong agreement between the behavior of the continuous-time models and the discrete implementations, thus providing a new perspective on the theoretical understanding of adaptive optimization methods.
\end{abstract}


\section{Introduction}

Central to numerous machine learning tasks is the challenge of solving the following optimization problem:
\[
\arg \min_{\theta \in \mathbb{R}^n} f(\theta),
\]
where \( f: \mathbb{R}^n \rightarrow \mathbb{R} \) denotes a typically non-convex and differentiable objective (or loss) function. The pursuit of finding the global minima of such functions presents a significant challenge due to the inherent complexity and non-convexity of the landscape.

Gradient Descent (GD) remains one of the most prominent algorithms for minimizing the function \( f \) by iteratively finding the optimal parameters \( \theta \) \cite{boyd2004}. It operates by adjusting the parameters in the direction of the steepest descent of \( f \) with a fixed step size \( \alpha \) (learning rate). At each iteration, the algorithm computes the gradient of \( f \) with respect to \( \theta \), guiding the parameter updates to minimize \( f \) progressively \cite{Rumelhart1986}:
\begin{equation}\label{intro:GD}
    \theta_{k} \leftarrow \theta_{k-1}  - \alpha \nabla_\theta f_{k}(\theta_{k-1}).
\end{equation}
While Stochastic Gradient Descent (SGD) \cite{Bottou2010} extends the Gradient Descent method by using mini-batches, or randomly selected subsets of data, to compute gradients, this article focuses on a non-stochastic perspective. The continuous nature of these methods permits a more direct application of differential equation techniques, as we demonstrate in Section~\ref{Section:Maths}. For readers interested in a continuous description of the stochastic method, we refer to \cite{sirignano2017stochastic}.

Adaptive optimization methods such as AdaGrad \cite{AdaGra} and RMSProp \cite{hinton_lecture6} have been pivotal in advancing gradient-based algorithms. AdaGrad adapts the learning rate for each parameter by accumulating the sum of past squared gradients, which can lead to a significant decrease in the learning rate over time. In contrast, RMSProp mitigates this issue by maintaining an exponentially weighted average of squared gradients, ensuring a more stable learning process. These innovations laid the groundwork for more sophisticated algorithms like Adam.

Adam (Adaptive Moment Estimation) \cite{kingma2014} refines gradient descent by computing adaptive learning rates for each parameter, leveraging both the first moment and the second moment of the gradients. Variants like AdamW \cite{loshchilov2017} and AdamL2 have been proposed to address specific issues such as weight decay and improved regularization.

To analyze these optimization algorithms through a continuous framework, we employ well-established tools from differential equations to examine the convergence and stability of these methods. For example, under the scaling \( t = \alpha\,k \), the gradient descent algorithm, interpreted via the Euler method, can be approximated by the following first-order differential equation:
\[
\dot{\theta} = -\nabla_\theta f(\theta).
\]
Given that the AdaGrad, RMSProp, and Adam algorithms incorporate memory effects through the accumulation of past gradients, they are naturally expected to be represented by integro-differential operators. The primary contributions of this work are summarized as follows:
\begin{itemize}
    \item We propose continuous formulations for the AdaGrad, RMSProp, and Adam algorithms using integro-differential equations, as described in Propositions \ref{prop:AdaGrad} to \ref{prop:Adam}. We emphasize that the memory effects are encapsulated within the nonlocal terms of these equations.
    
    \item We establish stability and convergence results for the proposed continuous-time models under explicit structural assumptions. Theorems~\ref{thm:adagrad_rmsprop_memory} to \ref{thm:nonconvex_adam_shifted}.
    
    \item We provide a numerical comparison between the results obtained from these continuous models and the original discrete algorithms, focusing on the accuracy and dynamics of the continuous approximations: Figures \ref{fig:AdaGrad_Nonlocal_theta} to \ref{fig:adagrad-nonlocal-ncx_0}. Additionally, we detail the numerical methods employed to solve the integro-differential equations.
\end{itemize}

This article is organized as follows:
\begin{enumerate}[label=\textbf{Sec. \arabic*.}, leftmargin=*, align=left, labelsep=0.5em]
    \setcounter{enumi}{1} 
    \item \textbf{Optimization Algorithms.} We review the AdaGrad, RMSProp, and Adam algorithms.
    \item \textbf{Continuous Representations.} We introduce the continuous versions of these algorithms as first-order integro-differential equations.
    \item \textbf{Convergence and Stability Analysis.} We establish stability and convergence theorems for these continuous nonlocal models
    \item \textbf{Numerical Simulations.} We compare the numerical results obtained from these continuous models with those from the discrete algorithms, and we provide a detailed explanation of the numerical method used to solve the integro-differential equations.
\end{enumerate}

\subsection{Notation}
In this subsection, we introduce and define the notation used throughout Section~~\ref{sec:OPC} and~\ref{sec:Continuous}. Although the notation may appear ``highly" physical, it is adopted here for its simplicity and ease when presenting the results.

Our discussion is set within an \( n \)-dimensional Euclidean space, and the key elements of the notation are as follows:
\begin{itemize}
    \item \textbf{Parameter Space:} The parameters \( \theta \) are represented as vectors in \( \mathbb{R}^n \), where \( \mathbb{R} \) denotes the set of real numbers. These parameters are assumed to be differentiable functions.
    
    \item \textbf{Euclidean Metric:} The Kronecker delta \( \delta_{ij} \) denotes the components of the standard Euclidean metric in \( \mathbb{R}^n \). Indices are raised and lowered using this metric.
    
    \item \textbf{Einstein Summation Convention:} We adopt the Einstein summation convention, wherein repeated indices imply summation over the corresponding index range. For example:
    \[
    \theta^2 := \|\theta\|^2 := \theta_i \theta^i = \delta_{ij} \theta^i \theta^j = \sum_{i=1}^n \theta^i \theta^i,
    \]
    which represents the squared Euclidean norm of the vector \( \theta \).

    \item \textbf{Derivative Notation:} The partial derivative with respect to the \( i \)-th component of \( \theta \) is denoted by \( \partial_i \). Specifically, the gradient operator is expressed as:
    \[
    \nabla_\theta =  \left(\nabla_1, \ldots, \nabla_i, \ldots, \nabla_n\right) = \left(\frac{\partial}{\partial \theta^1}, \ldots, \frac{\partial}{\partial \theta^i}, \ldots, \frac{\partial}{\partial \theta^n}\right) :=  \left(\partial_1, \ldots, \partial_i, \ldots, \partial_n\right).
    \]
\end{itemize}

In Section~\ref{Section:Maths}, we use the traditional notation used in optimization textbooks, where $\langle \cdot,\cdot \rangle$ denotes the scalar product in the $n$-dimensional Euclidean space, as we wish to adhere as closely as possible to standard mathematical notation.

\subsection{Related Work}

Adaptive optimization algorithms like AdaGrad, RMSProp, and Adam are widely adopted in deep learning due to their robustness and efficiency. However, the inherently discrete nature of deep learning algorithms has led researchers to explore continuous representations to gain deeper theoretical insights and leverage mathematical tools from continuous dynamical systems.

\paragraph{Continuous Representations:}
Researchers have extended discrete optimization algorithms to continuous settings using differential equations. For instance, \cite{Su2016} provide a theoretical framework that connects Nesterov's accelerated gradient method with a second-order ordinary differential equation (ODE), offering insights into the convergence properties and dynamics of optimization algorithms. Similarly, \cite{Wibisono2016} present a variational perspective on accelerated methods by deriving ODEs using a Lagrangian framework, capturing the essence of momentum in optimization.

These continuous formulations allow for the analysis of optimization dynamics and convergence properties, such as asymptotic Lyapunov stability \cite{gantmakher1970lectures}. Stability analysis helps determine whether the solutions of optimization algorithms remain near an equilibrium point or diverge over time, providing deeper insights into their robustness \cite{Khalil2002}. Additionally,  \cite{Romero2022} focus on the challenge of discretizing continuous-time analogues of optimization algorithms, particularly gradient-based methods, while retaining the optimality properties inherent in the continuous formulation.

Extending this continuous perspective to neural networks,  \cite{chen2018neural} introduce Neural Ordinary Differential Equations (Neural ODEs), which model the evolution of hidden states in continuous time, bridging the gap between discrete neural networks and continuous dynamical systems. Similarly, \cite{ruthotto2020deep} explore how partial differential equations (PDEs) can be used to model and design deep neural networks. By interpreting layers in a neural network as time steps in a PDE solver, they provide a continuous-time perspective that can lead to new architectures and training algorithms, such as parabolic, elliptic, and hyperbolic CNNs.

On another front, \cite{li2017stochastic} explore stochastic gradient algorithms through stochastic differential equations (SDEs), providing a deeper understanding of their behavior in the continuous limit. This perspective is valuable for analyzing the convergence and stability properties of stochastic optimization methods.

\paragraph{Integro-Differential Equations and Memory Effects:}
Recent research has examined the use of integro-differential equations (IDEs) to model biological and physical processes, as they can more accurately capture memory and cumulative effects than ODEs. For instance, \cite{zappala2022neural} introduce Neural Integro-Differential Equations, combining neural networks with IDEs to model systems where both past and future states influence the current state. This approach is particularly effective in extrapolating temporal data and generalizing to unseen initial conditions, providing a framework for modeling complex dynamics with nonlocal interactions.

\paragraph{Physical Perspectives on Optimization:}
Weinan \cite{weinan2017proposal} suggests rethinking machine learning algorithms through the lens of dynamical systems and control theory. In this view, optimization algorithms are modeled as continuous-time dynamical systems, treating learning as a control problem where parameters are adjusted for optimal outcomes. He also proposes that this continuous framework makes it possible to apply advanced control techniques, such as adaptive time-stepping, Hamiltonian structures, and constraints, to improve the efficiency and stability of machine learning models.

In this context, departing from traditional neural networks that focus on learning mappings between inputs and outputs, Hamiltonian Neural Networks (HNNs) \cite{greydanus2019hamiltonian} and Lagrangian Neural Networks (LNNs) \cite{cranmer2020lagrangian} incorporate fundamental physical principles directly into their architectures. HNNs are rooted in Hamiltonian mechanics and are trained to learn a Hamiltonian function that inherently respects exact conservation laws in an unsupervised manner. This approach makes them particularly effective for modeling systems where energy conservation is critical, such as the two-body problem. Conversely, LNNs employ the Euler-Lagrange equations by parameterizing arbitrary Lagrangian functions using neural networks without requiring canonical coordinates. This capability enables LNNs to perform well in situations where canonical momenta are unknown or difficult to compute. 

Noether's theorem, which establishes a fundamental connection between symmetries and conserved quantities in physical systems, has also inspired advancements in machine learning. For example, Noether Networks \cite{alet2021noether} leverage this theorem to meta-learn conserved quantities, improving prediction quality in sequential problems. Furthermore, \cite{tanaka2021noether} develop a theoretical framework to study the geometry of learning dynamics in neural networks, revealing a key mechanism of explicit symmetry breaking behind the efficiency and stability of modern neural networks.

\vspace{0.4cm}

While previous works have successfully employed ordinary differential equations (ODEs) and partial differential equations (PDEs) to model optimization algorithms and neural networks, they often focus on methods without explicit memory terms or cumulative effects. Our work distinguishes itself by proposing continuous formulations of AdaGrad, RMSProp, and Adam using integro-differential equations. By explicitly modeling the memory effects inherent in these adaptive algorithms through integral operators, we provide a novel perspective that captures their nonlocal dynamics. This distinction is crucial for achieving a deeper theoretical understanding and for developing new optimization strategies inspired by nonlocal behaviors, setting our work apart from prior studies that model optimization algorithms solely as ODEs.

Therefore, our approach allows us to: (a) accurately represent cumulative effects by directly modeling how past gradients influence current updates—a defining characteristic of adaptive optimization algorithms; (b) facilitate theoretical analysis, as the continuous framework enables the application of advanced mathematical tools from the theory of integro-differential equations, such as stability analysis and convergence proofs; and (c) bridge discrete and continuous dynamics, offering insights into the behavior of these algorithms and potential improvements.

\section{Optimization Algorithms}\label{sec:OPC}
In this section, we present the optimization algorithms\footnote{For a comprehensive overview of Gradient Descent optimization algorithms, see \cite{ruder2016overview}.}: AdaGrad, RMSProp, and Adam, with the aim of illustrating the evolution of the parameters as they converge towards the minimum of two representative toy models. The first model is a convex and differentiable function,  
\[
f(\theta) = (\theta - 4)^2,
\]
chosen for its simplicity and for having a known minimum at \( \theta = 4 \), which allows for straightforward analytical determination and facilitates direct comparisons with numerical simulations.  

To complement this, we also consider a non-convex toy model,  
\[
f(\theta) = \frac{1}{4}(\theta^2 - 1)^2,
\]
which exhibits multiple local minima, $\theta=\pm 1$, and an unstable critical point (a local maxima) at $\theta=0$. With these two toy models, we can analyze the dynamics of the continuous-time formulations in both convex and non-convex regimes.

\subsection{Fundamentals of AdaGrad}
The Adaptive Gradient Algorithm (AdaGrad), introduced by \cite{AdaGra}, marks a significant advancement in the development of optimization techniques. The primary innovation of AdaGrad is its ability to adapt the learning rate for each parameter individually, based on the historical gradients observed during training. 

AdaGrad is distinguished by maintaining a cumulative sum of the squares of past gradients for each parameter, denoted as \( G_{k,i} \). This accumulation effectively scales the learning rate inversely proportional to the square root of \( \text{diag}(G_{k,i} )\), ensuring that parameters frequently updated receive smaller updates, while parameters updated less frequently receive larger updates. This particular variant of AdaGrad is known as AdaGrad-Diagonal. However, for simplicity of analysis, in this work we focus on the AdaGrad-Norm variant, in which all parameters are scaled uniformly, namely, $G_k$. 

A notable strength of AdaGrad is its capability to handle scenarios involving sparse features, as it allows for larger updates for parameters associated with infrequent features. However, a limitation arises when the accumulated gradient sum \( G_k \) increases excessively over time, causing the effective learning rate to diminish and potentially leading to premature convergence or slower learning. As will be discussed, this issue is mitigated by the RMSProp algorithm.

\paragraph{Mathematical Formulation of AdaGrad:} Consider an objective function \( f: \mathbb{R}^n\to \mathbb{R} \), parameterized by \( \theta \in \mathbb{R}^n \), with a learning rate \( \alpha\in(0,1] \). AdaGrad iteratively updates the parameters as described in Algorithm \ref{alg:AdaGrad}, where \( \epsilon \) is a small constant added to prevent division by zero (typically \( \sim 10^{-8} \)), and \( k \) denotes the current timestep.

\begin{algorithm}
\caption{AdaGrad Optimization Algorithm}
\label{alg:AdaGrad}
\begin{algorithmic}[1] 
    \State Initialize parameters $\theta^i_{k=0}$ and the accumulated squared gradient sum $G_{k=0} = 0$.
    \While{the parameters $\theta^i_k$ have not converged}
        \State $k \leftarrow k + 1$
        \State Compute the gradient: $g^i_{k} \leftarrow \delta^{ij} \partial_j f(\theta_{k-1})$.
        \State Accumulate gradients: $G_{k} \leftarrow G_{k-1} + g^2_k$
        \State Update parameters: $\theta^i_{k} \leftarrow \theta^i_{k-1} - \alpha\, g^i_{k}/(\sqrt{G_{k}} + \epsilon)$
    \EndWhile
\end{algorithmic}
\end{algorithm}

\subsection{Fundamentals of RMSProp}

The Root Mean Square Propagation (RMSProp) optimizer, introduced by Hinton in his lecture series \cite{hinton_lecture6}, was developed primarily to address the issue of diminishing learning rates encountered in AdaGrad, which can slow down convergence over time.

RMSProp distinguishes itself by maintaining a moving average of the squared gradients for each parameter, denoted as \( v \). This method allows the algorithm to adaptively adjust the learning rate based on the recent magnitude of the gradients, effectively preventing the learning rate from becoming excessively small. The moving average is computed using an exponential decay rate \( \beta \), which determines the influence of recent gradients on the current value of \( v \). A \(\beta\) value close to 1 reduces the impact of recent gradients, resulting in a smoother and more stable average.

A key advantage of RMSProp is its ability to maintain a more consistent learning rate throughout the optimization process. By avoiding the drastic reduction in learning rates seen in AdaGrad, RMSProp allows for more sustained and effective learning over time.

\paragraph{Mathematical Formulation of RMSProp:} Consider an objective function \( f: \mathbb{R}^n \to \mathbb{R} \), parameterized by \( \theta \in \mathbb{R}^n \), and an exponential decay rate \( \beta \in [0,1) \). RMSProp iteratively updates the parameters as described in Algorithm \ref{alg:RMSProp}, where \( \alpha \in (0,1] \) denotes the learning rate, \( \epsilon \) is a small constant to prevent division by zero (typically \( \sim 10^{-8} \)), and \( k \) is the current timestep.

\begin{algorithm}
\caption{RMSProp Optimization Algorithm}
\label{alg:RMSProp}
\begin{algorithmic}[1] 
    \State Initialize parameters $\theta^i_{k=0}$ and the squared gradient moving average $v_{k=0} = 0$.
    \While{the parameters $\theta^i_k$ have not converged}
        \State $k \leftarrow k + 1$
        \State Compute the gradient: $g^i_{k} \leftarrow \delta^{ij} \partial_j f(\theta_{k-1})$.
        \State Update the squared gradient moving average: $v_{k} \leftarrow \beta v_{k-1} + (1 - \beta) g_{k}^2$
        \State Update parameters: $\theta^i_{k} \leftarrow \theta^i_{k-1} - \alpha\, g^i_{k}/(\sqrt{v_{k}} + \epsilon)$
    \EndWhile
\end{algorithmic}
\end{algorithm}

\subsection{Fundamentals of Adam}

The Adaptive Moment Estimation (Adam) optimizer, introduced by \cite{kingma2014}, represents also a significant advancement in optimization techniques, particularly in the fields of deep learning. It combines the strengths of two other gradient-based optimization methods: the Adaptive Gradient Algorithm and Root Mean Square Propagation. 

Adam distinguishes itself by maintaining two separate moving averages for each parameter: the first moment, \( m^i \), and the second moment, \( v \). The first moment estimates the gradients, while the second moment estimates the squared gradients, similar to RMSProp's adaptive learning rates. These moving averages are computed using exponential decay rates, \( \beta_1 \) and \( \beta_2 \), respectively.

Values of \( \beta_1 \) and \( \beta_2 \) close to 1 imply a longer memory of past gradients, effectively smoothing the averages over a larger number of iterations. This smoothing helps stabilize the optimization process by reducing the impact of sudden changes in gradient values.

\paragraph{Mathematical Formulation of Adam:} Consider an objective function \( f: \mathbb{R}^n \to \mathbb{R} \), parameterized by \( \theta \in \mathbb{R}^n \), with exponential decay rates \( \beta_1, \beta_2 \in [0,1) \). Adam iteratively updates the parameters as described in Algorithm \ref{alg:Adam}, where \( \alpha \in (0,1] \) denotes the learning rate, \( \epsilon \) is a small constant to prevent division by zero (typically \( \sim 10^{-8} \)), and \( k \) is the current timestep.

\begin{algorithm}
\caption{Adam Optimization Algorithm}
\label{alg:Adam}
\begin{algorithmic}[1] 
    \State Initialize the parameters $\theta^i_{k=0}$, the first moment $m^i_{k=0} = 0$, and the second moment $v_{k=0} = 0$.
    \While{the parameters $\theta^i_k$ have not converged}
        \State $k \leftarrow k + 1$
        \State Compute the gradient: $g^i_{k} \leftarrow \delta^{ij} \partial_j f(\theta_{k-1})$.
        \State Update first moment: $m^i_{k} \leftarrow \beta_1 m^i_{k-1} + (1 - \beta_1) g^i_{k}$
        \State Update second moment: $v_{k} \leftarrow \beta_2 v_{k-1} + (1 - \beta_2) g_{k}^2$
        \State Correct first moment: $\hat{m}^i_{k} \leftarrow m^i_{k}/(1 - \beta_1^k)$
        \State Correct second moment: $\hat{v}_{k} \leftarrow v_{k}/(1 - \beta_2^k)$
        \State Update parameters: $\theta^i_{k} \leftarrow \theta^i_{k-1} - \alpha\, \hat{m}^i_{k}/(\sqrt{\hat{v}_{k}} + \epsilon)$
    \EndWhile
\end{algorithmic}
\end{algorithm}

\section{Continuous-time models for AdaGrad, RMSProp and Adam}\label{sec:Continuous}
Expressing the optimizers discussed in the previous section within a continuous framework offers several advantages, particularly by enhancing the understanding and analysis of their temporal behavior. Continuous formulations allow the application of advanced mathematical tools, such as Lyapunov stability analysis, to investigate the dynamics of parameter updates and the convergence properties of the algorithm \cite{kovachki2021continuous}.

Next, we present three propositions (three continuous-time models) for the optimization algorithms AdaGrad, RMSProp, and Adam. In all cases, we assume the following conditions:

\begin{assumption}\label{Ass:Assumption_t}
The relationship between the temporal variable \( t \) and the iteration \( k \) is governed by the learning rate \( \alpha \) such that
\begin{equation}
 t = \alpha\,k\,.
\end{equation}
\end{assumption}

Building on this assumption,
\begin{assumption}\label{Ass:Assumption}
Given a small learning rate \( \alpha \), the timescale relationship is expressed as:
\[
z^i_{\frac{t+\alpha}{\alpha}} := z^i(t+\alpha) \sim z^i(t) + \alpha\,\dot z^i(t) + \mathcal{O}(\alpha^2) \quad \text{and} \quad z^i_{\frac{t}{\alpha}} := z^i(t) \quad \text{for} \quad  t \geq 0\,.
\]
\end{assumption}

In what follows, we work at first order in \(\alpha\), neglecting \(O(\alpha^2)\) remainder terms for simplicity. A second-order extension of this framework is discussed in~\cite{heredia2026adamadamlikelagrangianssecondorder}. Hence, 
\begin{proposition}[\textbf{Nonlocal Continuous Dynamics of AdaGrad}]\label{prop:AdaGrad}
Under Assumptions \ref{Ass:Assumption_t} and \ref{Ass:Assumption}, and with an initial value for the accumulated gradients \( G_0 = 0 \), the continuous nonlocal dynamics of AdaGrad can be characterized by the following equation:
\begin{equation}\label{eq:AdaGrad_cont}
     \dot{\theta}^i(t) = - \frac{\partial^i f(\theta)}{\sqrt{G(t+\alpha, \theta)} + \epsilon}\,  \quad \text{with} \quad t \geq 0,
\end{equation}
where \( \epsilon \) is a small real value (typically \( \sim 10^{-8} \)), and the nonlocal term \( G(t, \theta) \) is defined as:
\begin{equation*}
    G(t, \theta) = \frac{1}{\alpha} \int^{t}_0 \mathrm{d}\tau\, \partial^i f(\theta(\tau)) \partial_i f(\theta(\tau))\,.
\end{equation*}
\end{proposition}
\begin{proof}
To facilitate the proof, we begin by shifting the temporal indexing from \( k \leftarrow k+1 \) to \( k+1 \leftarrow k \). Consequently, the parameter update rule becomes:
\[
\theta^i_{k+1} = \theta^i_{k} - \alpha\, \frac{\partial^i f(\theta_k)}{\sqrt{G_{k+1}} + \epsilon}\,.
\]
By applying Assumption \ref{Ass:Assumption}, the evolution of the accumulated gradients can be described by the differential equation:
\[
\dot{G}(t,\theta) = \frac{1}{\alpha} \partial^i f(\theta) \partial_i f(\theta)\,,
\]
whose solution is:
\[
G(t,\theta) = \frac{1}{\alpha} \int^t_0 \, \mathrm{d}\tau \, \partial^i f(\theta(\tau)) \partial_i f(\theta(\tau))\, \quad \text{with} \quad t\geq 0.
\]
Using Assumption \ref{Ass:Assumption} again, we derive the integro-differential equation that governs the updated parameters:
\[
\dot{\theta}^i(t) = - \frac{\partial^i f(\theta)}{\sqrt{G(t+\alpha, \theta)} + \epsilon}\,.
\]
\end{proof}
This proposition highlights that the continuous-time model for AdaGrad is described by a first-order integro-differential equation. Notice that the nonlocality is embedded in the accumulated gradient term, where the entire trajectory of \( \theta \) up to time \( t \) must be known, along with a small differential time step (future) described by \( \alpha \). The appearance of \(t+\alpha\) is a direct consequence of the reindexed discrete update: the parameter \(\theta_{k+1}\) is updated using the accumulated quantity \(G_{k+1}\). Thus, the shifted argument \(t+\alpha\) is the continuous counterpart of the one-step-ahead memory term in the
discrete AdaGrad recursion.

Now, we introduce the continuous model for RMSProp. Hence,
\begin{proposition}[\textbf{Nonlocal Continuous Dynamics of RMSProp}]\label{prop:RMSProp}
Under Assumptions \ref{Ass:Assumption_t} and \ref{Ass:Assumption}, and given an initial value for the squared gradient moving average \( v_0 = 0 \), the continuous nonlocal dynamics of RMSProp can be characterized by the following equation:
\begin{equation}\label{eq:RMSProp_cont}
     \dot{\theta}^i(t) = - \frac{\partial^i f(\theta)}{\sqrt{v(t + \alpha, \theta)} + \epsilon}\,  \quad \text{with} \quad t \geq 0,
\end{equation}
where \( \epsilon \) is a small real value (typically \( \sim 10^{-8} \)), and the nonlocal term \( v(t, \theta) \) is defined as:
\begin{equation*}
    v(t,\theta) = \frac{1-\beta}{\alpha} \int^t_0 \mathrm{d}\tau\,e^{-\frac{1-\beta}{\alpha} (t-\tau)} \partial^i f(\theta(\tau)) \partial_i f(\theta(\tau))\,.
\end{equation*}
\end{proposition}

\begin{proof}
To simplify the proof, as in Proposition \ref{prop:AdaGrad}, we shift the indexing such that \(k+1 \leftarrow k\). The parameter update then becomes:
\[
\theta^i_{k+1} = \theta^i_{k} - \alpha\, \frac{\partial^i f(\theta_k)}{\sqrt{v_{k+1}} + \epsilon}\,.
\]
By applying Assumption \ref{Ass:Assumption}, the evolution of the squared gradient moving average is governed by the differential equation:
\begin{equation}
\dot{v} + \frac{1-\beta}{\alpha} v = \frac{1-\beta}{\alpha} \partial^i f(\theta) \partial_i f(\theta)\,.
\end{equation}
Rewriting this equation yields:
\begin{equation}
\frac{\D}{\D t}\left[e^{\frac{1-\beta}{\alpha} t} v \right] = \frac{1-\beta}{\alpha} e^{\frac{1-\beta}{\alpha}t} \partial^i f(\theta) \partial_i f(\theta)\,,
\end{equation}
which leads to the solution:
\begin{equation}
v(t,\theta) = \frac{1-\beta}{\alpha} \int^t_0 \D \tau \, e^{-\frac{1-\beta}{\alpha}(t-\tau)} \partial^i f(\theta(\tau)) \partial_i f(\theta(\tau))\, \quad \text{with} \quad t \geq 0\,.
\end{equation}
Finally, by applying Assumption \ref{Ass:Assumption}, the parameter updates follow the integro-differential equation:
\begin{equation}
\dot{\theta}^i(t) = - \frac{\partial^i f(\theta)}{\sqrt{v(t + \alpha, \theta)} + \epsilon}\,.
\end{equation}
\end{proof}
As discussed earlier, a similar situation arises with RMSProp: the dynamics are governed by the integro-differential equation (\ref{eq:RMSProp_cont}). The nonlocal behavior is encapsulated within the squared gradient moving average. Thus, the dynamics require knowledge of the trajectory of \(\theta\) up to time \(t\), together with the small forward increment determined by \(\alpha\). As in AdaGrad, the appearance of the shifted argument \(t+\alpha\) is a consequence of the reindexing, since the update of \(\theta_{k+1}\) depends on the memory term evaluated at step \(k+1\).

The key difference between RMSProp and AdaGrad lies in the kernel of the integral operator. For AdaGrad, the kernel is 
\[
K_{AG}(t) = \frac{1}{\alpha}\,.
\] 
However, for RMSProp, the kernel is modified to:
\[
K_{RP}(t) = \frac{1-\beta}{\alpha} e^{-\frac{1-\beta}{\alpha}t}\,.
\]

Finally, we introduce the continuous model for Adam:

\begin{proposition}[\textbf{Nonlocal Continuous Dynamics of Adam}]\label{prop:Adam}
Under Assumption \ref{Ass:Assumption_t} and \ref{Ass:Assumption}, and with initial values for the moments \( m^i_0 = 0 \) and \( v_0 = 0 \), the Adam optimizer can be characterized by the following continuous nonlocal dynamics:
\begin{equation}\label{eq:Adam_cont}
    \dot{\theta}^i(t) = - \eta(t+\alpha)\,T^i(t+\alpha,\theta) \quad \text{with} \quad t \geq 0\,,
\end{equation}
where
\begin{equation}
    \eta(t) = \frac{\sqrt{1-\beta_2^{\frac{t}{\alpha}}}}{1-\beta_1^{\frac{t}{\alpha}}} , \qquad \text{and} \qquad T^i(t,\theta) = \frac{m^i(t,\theta)}{\sqrt{v(t,\theta)}+\varepsilon (t)},
\end{equation}
and the moments are given by:
\begin{equation*}
    m^i(t,\theta) = \int^t_0 \D\tau\, K_1(t-\tau) \,\partial^i f(\theta(\tau)), \qquad \text{and} \qquad v(t,\theta) = \int^t_0 \D\tau\, K_2(t-\tau) \,\partial^i f(\theta(\tau)) \,\partial_i f(\theta(\tau)),
\end{equation*}
with the kernel function \( K_a(t) \) defined as:
\begin{equation*}
     K_a(t) = \frac{1-\beta_a}{\alpha} e^{-\frac{1-\beta_a}{\alpha} t}, \qquad a = 1,2\,,
\end{equation*}
and 
\[
\varepsilon(t) = \epsilon\sqrt{1-\beta_2^\frac{t}{\alpha}}\,.
\]
\end{proposition}

\begin{proof}
To prove this proposition, we start from step 5-6 of Algorithm \ref{alg:Adam}: \(m^i_{k} = \beta_1 m^i_{k-1} + (1 - \beta_1) g^i_{k}\) and \(v_{k} = \beta_2 v_{k-1} + (1 - \beta_2) g^2_{k}\). Since the structure is identical for both moments, we demonstrate the process for one, knowing the other will follow similarly, except for differences in the exponential decay or gradient. We redefine the indices as \(k+1 \leftarrow k\) to avoid involving \(\theta_{t-1}\) in the gradient.

Focusing on the second moment \(v\), we use the same approach as in Proposition \ref{prop:RMSProp} to obtain:
\begin{equation}
v(t,\theta) = \frac{1-\beta_2}{\alpha} \int^t_0 \D \tau \, e^{-\frac{1-\beta_2}{\alpha}(t-\tau)} \partial^i f(\theta(\tau)) \partial_i f(\theta(\tau))\, \quad \text{with} \quad t\geq 0.
\end{equation}
The first moment \(m^i(t,\theta)\) follows a similar pattern, differing by the absence of the squared gradient and the exponential decay factor \(\beta_2\):
\begin{equation}
m^i(t,\theta) = \frac{1-\beta_1}{\alpha} \int^t_0 \D \tau \, e^{-\frac{1-\beta_1}{\alpha}(t-\tau)} \partial^i f(\theta(\tau)) \quad \text{with} \quad t \geq 0.
\end{equation}
Applying Assumptions \ref{Ass:Assumption_t} and \ref{Ass:Assumption} and performing some algebraic manipulations yields:
\begin{equation}\label{eq:Adam1cont}
\dot{\theta}^i(t) = - \eta(t+\alpha)\,T(t+\alpha,\theta),
\end{equation}
where
\begin{equation}
    \eta(t) = \frac{\sqrt{1-\beta_2^{\frac{t}{\alpha}}}}{1-\beta_1^{\frac{t}{\alpha}}} , \qquad \text{and} \qquad T^i(t,\theta) = \frac{m^i(t,\theta)}{\sqrt{v(t,\theta)}+\varepsilon (t)},
\end{equation}
with 
\[
\varepsilon(t) = \epsilon\sqrt{1-\beta_2^\frac{t}{\alpha}}\,.
\]
\end{proof}
In the case of Adam, the nonlocality is encoded in both moments, \(m^i\) and \(v\), as expected. With the indexing convention adopted
above, the Adam dynamics also involve the shifted time \(t+\alpha\). This shift is not an additional modeling assumption; rather, it is the
continuous counterpart of the fact that the discrete update of \(\theta_{k+1}\) depends on the bias-corrected moments evaluated at the same step, namely \(m_{k+1}\) and \(v_{k+1}\). Therefore, as in the AdaGrad and RMSProp cases, the dynamics require the trajectory of \(\theta\) up to the small forward increment determined by \(\alpha\). Although \(T^i(0,\theta)\) is formally undefined, since all the terms in its denominator vanish at \(t=0\), this singularity is never encountered in the dynamics: the argument of \(T^i\) is always \(t+\alpha\).

These propositions confirm that the accumulative effects present in adaptive algorithms such as AdaGrad, RMSProp, and Adam are represented in continuous form as integro-differential equations (or nonlocal equations), where the presence of the kernel in the integral operator regulates the influence of the gradient on the dynamics.

\section{Theoretical Framework for Continuous-Time Dynamics}\label{Section:Maths}

Thanks to the continuous-time formulation of these algorithms, we can investigate their convergence and stability properties using mathematical tools such as Lyapunov stability analysis and convergence theory for both convex and non-convex objective functions. To enhance clarity, this section is divided into two parts: the first focuses on convex functions, while the second addresses non-convex functions. Readers who are less interested in the mathematical details may proceed directly to the next section.

\subsection{General Assumptions}
To begin, we assume that \( f : \mathbb{R}^n \to \mathbb{R} \) is of class \( \mathcal{C}^1 \). We denote by \( \theta(t) \in \mathbb{R}^n \) the continuous-time trajectory of the parameter,  and by \( g(t) := \nabla_\theta f(\theta(t)) \) the gradient evaluated along this trajectory. For notational simplicity, we suppress component indices whenever possible and make them explicit only when necessary. Furthermore, in order to represent compactly the kernels of the integral operators appearing in the integro-differential equations, we introduce the following definition.

\begin{definition}\label{def:Borel}
Let \(\nu\) be a finite positive Borel measure on \([0,\infty)\). We define the kernel function
\begin{equation}
    K_\nu(t) := \int_{[0,\infty)} e^{-\lambda t} \,\mathrm{d}\nu(\lambda), \qquad t \geq 0.
\end{equation}
Given a function \( g: [0,\infty) \to \mathbb{R}^n \) that is measurable and locally integrable, the associated memory operator is defined as
\begin{equation}
    M_{\nu}[g](t) := \int_0^t K_\nu(t - \tau) \, g(\tau) \,\mathrm{d}\tau.
\end{equation}
\end{definition}
\begin{lemma}\label{lemma:kernels}
Let \( \alpha > 0 \) and \( \beta_a \in [0,1) \). If
\begin{equation}
    \nu_{\mathrm{AG}} = \frac{1}{\alpha} \, \delta_0, 
    \qquad 
    \nu_{\mathrm{RM/Adam}} = \frac{1-\beta_a}{\alpha} \, \delta_{\frac{1-\beta_a}{\alpha}},
\end{equation}
then the corresponding kernels defined in Definition~\ref{def:Borel} are
\begin{equation}
    K_{\nu_{\mathrm{AG}}}(t) = \frac{1}{\alpha}, 
    \qquad 
    K_{\nu_{\mathrm{RM/Adam}}}(t) = \frac{1-\beta_a}{\alpha} e^{-\frac{1-\beta_a}{\alpha} t},
\end{equation}
where \( \delta_a \) denotes the Dirac measure, i.e.,
\begin{equation}
    \delta_a(B) = 
    \begin{cases}
        1, & \text{if } a \in B, \\
        0, & \text{otherwise}.
    \end{cases}
\end{equation}
\end{lemma}

\begin{proof}
    Straightforward calculation.
\end{proof}

Next, we introduce the necessary lemmas, propositions, and regularity assumptions that will serve as the basis for proving more specific results---both lemmas and theorems---for convex and non-convex functions.

\begin{lemma}[Bounds on $K_\nu$]\label{lem:puntual}
For all \( t \geq 0 \),
\[
    0 \ \leq\ K_\nu(t) \ \leq\ \|\nu\|,
    \quad\text{and in particular}\quad
    K_\nu(0) = \|\nu\|.
\]
\end{lemma}
\begin{proof}
Since \(\nu\) is a positive measure and \( e^{-\lambda t} \in [0,1] \) for every \( \lambda \geq 0 \) and \( t \geq 0 \), we obtain
\[
    0 \ \leq \ \int_{[0,\infty)} e^{-\lambda t} \,\mathrm{d}\nu(\lambda) 
      \ \leq \ \int_{[0,\infty)} 1 \,\mathrm{d}\nu(\lambda) 
      = \nu([0,\infty)) 
\]
Since $\nu$ is positive, its total variation coincides with its total mass, namely, $\|\nu\| = \nu([0,\infty))$. Therefore
\[
0\leq K_\nu(t) \leq \| \nu\|\,.
\]
The identity \( K_\nu(0) = \|\nu\| \) follows immediately by taking \( t = 0 \) in the definition of \(K_\nu\). For AdaGrad, this yields $\|\nu_\mathrm{AG}\| = 1/\alpha$, and for RMSProp/Adam, $\|\nu_\mathrm{RM/Adam}\| = (1-\beta_a)/\alpha$.
\end{proof}

\begin{assumption}[\(L\)-smoothness]\label{ass:regularidad}
Let \(f:\mathbb{R}^n\to\mathbb{R}\) be of class \(C^1\). There exists \(L>0\) such that
\[
\|\nabla f(x)-\nabla f(y)\|\ \le\ L\,\|x-y\|\qquad \forall\,x,y\in\mathbb{R}^n,
\]
equivalently,
\[
f(y)\ \le\ f(x)+\langle \nabla f(x),\,y-x\rangle+\tfrac{L}{2}\|y-x\|^2\qquad \forall\,x,y\in\mathbb{R}^n.
\]
\end{assumption}

To conclude the general lemmas and propositions, 
\begin{definition}\label{def:M}
Throughout, the notation \(M_\nu[g^2](t)\) is understood as
\[
    M_\nu[g^2](t) \ :=\ \int_0^t K_\nu(t-\tau)\,\|g(\tau)\|^2\,\mathrm{d}\tau.
\]
This convention avoids ambiguities that could arise from a component-wise interpretation.
\end{definition}



With these definitions in place, we are ready to treat the convex and non-convex settings. Before doing so, Table~\ref{tab:placeholder} summarizes the correspondence between the notation used in Section~\ref{sec:Continuous} and the more compact notation introduced in the present section. This notation will be used throughout the analysis, since it allows the stability and convergence results to be stated in a unified form.

\begin{table}[htbp]
    \centering
    \setlength{\tabcolsep}{12pt} 
    \renewcommand{\arraystretch}{1.2} 
    \begin{tabular}{@{}cc@{}}
      \toprule
      \textbf{Section 3} & \textbf{Section 4} \\
      \midrule
      $G(t)$      & $M_{\nu_{\text{AG}}}[g^2](t)$ \\
      $v(t)$      & $M_{\nu_{\text{RM/Adam}}}[g^2](t)$ \\
      $m^{i}(t)$  & $M^{i}_{\nu_{\text{Adam}}}[g](t)$ \\
      \bottomrule
    \end{tabular}
    \caption{Correspondence between the notation used in Section~\ref{sec:Continuous} and Section~\ref{Section:Maths}.}
    \label{tab:placeholder}
  \end{table}

Finally, throughout the following subsections, we assume that the shifted continuous-time systems under consideration admit global classical solutions for the prescribed initial data.

\subsection{Case I: Convex Objective Functions}\label{Section:Convex_math}

In this subsection, we shall concentrate on convex objective functions, under the assumptions of regularity---Assumption~\ref{ass:regularidad}---and convexity \cite{BauschkeCombettes2017}:

\begin{assumption}[Strong convexity]\label{ass:convexidad}
Let \(f:\mathbb{R}^n\to\mathbb{R}\) be of class \(C^1\). There exists \(m>0\) such that
\[
\langle \nabla f(x)-\nabla f(y),\,x-y\rangle\ \ge\ m\,\|x-y\|^2\qquad \forall\,x,y\in\mathbb{R}^n.
\]
equivalently,
\[
f(y)\ \ge\ f(x)+\langle \nabla f(x),\,y-x\rangle+\tfrac{m}{2}\|y-x\|^2\qquad \forall\,x,y\in\mathbb{R}^n,
\]
\end{assumption}

\begin{lemma}\label{lem:grad_bounds}
Let \( f : \mathbb{R}^n \to \mathbb{R} \) be \(m\)-strongly convex and \(L\)-smooth, with minimizer \(x^*\).  
Then, for all \( x \in \mathbb{R}^n \),
\[
    \frac{1}{2L} \, \|\nabla f(x)\|^2 
    \ \leq\ f(x) - f(x^*) 
    \ \leq\ \frac{1}{2m} \, \|\nabla f(x)\|^2.
\]
\end{lemma}

\begin{proof}
\noindent\textbf{(a) Lower bound.} Since \(f\) is \(L\)-smooth, for every \(x,y\in\mathbb R^n\),
\[
f(y)
\leq
f(x)+\langle \nabla f(x),y-x\rangle
+\frac L2\|y-x\|^2.
\]
Choose
\[
y=x-\frac1L\nabla f(x).
\]
Then
\[
f\left(x-\frac1L\nabla f(x)\right)
\leq
f(x)
-\frac1L\|\nabla f(x)\|^2
+\frac L2\left\|\frac1L\nabla f(x)\right\|^2.
\]
Hence
\[
f\left(x-\frac1L\nabla f(x)\right)
\leq
f(x)-\frac1{2L}\|\nabla f(x)\|^2.
\]
Since \(x^*\) is a global minimizer of \(f\),
\[
f(x^*)\leq f\left(x-\frac1L\nabla f(x)\right).
\]
Therefore,
\[
f(x^*)
\leq
f(x)-\frac1{2L}\|\nabla f(x)\|^2,
\]
which gives
\[
f(x)-f(x^*)
\geq
\frac1{2L}\|\nabla f(x)\|^2.
\]
\noindent\textbf{(b) Upper bound.}  
If \(f\) is \(m\)-strongly convex, then for all \(x,y\),
\[
    f(y) \ \geq\ f(x) + \langle \nabla f(x),\, y - x \rangle + \frac{m}{2} \|y - x\|^2.
\]
Taking \(y = x^*\), we get
\[
    f(x^*) \ \geq\ f(x) + \langle \nabla f(x),\, x^* - x \rangle + \frac{m}{2} \|x - x^*\|^2,
\]
which is equivalent to
\[
    f(x) - f(x^*) \ \leq\ \langle \nabla f(x),\, x - x^* \rangle - \frac{m}{2} \|x - x^*\|^2.
\]
Letting \(g := \nabla f(x)\) and \(d := x - x^*\), we maximize the quadratic form
\[
    \psi(d) := \langle g,\, d \rangle - \frac{m}{2} \|d\|^2
\]
with respect to \(d\). The maximizer is \(d = g/m\), and the maximum value is
\[
    \max_{d} \psi(d) = \frac{1}{2m} \|g\|^2.
\]
Therefore,
\[
    f(x) - f(x^*) \ \leq\ \frac{1}{2m} \|\nabla f(x)\|^2.
\]
\end{proof}

As the dynamics of AdaGrad and RMSProp have the same structural form, differing only in the choice of the memory kernel, we first state a unified preliminary estimate for both methods. The purpose of this result is to provide an a priori convergence bound that is robust enough to control the accumulated memory term.  

\begin{theorem}[Preliminary convergence bound for AdaGrad/RMSProp]
\label{thm:adagrad_rmsprop_memory}
Let \(f:\mathbb R^n\to\mathbb R\) be \(L\)-smooth and \(m\)-strongly convex, with unique minimizer \(\theta^*\). Let \(\nu\) be a finite positive Borel measure on \([0,\infty)\), and define \(M_\nu[g^2](t)\) as in Definition~\ref{def:M}. Assume that \(\theta(t)\) is a global classical solution of
\[
\dot\theta(t)
=
-
\frac{\nabla f(\theta(t))}
{\sqrt{M_\nu[g^2](t+\alpha)}+\epsilon},
\qquad t\ge 0,
\]
with \(\epsilon>0\) and \(\alpha>0\). Moreover, set
\[
\Phi_0:=f(\theta(0))-f(\theta^*),
\qquad
G_0:=L\sqrt{\frac{2\Phi_0}{m}},
\qquad
B:=G_0\sqrt{\|\nu\|}.
\]
Then, for every \(t\ge 0\),
\[
\|\theta(t)-\theta^*\|
\le
\|\theta(0)-\theta^*\|
\exp\!\left(-m\,I_\alpha(t)\right),
\]
where
\[
I_\alpha(t)
:=
\int_0^t
\frac{ds}{\epsilon+B\sqrt{s+\alpha}}.
\]
Equivalently, if \(B>0\),
\[
I_\alpha(t)
=
\frac{2}{B}
\left(\sqrt{t+\alpha}-\sqrt{\alpha}\right)
-
\frac{2\epsilon}{B^2}
\log\!\left(
\frac{\epsilon+B\sqrt{t+\alpha}}
{\epsilon+B\sqrt{\alpha}}
\right).
\]
In particular,
\[
I_\alpha(t)\sim \frac{2}{B}\sqrt t,
\qquad t\to\infty,
\]
and therefore
\begin{equation}\label{th1:bound}
\|\theta(t)-\theta^*\|
\le
C\exp(-c\sqrt t)
\end{equation}
for suitable constants \(C,c>0\).
\end{theorem}

\begin{proof}
Let
\[
g(t):=\nabla f(\theta(t)),
\qquad
H_\alpha(t)
:=
\sqrt{M_\nu[g^2](t+\alpha)}+\epsilon.
\]
With this notation, the dynamics of Proposition \ref{prop:AdaGrad} and \ref{prop:RMSProp} can be written as
\[
\dot\theta(t)
=
-
\frac{g(t)}{H_\alpha(t)}.
\]
Since \(M_\nu[g^2](t+\alpha)\ge 0\) and \(\epsilon>0\), we have
\[
H_\alpha(t)>0
\qquad \text{for all } t\ge 0.
\]

We define the objective gap as
\[
\Phi(t)
:=
f(\theta(t))-f(\theta^*).
\]
Differentiating along the flow gives
\[
\dot\Phi(t)
=
\langle \nabla f(\theta(t)),\dot\theta(t)\rangle
=
\left\langle g(t),-\frac{g(t)}{H_\alpha(t)}\right\rangle
=
-
\frac{\|g(t)\|^2}{H_\alpha(t)}
\le 0.
\]
Therefore \(\Phi\) is nonincreasing, and hence
\begin{equation}
\label{eq:Phi_decreasing}
\Phi(t)\le \Phi(0)=\Phi_0,
\qquad t\ge 0.
\end{equation}

Since \(f\) is \(m\)-strongly convex and \(\theta^*\) is its unique minimizer, we have
\[
\Phi(t)
=
f(\theta(t))-f(\theta^*)
\ge
\frac m2\|\theta(t)-\theta^*\|^2.
\]
Combining this estimate with equation~(\ref{eq:Phi_decreasing}), we obtain
\begin{equation}
\label{eq:theta_bound}
\|\theta(t)-\theta^*\|
\le
\sqrt{\frac{2\Phi_0}{m}},
\qquad t\ge 0.
\end{equation}
Moreover, since \(f\) is \(L\)-smooth and \(\nabla f(\theta^*)=0\),
\[
\|g(t)\|
=
\|\nabla f(\theta(t))-\nabla f(\theta^*)\|
\le
L\|\theta(t)-\theta^*\|.
\]
Using equation~(\ref{eq:theta_bound}), we get the uniform bound
\begin{equation}
\label{eq:g_bound}
\|g(t)\|
\le
L\sqrt{\frac{2\Phi_0}{m}}
=
G_0,
\qquad t\ge 0.
\end{equation}

We now estimate the memory term. By Lemma~\ref{lem:puntual}, for every \(r\ge 0\),
\[
0\le K_\nu(r)\le \|\nu\|.
\]
Hence, using equation~(\ref{eq:g_bound}), for every \(t\ge 0\),
\[
M_\nu[g^2](t+\alpha)
=
\int_0^{t+\alpha}
K_\nu(t+\alpha-\tau)\|g(\tau)\|^2\,\mathrm d\tau
\le
\int_0^{t+\alpha}
\|\nu\|\,G_0^2\,\mathrm d\tau.
\]
Therefore,
\begin{equation}
\label{eq:memory_bound}
M_\nu[g^2](t+\alpha)
\le
\|\nu\|G_0^2(t+\alpha).
\end{equation}
Taking square roots gives
\[
\sqrt{M_\nu[g^2](t+\alpha)}
\le
G_0\sqrt{\|\nu\|}\sqrt{t+\alpha}
=
B\sqrt{t+\alpha}.
\]
Consequently,
\begin{equation}
\label{eq:H_alpha_upper}
H_\alpha(t)
=
\sqrt{M_\nu[g^2](t+\alpha)}+\epsilon
\le
\epsilon+B\sqrt{t+\alpha}.
\end{equation}
Since \(H_\alpha(t)>0\), from equation~(\ref{eq:H_alpha_upper}), we get
\begin{equation}
\label{eq:mu_lower_corrected}
\frac1{H_\alpha(t)}
\ge
\frac1{\epsilon+B\sqrt{t+\alpha}}.
\end{equation}

Now consider the Lyapunov function
\begin{equation}\label{th1:liapunov}
 \mathcal L(t)
:=
\frac12\|\theta(t)-\theta^*\|^2.   
\end{equation}
Differentiating and using the dynamics,
\[
\dot{\mathcal L}(t)
=
\langle \theta(t)-\theta^*,\dot\theta(t)\rangle
=
-
\frac{
\langle \theta(t)-\theta^*,g(t)\rangle
}
{H_\alpha(t)}.
\]
By \(m\)-strong convexity,
\[
\langle \theta(t)-\theta^*,g(t)\rangle
=
\langle \theta(t)-\theta^*,
\nabla f(\theta(t))-\nabla f(\theta^*)\rangle
\ge
m\|\theta(t)-\theta^*\|^2
=
2m\mathcal L(t).
\]
Therefore,
\[
\dot{\mathcal L}(t)
\le
-
\frac{2m}{H_\alpha(t)}
\mathcal L(t).
\]
Using the lower bound (\ref{eq:mu_lower_corrected}), we obtain
\[
\dot{\mathcal L}(t)
\le
-
\frac{2m}{\epsilon+B\sqrt{t+\alpha}}
\mathcal L(t).
\]
By Gronwall's inequality,
\[
\mathcal L(t)
\le
\mathcal L(0)
\exp\!\left(
-2m
\int_0^t
\frac{\mathrm ds}
{\epsilon+B\sqrt{s+\alpha}}
\right).
\]
Taking square roots yields
\[
\|\theta(t)-\theta^*\|
\le
\|\theta(0)-\theta^*\|
\exp\!\left(
-m
\int_0^t
\frac{\mathrm ds}
{\epsilon+B\sqrt{s+\alpha}}
\right),
\]
which proves the first inequality. It remains only to compute the integral explicitly. If \(B>0\), set
\[
u=\sqrt{s+\alpha}.
\]
Then
\[
s=u^2-\alpha,
\qquad
\mathrm ds=2u\,\mathrm du.
\]
Hence
\[
I_\alpha(t)
=
\int_{\sqrt{\alpha}}^{\sqrt{t+\alpha}}
\frac{2u}{\epsilon+Bu}\,\mathrm du.
\]
Since
\[
\frac{2u}{\epsilon+Bu}
=
\frac{2}{B}
-
\frac{2\epsilon}{B(\epsilon+Bu)},
\]
we get
\[
I_\alpha(t)
=
\left[
\frac{2}{B}u
-
\frac{2\epsilon}{B^2}\log(\epsilon+Bu)
\right]_{\sqrt{\alpha}}^{\sqrt{t+\alpha}}.
\]
Therefore,
\[
I_\alpha(t)
=
\frac{2}{B}
\left(
\sqrt{t+\alpha}-\sqrt{\alpha}
\right)
-
\frac{2\epsilon}{B^2}
\log\!\left(
\frac{\epsilon+B\sqrt{t+\alpha}}
{\epsilon+B\sqrt{\alpha}}
\right).
\]
In particular,
\[
I_\alpha(t)
\sim
\frac{2}{B}\sqrt t,
\qquad t\to\infty.
\]
Thus the convergence rate is at least of order \(\exp(-c\sqrt t)\) for suitable constants \(C,c>0\).

Finally, if \(B=0\), then \(G_0=0\), hence \(\Phi_0=0\). Since \(f\) is strongly convex, this implies
\[
\theta(0)=\theta^*.
\]
Therefore \(g(0)=0\), and the constant trajectory \(\theta(t)\equiv \theta^*\) solves the system. This completes the proof.
\end{proof}

Let us emphasize the role of Theorem~\ref{thm:adagrad_rmsprop_memory}. The estimate above already implies that
\[
\theta(t)\to\theta^*
\qquad\text{as }t\to\infty,
\]
for both AdaGrad and RMSProp in the strongly convex setting. However, this theorem should be understood as a preliminary a priori bound rather than as the final sharp convergence rate of the
methods. The reason is that the proof only uses the crude estimate
\[
M_\nu[g^2](t+\alpha)
\le
\|\nu\|G_0^2(t+\alpha),
\]
which treats both AdaGrad and RMSProp in the same way and does not exploit the exponential forgetting present in RMSProp.

Consequently, Theorem~\ref{thm:adagrad_rmsprop_memory} yields the robust subexponential bound
\[
\|\theta(t)-\theta^*\|
\le
C e^{-c\sqrt t}.
\]
As we will see in the following theorem, this estimate is particularly useful for AdaGrad, because \(\exp(-c\sqrt t)\) is integrable on \([0,\infty)\). Hence it can be used to prove that the accumulated gradient memory remains
uniformly bounded. Once this boundedness is obtained, the effective learning factor
\[
\mu(t)
=
\frac{1}{\sqrt{M_\nu[g^2](t+\alpha)}+\epsilon}
\]
is bounded from below by a positive constant, which then leads to exponential decay of the objective gap and of the distance to the minimizer.

For RMSProp, the estimate of Theorem~\ref{thm:adagrad_rmsprop_memory} is not optimal, since RMSProp contains an explicit forgetting term in its memory equation. This additional structure is
captured by the extended memory functional introduced in the next theorem, which separates the cumulative case \(\xi=0\) from the forgetting case \(\xi=\lambda\).

\begin{theorem}[Extended memory functional for AdaGrad/RMSProp]
\label{thm:extended_memory_adagrad_rmsprop}
Let \(f:\mathbb R^n\to\mathbb R\) be \(L\)-smooth and \(m\)-strongly convex, with unique minimizer \(\theta^*\). Let \(g(t):=\nabla f(\theta(t))\), and consider the memory dynamics
\begin{equation}
\label{eq:shifted_memory_dynamics_thm2}
\dot\theta(t)
=
-
\frac{g(t)}
{\sqrt{M_\nu[g^2](t+\alpha)}+\epsilon},
\qquad t\ge 0,
\end{equation}
where \(\alpha>0\), \(\epsilon>0\), and \(M_\nu[g^2]\) is defined as in Definition~\ref{def:M}. Assume that the memory satisfies
\begin{equation}
\label{eq:memory_ode_thm2}
\dot M_\nu[g^2](t)
=
\lambda \|g(t)\|^2-\xi M_\nu[g^2](t),
\qquad 
\lambda>0,
\qquad 
\xi\in\{0,\lambda\}.
\end{equation}
Here \(\xi=0\) corresponds to AdaGrad and \(\xi=\lambda\) to RMSProp. Then there exists a constant \(\mu_0>0\) such that
\[
\frac{1}{\sqrt{M_\nu[g^2](t+\alpha)}+\epsilon}
\ge
\mu_0,
\qquad t\ge 0.
\]
Define
\[
\Phi(t):=f(\theta(t))-f(\theta^*),
\qquad
E_\rho(t):=\Phi(t)+\rho M_\nu[g^2](t),
\]
and choose
\[
\rho:=\frac{\mu_0}{2\lambda}.
\]
Then, for every \(t\ge 0\),
\begin{equation}
\label{eq:E_dissipation_thm2}
\dot E_\rho(t)
\le
-
m\mu_0\Phi(t)
-
\rho\xi M_\nu[g^2](t).
\end{equation}

Consequently:
\begin{enumerate}
\item If \(\xi=\lambda\), i.e. in the RMSProp case, then
\[
E_\rho(t)
\le
E_\rho(0)e^{-\kappa_0 t},
\qquad
\kappa_0:=\min\{m\mu_0,\xi\}>0.
\]
In particular,
\[
\Phi(t)\le E_\rho(0)e^{-\kappa_0 t},
\qquad
M_\nu[g^2](t)
\le
\frac{1}{\rho}E_\rho(0)e^{-\kappa_0 t},
\]
and
\[
\|\theta(t)-\theta^*\|
\le
\sqrt{\frac{2E_\rho(0)}{m}}\,e^{-\kappa_0 t/2}.
\]

\item If \(\xi=0\), i.e. in the AdaGrad case, then
\[
\dot E_\rho(t)\le -m\mu_0\Phi(t)\le 0.
\]
Hence \(E_\rho\) is nonincreasing. Moreover,
\[
\|\theta(t)-\theta^*\|
\le
\|\theta(0)-\theta^*\|e^{-m\mu_0 t},
\]
and \(M_\nu[g^2](t)\) converges to a finite limit as \(t\to\infty\). In general, this limit is not zero.
\end{enumerate}
\end{theorem}

\begin{proof}
We first treat the nontrivial case \(\theta(0)\neq \theta^*\). By Theorem~\ref{thm:adagrad_rmsprop_memory}, we know that
\[
\|\theta(t)-\theta^*\|
\le
R_0 e^{-mI_\alpha(t)},
\qquad
R_0:=\|\theta(0)-\theta^*\|.
\]
Since \(f\) is \(L\)-smooth and \(\nabla f(\theta^*)=0\), we have
\[
\|g(t)\|
=
\|\nabla f(\theta(t))-\nabla f(\theta^*)\|
\le
L\|\theta(t)-\theta^*\|.
\]
Therefore,
\begin{equation}
\label{eq:g_decay_from_thm1}
\|g(t)\|^2
\le
L^2R_0^2 e^{-2mI_\alpha(t)}.
\end{equation}

Moreover, by the explicit expression of \(I_\alpha(t)\), we have
\[
I_\alpha(t)\sim \frac{2}{B}\sqrt t,
\qquad t\to\infty.
\]
Hence
\[
e^{-2mI_\alpha(t)}
\sim
e^{-\frac{4m}{B}\sqrt t},
\qquad t\to\infty.
\]
Since \(\exp(-c\sqrt t)\) is integrable on \([0,\infty)\) for every \(c>0\), it follows that
\[
\mathcal J_\alpha
:=
\int_0^\infty e^{-2mI_\alpha(s)}\,\mathrm ds
<
\infty.
\]

We now prove that the memory is uniformly bounded.  Solving the linear equation
\begin{equation}\label{eq:M_nu_differential}
\dot M_\nu[g^2](t)
=
\lambda\|g(t)\|^2-\xi M_\nu[g^2](t)
\end{equation}
gives
\[
M_\nu[g^2](t)
=
e^{-\xi t}M_\nu[g^2](0)
+
\lambda
\int_0^t
e^{-\xi(t-s)}\|g(s)\|^2\,\mathrm ds.
\]
Since \(\xi\ge 0\), we have \(e^{-\xi(t-s)}\le 1\). Hence, using equation~(\ref{eq:g_decay_from_thm1}),
\[
M_\nu[g^2](t)
\le
M_\nu[g^2](0)
+
\lambda
\int_0^t
\|g(s)\|^2\,\mathrm ds
\le
M_\nu[g^2](0)
+
\lambda L^2R_0^2
\int_0^t
e^{-2mI_\alpha(s)}\,\mathrm ds.
\]
Therefore,
\[
M_\nu[g^2](t)
\le
M_\nu[g^2](0)
+
\lambda L^2R_0^2\mathcal J_\alpha
=
\overline M,
\qquad
t\ge 0.
\]
In particular,
\[
M_\nu[g^2](t+\alpha)\le \overline M,
\qquad
t\ge 0.
\]
Consequently,
\[
\sqrt{M_\nu[g^2](t+\alpha)}+\epsilon
\le
\sqrt{\overline M}+\epsilon,
\]
and so
\begin{equation}
\label{eq:mu_lower_thm2}
\mu(t)
:=
\frac{1}{\sqrt{M_\nu[g^2](t+\alpha)}+\epsilon}
\ge
\frac{1}{\epsilon+\sqrt{\overline M}}
=
\mu_0.
\end{equation}

We now compute the derivative of the objective gap $\Phi$. Since
\[
\dot\theta(t)
=
-
\mu(t)g(t),
\]
we obtain
\begin{equation}
\label{eq:Phi_dot_thm2}
\dot\Phi(t)
=
\langle g(t),\dot\theta(t)\rangle
=
-\mu(t)\|g(t)\|^2.
\end{equation}
For
\[
E_\rho(t)
=
\Phi(t)+\rho M_\nu[g^2](t),
\]
we have
\[
\dot E_\rho(t)
=
\dot\Phi(t)+\rho \dot M_\nu[g^2](t).
\]
Substituting equations (\ref{eq:M_nu_differential}) and (\ref{eq:Phi_dot_thm2}) gives
\begin{equation}
\label{eq:E_dot_raw_thm2}
\dot E_\rho(t)
=
-
\bigl(\mu(t)-\rho\lambda\bigr)\|g(t)\|^2
-
\rho\xi M_\nu[g^2](t).
\end{equation}
Choosing
\[
\rho
=
\frac{\mu_0}{2\lambda},
\]
and using \(\mu(t)\ge \mu_0\), we get
\[
\mu(t)-\rho\lambda
\ge
\mu_0-\frac{\mu_0}{2}
=
\frac{\mu_0}{2}.
\]
Therefore,
\begin{equation}
\label{eq:E_dot_first_bound_thm2}
\dot E_\rho(t)
\le
-
\frac{\mu_0}{2}\|g(t)\|^2
-
\rho\xi M_\nu[g^2](t).
\end{equation}

By Lemma~\ref{lem:grad_bounds}, since \(f\) is \(m\)-strongly convex and \(L\)-smooth,
\[
\Phi(t)
=
f(\theta(t))-f(\theta^*)
\le
\frac{1}{2m}\|g(t)\|^2.
\]
Hence
\[
\|g(t)\|^2
\ge
2m\Phi(t).
\]
Substituting this into equation~(\ref{eq:E_dot_first_bound_thm2}), we obtain
\[
\dot E_\rho(t)
\le
-
m\mu_0\Phi(t)
-
\rho\xi M_\nu[g^2](t),
\]
which proves equation~(\ref{eq:E_dissipation_thm2}). We now distinguish the two cases:

\medskip

\noindent\textbf{RMSProp case: \(\xi=\lambda\).}
If \(\xi=\lambda>0\), then
\[
\dot E_\rho(t)
\le
-
m\mu_0\Phi(t)
-
\rho\xi M_\nu[g^2](t).
\]
Let
\[
\kappa_0
:=
\min\{m\mu_0,\xi\}.
\]
Then
\[
m\mu_0\Phi(t)
+
\rho\xi M_\nu[g^2](t)
\ge
\kappa_0
\bigl(
\Phi(t)+\rho M_\nu[g^2](t)
\bigr)
=
\kappa_0 E_\rho(t).
\]
Therefore,
\[
\dot E_\rho(t)
\le
-\kappa_0 E_\rho(t).
\]
By Gronwall's inequality,
\[
E_\rho(t)
\le
E_\rho(0)e^{-\kappa_0 t}.
\]
Since both terms in \(E_\rho(t)=\Phi(t)+\rho M_\nu[g^2](t)\) are nonnegative, we get
\[
\Phi(t)
\le
E_\rho(0)e^{-\kappa_0 t},
\]
and
\[
M_\nu[g^2](t)
\le
\frac{1}{\rho}E_\rho(0)e^{-\kappa_0 t}.
\]
Finally, strong convexity gives
\[
\Phi(t)
\ge
\frac m2\|\theta(t)-\theta^*\|^2.
\]
Thus
\[
\|\theta(t)-\theta^*\|
\le
\sqrt{\frac{2}{m}\Phi(t)}
\le
\sqrt{\frac{2E_\rho(0)}{m}}e^{-\kappa_0 t/2}.
\]
This proves the RMSProp claims.

\medskip

\noindent\textbf{AdaGrad case: \(\xi=0\).}
If \(\xi=0\), then equation~(\ref{eq:E_dissipation_thm2}) becomes
\[
\dot E_\rho(t)
\le
-
m\mu_0\Phi(t)
\le 0.
\]
Hence \(E_\rho\) is nonincreasing. However, because there is no negative term proportional to \(M_\nu[g^2](t)\), this inequality cannot be closed in the form
\[
\dot E_\rho(t)\le -\kappa E_\rho(t)
\]
with some \(\kappa>0\). This reflects the purely accumulative character of AdaGrad. Nevertheless, the distance to the minimizer itself decays exponentially with the initial constant appearing in the statement. Indeed, define
\[
\mathcal L(t)
:=
\frac12\|\theta(t)-\theta^*\|^2.
\]
Since
\[
\dot\theta(t)=-\mu(t)g(t),
\qquad
\mu(t):=
\frac{1}{\sqrt{M_\nu[g^2](t+\alpha)}+\epsilon},
\]
we have
\[
\dot{\mathcal L}(t)
=
\langle \theta(t)-\theta^*,\dot\theta(t)\rangle
=
-\mu(t)\langle \theta(t)-\theta^*,g(t)\rangle.
\]
By \(m\)-strong convexity,
\[
\langle \theta(t)-\theta^*,g(t)\rangle
=
\langle \theta(t)-\theta^*,
\nabla f(\theta(t))-\nabla f(\theta^*)\rangle
\ge
m\|\theta(t)-\theta^*\|^2
=
2m\mathcal L(t).
\]
Using also \(\mu(t)\ge\mu_0\), we obtain
\[
\dot{\mathcal L}(t)
\le
-2m\mu_0\mathcal L(t).
\]
Therefore, by Gronwall's inequality,
\[
\mathcal L(t)
\le
\mathcal L(0)e^{-2m\mu_0t}.
\]
Taking square roots gives
\[
\|\theta(t)-\theta^*\|
\le
\|\theta(0)-\theta^*\|e^{-m\mu_0t}.
\]

Finally, in the AdaGrad case,
\[
\dot M_\nu[g^2](t)
=
\lambda\|g(t)\|^2
\ge 0,
\]
so \(M_\nu[g^2](t)\) is nondecreasing. Moreover, since \(f\) is \(L\)-smooth and \(\nabla f(\theta^*)=0\), the previous estimate gives
\[
\|g(t)\|
=
\|\nabla f(\theta(t))-\nabla f(\theta^*)\|
\le
L\|\theta(t)-\theta^*\|
\le
L\|\theta(0)-\theta^*\|e^{-m\mu_0t}.
\]
Hence \(\|g(t)\|^2\in L^1(0,\infty)\). Therefore
\[
M_\nu[g^2](t)
=
M_\nu[g^2](0)
+
\lambda\int_0^t\|g(s)\|^2\,\mathrm ds
\]
converges to a finite limit as \(t\to\infty\). In general, this limit is positive, so AdaGrad retains a persistent accumulated memory.

The case \(\theta(0)=\theta^*\) is immediate: then \(g(0)=0\), and the constant trajectory
\[
\theta(t)\equiv \theta^*
\]
solves the dynamics, with zero memory evolution if the memory is initialized at zero. This completes the proof.
\end{proof}

Once again, we emphasize this theorem. In short, \(E_\rho\) couples state and memory. The proof first uses Theorem~\ref{thm:adagrad_rmsprop_memory} to obtain a uniform lower bound \(\mu(t)\ge \mu_0>0\), which then yields the dissipation estimate in equation~(\ref{eq:E_dissipation_thm2}). When \(\xi=\lambda\) (RMSProp), the forgetting term \(-\xi M_\nu[g^2](t)\) closes the estimate as \(\dot E_\rho\le -\kappa_0 E_\rho\), with \(\kappa_0=\min\{m\mu_0,\xi\}>0\), and therefore \(E_\rho(t)\), \(\Phi(t)\), and \(M_\nu[g^2](t)\) all decay exponentially. In particular, no persistent memory remains in the limit. In contrast, when \(\xi=0\) (AdaGrad), \(E_\rho\) is nonincreasing, but the estimate cannot be closed as an exponential decay for \(E_\rho\), since there is no forgetting term. Nevertheless, the objective gap still decays exponentially, and \(M_\nu[g^2](t)\) converges to a finite, generally nonzero, limit, consistently with the cumulative nature of AdaGrad.

We now turn to the case of Adam. This case must be treated separately because the dynamics are not governed only by a normalized gradient and a scalar memory term. Instead, the update involves a shifted first moment and an effective time-dependent gain.

In the Lyapunov computation for Adam, an additional term appears involving the logarithmic derivative of this effective gain. Therefore, unlike in the AdaGrad/RMSProp case, stability cannot be obtained only from the positivity or boundedness of the memory kernel. We need a scalar condition ensuring that the effective Adam gain does not grow too fast compared with the relaxation rate of the first moment. The following definition encodes precisely this requirement.

\begin{definition}[Scalar stability margin for the Adam dynamics]
\label{def:adam_scalar_margin}
For \(0\leq \beta<1\) and \(q\geq1\), define
\[
    \psi_\beta(q)
    :=
    \begin{cases}
    \displaystyle
    \frac{-\log\beta}{\exp\bigl((-\log\beta)q\bigr)-1},
    & 0<\beta<1,\\[1.2em]
    0,
    & \beta=0.
    \end{cases}
\]
For \(0\leq\beta_1,\beta_2<1\), define
\begin{equation}\label{eq:H_beta_def}
    \mathcal H_{\beta_1,\beta_2}
    :=
    \sup_{q\geq1}
    \left\{
        \frac12\psi_{\beta_2}(q)
        -
        \psi_{\beta_1}(q)
        +
        \frac{1-\beta_2}{2}
    \right\}.
\end{equation}
We say that the Adam dynamics satisfies the scalar Lyapunov condition if
\begin{equation}\label{eq:natural_scalar_condition}
    \mathcal H_{\beta_1,\beta_2}
    <
    2(1-\beta_1).
\end{equation}
In that case we define
\begin{equation}\label{eq:c_beta_def}
    c_\beta
    :=
    1-
    \frac{\mathcal H_{\beta_1,\beta_2}}
    {2(1-\beta_1)}
    >0.
\end{equation}
\end{definition}

\begin{lemma}[Logarithmic growth of the Adam gain]
\label{lem:h_bound_shifted_adam}
Let \(0\leq\beta_1,\beta_2<1\), \(\alpha>0\), and set
\[
    \lambda_1:=\frac{1-\beta_1}{\alpha},
    \qquad
    \lambda_2:=\frac{1-\beta_2}{\alpha}.
\]
Let $\widehat v(t):=v(t+\alpha)$, and define the Adam gain
\begin{equation}\label{eq:ahat_lemma_def}
    \widehat a(t)
    :=
    \frac{\eta(t+\alpha)}
    {\sqrt{\widehat v(t)}+\varepsilon(t+\alpha)},
\end{equation}
where $\eta(t)$ and $\varepsilon(t)$ are defined in Proposition~\ref{prop:Adam}. Assume that \(\widehat v\) satisfies
\[
    \dot{\widehat v}(t)+\lambda_2\widehat v(t)
    =
    \lambda_2\|g(t+\alpha)\|^2.
\]
Define
\[
    h(t):=\frac{\mathrm d}{\mathrm dt}\log \widehat a(t).
\]
Then, for all \(t\geq0\),
\begin{equation}\label{eq:h_bound_by_H}
    h(t)
    \leq
    \frac{\mathcal H_{\beta_1,\beta_2}}{\alpha}.
\end{equation}
Consequently, if
\[
    \mathcal H_{\beta_1,\beta_2}<2(1-\beta_1),
\]
then
\begin{equation}\label{eq:h_less_than_2lambda}
    h(t)<2\lambda_1,
    \qquad t\geq0.
\end{equation}
\end{lemma}

\begin{proof}
We write
\[
    \widehat a(t)
    =
    \eta(t+\alpha)
    \left(\sqrt{\widehat v(t)}+\varepsilon(t+\alpha)\right)^{-1}.
\]
Therefore
\[
    h(t)
    =
    \frac{\mathrm d}{\mathrm dt}\log\eta(t+\alpha)
    +
    \frac{\mathrm d}{\mathrm dt}
    \log
    \left(
        \frac{1}{\sqrt{\widehat v(t)}+\varepsilon(t+\alpha)}
    \right).
\]

We first estimate the second term. Since
\[
    \dot{\widehat v}(t)+\lambda_2\widehat v(t)
    =
    \lambda_2\|g(t+\alpha)\|^2,
\]
we have
\[
    \dot{\widehat v}(t)\geq -\lambda_2\widehat v(t).
\]
Moreover,
\[
    \varepsilon(t+\alpha)
    =
    \epsilon\sqrt{1-\beta_2^{1+t/\alpha}}
\]
is nondecreasing in \(t\). Hence, at every point where \(\widehat v(t)>0\),
\[
\begin{aligned}
    \frac{\mathrm d}{\mathrm dt}
    \log
    \left(
        \frac{1}{\sqrt{\widehat v(t)}+\varepsilon(t+\alpha)}
    \right)
    &=
    -
    \frac{
        \dfrac{\dot{\widehat v}(t)}{2\sqrt{\widehat v(t)}}
        +
        \dot\varepsilon(t+\alpha)
    }
    {
        \sqrt{\widehat v(t)}+\varepsilon(t+\alpha)
    }\\
    &\leq
    \frac{\lambda_2}{2}
    \frac{\sqrt{\widehat v(t)}}
    {\sqrt{\widehat v(t)}+\varepsilon(t+\alpha)}
    \leq
    \frac{\lambda_2}{2}.
\end{aligned}
\]
At points where \(\widehat v(t)=0\), it follows by approximation from \(\widehat v+\delta\) and then letting \(\delta\downarrow0\). Thus
\begin{equation}\label{eq:h_pre_estimate}
    h(t)
    \leq
    \frac{\mathrm d}{\mathrm dt}\log\eta(t+\alpha)
    +
    \frac{\lambda_2}{2}.
\end{equation}

Now put
\[
    q:=1+\frac{t}{\alpha}.
\]
Then \(q\geq1\), and
\[
    \eta(t+\alpha)
    =
    \frac{\sqrt{1-\beta_2^q}}
    {1-\beta_1^q}.
\]
Using Definition~\ref{def:adam_scalar_margin}, direct differentiation gives
\[
    \frac{\mathrm d}{\mathrm dt}\log\eta(t+\alpha)
    =
    \frac1\alpha
    \left[
        \frac12\psi_{\beta_2}(q)
        -
        \psi_{\beta_1}(q)
    \right].
\]
Also,
\[
    \frac{\lambda_2}{2}
    =
    \frac{1-\beta_2}{2\alpha}.
\]
Substituting this into equation~(\ref{eq:h_pre_estimate}), we obtain
\[
    h(t)
    \leq
    \frac1\alpha
    \left[
        \frac12\psi_{\beta_2}(q)
        -
        \psi_{\beta_1}(q)
        +
        \frac{1-\beta_2}{2}
    \right].
\]
Taking the supremum over \(q\geq1\), we get
\[
    h(t)
    \leq
    \frac{\mathcal H_{\beta_1,\beta_2}}{\alpha}.
\]
Finally, since
\[
    2\lambda_1
    =
    \frac{2(1-\beta_1)}{\alpha},
\]
the condition
\[
    \mathcal H_{\beta_1,\beta_2}<2(1-\beta_1)
\]
implies
\[
    h(t)
    <
    2\lambda_1.
\]
This proves the lemma.
\end{proof}

The previous lemma explains the role of the scalar Lyapunov condition. It provides a uniform upper bound for the logarithmic growth of the Adam gain. In the Lyapunov functional below, this bound is exactly what makes the coefficient of \(\|\widehat m(t)\|^2\) strictly negative. Thus, the condition \(H_{\beta_1,\beta_2}<2(1-\beta_1)\) is not an auxiliary technical artifact; it is precisely the condition needed to close the Lyapunov estimate.

\begin{theorem}[\textbf{Conditional convergence and stability for the Adam flow}]
\label{thm:adam_shifted}
Let \(f:\mathbb R^n\to\mathbb R\) be \(L\)-smooth and \(m\)-strongly convex, with unique minimizer \(\theta^*\). Let $g(t):=\nabla f(\theta(t))$, and assume that \(0\leq\beta_1,\beta_2<1\), \(\alpha>0\), \(\epsilon>0\). Consider a global classical solution of the continuous Adam dynamics
\begin{equation}\label{eq:adam_shifted_dyn}
    \dot\theta(t)
    =
    -
    \widehat a(t)\widehat m(t),
    \qquad t\geq0,
\end{equation}
where
\[
    \widehat m(t):=m(t+\alpha),
    \qquad
    \widehat v(t):=v(t+\alpha),
\]
and
\begin{equation}\label{eq:ahat_def}
    \widehat a(t)
    :=
    \frac{\eta(t+\alpha)}
    {\sqrt{\widehat v(t)}+\varepsilon(t+\alpha)}\,,
\end{equation}
where $\eta(t)$ and $\varepsilon(t)$ are defined in Proposition~\ref{prop:Adam}, and the shifted moments satisfy
\begin{equation}\label{eq:shifted_moment_odes}
    \dot{\widehat m}(t)+\lambda_1\widehat m(t)
    =
    \lambda_1 g(t+\alpha),
    \qquad
    \dot{\widehat v}(t)+\lambda_2\widehat v(t)
    =
    \lambda_2\|g(t+\alpha)\|^2,
\end{equation}
with
\[
    \lambda_1:=\frac{1-\beta_1}{\alpha},
    \qquad
    \lambda_2:=\frac{1-\beta_2}{\alpha}.
\]

Assume the scalar Lyapunov condition
\begin{equation}\label{eq:H_condition_theorem}
    \mathcal H_{\beta_1,\beta_2}
    <
    2(1-\beta_1),
\end{equation}
and set
\[
    c_\beta
    :=
    1-
    \frac{\mathcal H_{\beta_1,\beta_2}}
    {2(1-\beta_1)}
    >0.
\]

Assume, moreover, that the shift defect satisfies the one-sided estimate\footnote{This is an additional structural assumption on the Adam flow; it is not claimed here to follow from \(L\)-smoothness and strong convexity
alone.}
\begin{equation}\label{eq:shift_defect_onesided}
    \bigl\langle g(t)-g(t+\alpha),\widehat m(t)\bigr\rangle
    \geq
    -\sigma_\alpha\|\widehat m(t)\|^2,
    \qquad t\geq0,
\end{equation}
for some constant \(0\leq\sigma_\alpha<c_\beta\). Then the functional
\begin{equation}\label{eq:adam_shifted_lyapunov}
    V(t)
    :=
    \Phi(t)
    +
    \frac{\widehat a(t)}{2\lambda_1}
    \|\widehat m(t)\|^2
\end{equation}
satisfies
\begin{equation}\label{eq:adam_shifted_dissipation}
    \dot V(t)
    \leq
    -
    \widehat a(t)
    \bigl(c_\beta-\sigma_\alpha\bigr)
    \|\widehat m(t)\|^2
    \leq0,
\end{equation}
where $\Phi(t) = f(\theta(t))-f(\theta^*).$ Consequently, \(V\) is a Lyapunov-type dissipation functional for the Adam dynamics \((\theta^*, \widehat{m}, \widehat{v})\), and
\[
    \widehat m(t)\to0,
    \qquad
    g(t)\to0,
    \qquad
    \theta(t)\to\theta^*,
    \qquad
    t\to\infty.
\]
\end{theorem}

\begin{proof}
Define
\[
    \Phi(t):=f(\theta(t))-f(\theta^*).
\]
Using the Adam dynamics,
\[
    \dot\Phi(t)
    =
    \langle g(t),\dot\theta(t)\rangle
    =
    -
    \widehat a(t)
    \langle g(t),\widehat m(t)\rangle.
\]
From the shifted first-moment equation,
\[
    \dot{\widehat m}(t)+\lambda_1\widehat m(t)
    =
    \lambda_1g(t+\alpha),
\]
we obtain
\[
    g(t+\alpha)
    =
    \widehat m(t)+\frac1{\lambda_1}\dot{\widehat m}(t).
\]
Therefore
\[
    \langle g(t+\alpha),\widehat m(t)\rangle
    =
    \|\widehat m(t)\|^2
    +
    \frac1{2\lambda_1}
    \frac{\mathrm d}{\mathrm dt}\|\widehat m(t)\|^2.
\]
Now we decompose
\[
    g(t)
    =
    g(t+\alpha)
    +
    \bigl(g(t)-g(t+\alpha)\bigr).
\]
Hence
\[
\begin{aligned}
    \langle g(t),\widehat m(t)\rangle
    &=
    \langle g(t+\alpha),\widehat m(t)\rangle
    +
    \langle g(t)-g(t+\alpha),\widehat m(t)\rangle\\
    &=
    \|\widehat m(t)\|^2
    +
    \frac1{2\lambda_1}
    \frac{\mathrm d}{\mathrm dt}\|\widehat m(t)\|^2
    +
    \langle g(t)-g(t+\alpha),\widehat m(t)\rangle.
\end{aligned}
\]
Substituting this into \(\dot\Phi(t)\), we get
\begin{equation}\label{eq:Phi_dot_shifted}
    \dot\Phi(t)
    =
    -
    \widehat a(t)\|\widehat m(t)\|^2
    -
    \frac{\widehat a(t)}{2\lambda_1}
    \frac{\mathrm d}{\mathrm dt}\|\widehat m(t)\|^2
    -
    \widehat a(t)
    \langle g(t)-g(t+\alpha),\widehat m(t)\rangle.
\end{equation}

Now we differentiate
\[
    V(t)
    =
    \Phi(t)
    +
    \frac{\widehat a(t)}{2\lambda_1}
    \|\widehat m(t)\|^2.
\]
Then
\[
    \dot V(t)
    =
    \dot\Phi(t)
    +
    \frac{\dot{\widehat a}(t)}{2\lambda_1}
    \|\widehat m(t)\|^2
    +
    \frac{\widehat a(t)}{2\lambda_1}
    \frac{\mathrm d}{\mathrm dt}\|\widehat m(t)\|^2.
\]
Using equation~(\ref{eq:Phi_dot_shifted}), the terms involving $\frac{\mathrm d}{\mathrm dt}\|\widehat m(t)\|^2$ cancel. Hence
\[
    \dot V(t)
    =
    -
    \widehat a(t)
    \left(
        1-
        \frac{1}{2\lambda_1}
        \frac{\dot{\widehat a}(t)}{\widehat a(t)}
    \right)
    \|\widehat m(t)\|^2
    -
    \widehat a(t)
    \langle g(t)-g(t+\alpha),\widehat m(t)\rangle.
\]
Since
\[
    h(t):=
    \frac{\dot{\widehat a}(t)}{\widehat a(t)}
    =
    \frac{\mathrm d}{\mathrm dt}\log\widehat a(t),
\]
Lemma~\ref{lem:h_bound_shifted_adam} gives
\[
    h(t)
    \leq
    \frac{\mathcal H_{\beta_1,\beta_2}}{\alpha}
    \qquad 
    \text{and assumes that}
    \qquad
    \mathcal H_{\beta_1,\beta_2} < 2(1-\beta_1).
\]
Because $2\lambda_1 = \frac{2(1-\beta_1)}{\alpha}$, we get
\[
    1-\frac{h(t)}{2\lambda_1}
    \geq
    1-
    \frac{\mathcal H_{\beta_1,\beta_2}}
    {2(1-\beta_1)}
    =
    c_\beta > 0.
\]
Therefore
\[
    \dot V(t)
    \leq
    -
    \widehat a(t)c_\beta\|\widehat m(t)\|^2
    -
    \widehat a(t)
    \langle g(t)-g(t+\alpha),\widehat m(t)\rangle.
\]
Using the shift-defect condition,
\[
    \langle g(t)-g(t+\alpha),\widehat m(t)\rangle
    \geq
    -\sigma_\alpha\|\widehat m(t)\|^2,
\]
we obtain
\[
    \dot V(t)
    \leq
    -
    \widehat a(t)c_\beta\|\widehat m(t)\|^2
    +
    \widehat a(t)\sigma_\alpha\|\widehat m(t)\|^2.
\]
Hence
\[
    \dot V(t)
    \leq
    -
    \widehat a(t)
    (c_\beta-\sigma_\alpha)
    \|\widehat m(t)\|^2
    \leq0,
\]
because \(\sigma_\alpha<c_\beta\). This proves the Lyapunov dissipation estimate.

It remains to prove convergence. Since \(V(t)\) is nonincreasing and
\[
    \Phi(t)\leq V(t)\leq V(0),
\]
strong convexity gives
\[
    \frac m2\|\theta(t)-\theta^*\|^2
    \leq
    \Phi(t)
    \leq
    V(0).
\]
Thus \(\theta(t)\) is bounded. Since \(f\) is \(L\)-smooth, \(g(t)\) is bounded. Since \(g(t)\) is bounded, there exists \(G>0\) such that
\[
    \|g(t)\|\leq G,\qquad t\geq0.
\]
Hence \(\|g(t+\alpha)\|^2\leq G^2\). By solving the differential equation
\[
    \dot{\widehat v}(t)+\lambda_2\widehat v(t)
    =
    \lambda_2\|g(t+\alpha)\|^2,
    \qquad \lambda_2>0,
\]
we obtain
\[
    \widehat v(t)
    =
    e^{-\lambda_2 t}\widehat v(0)
    +
    \lambda_2
    \int_0^t
    e^{-\lambda_2(t-s)}
    \|g(s+\alpha)\|^2\,\mathrm ds.
\]
Therefore
\[
    \widehat v(t)
    \leq
    e^{-\lambda_2 t}\widehat v(0)
    +
    G^2(1-e^{-\lambda_2 t})
    \leq
    \max\{\widehat v(0),G^2\}.
\]
Thus \(\widehat v(t)\) is bounded on \([0,\infty)\). Moreover, because $q=1+\frac{t}{\alpha}\geq1$, the factors
\[
    \eta(t+\alpha)
    =
    \frac{\sqrt{1-\beta_2^q}}{1-\beta_1^q},
    \qquad
    \varepsilon(t+\alpha)
    =
    \epsilon\sqrt{1-\beta_2^q}
\]
are bounded away from their singular regimes. Hence there exist constants \(a_*,a^*>0\) such that
\[
    0<a_*\leq \widehat a(t)\leq a^*<\infty,
    \qquad t\geq0.
\]

Set $\gamma:=c_\beta-\sigma_\alpha>0$. From the dissipation inequality,
\[
    \dot V(t)
    \leq
    -\gamma \widehat a(t)\|\widehat m(t)\|^2.
\]
Hence, for every \(T>0\),
\[
    \gamma
    \int_0^T
    \widehat a(t)\|\widehat m(t)\|^2\,\mathrm dt
    \leq
    V(0)-V(T) \leq V(0).
\]
Letting \(T\to\infty\), we get
\[
    \int_0^\infty
    \widehat a(t)\|\widehat m(t)\|^2\,\mathrm dt
    \leq
    \frac{V(0)}{\gamma}
    <\infty.
\]
Moreover, since there exists \(a_*>0\) such that
\[
    \widehat a(t)\geq a_*,
    \qquad t\geq0,
\]
we have
\[
    a_*
    \int_0^\infty
    \|\widehat m(t)\|^2\,\mathrm dt
    \leq
    \int_0^\infty
    \widehat a(t)\|\widehat m(t)\|^2\,\mathrm dt
    <\infty.
\]
Therefore
\[
    \int_0^\infty
    \|\widehat m(t)\|^2\,\mathrm dt
    <\infty.
\]
Also, since \(V(t)=\Phi(t)+\frac{\widehat a(t)}{2\lambda_1}\|\widehat m(t)\|^2\leq V(0)\) and \(\widehat a(t)\geq a_*>0\) and $\Phi(t)\geq 0$, the quantity \(\widehat m(t)\) is bounded. From
\[
    \dot{\widehat m}(t)
    =
    \lambda_1\bigl(g(t+\alpha)-\widehat m(t)\bigr),
\]
and since both \(g(t+\alpha)\) and \(\widehat m(t)\) are bounded, \(\dot{\widehat m}(t)\) is bounded. Therefore \(\widehat m(t)\) is uniformly continuous \cite[Theorem 5.10]{Rudin1976}. By Barbalat's lemma  \cite{Barbalat1959},
\[
    \widehat m(t)\to0.
\]

Since $\dot\theta(t) = -\widehat a(t)\widehat m(t)$, and \(\widehat a(t)\) is bounded, we also have
\[
    \dot\theta(t)\to0.
\]
Because \(f\) is \(L\)-smooth and \(\theta(t)\) is uniformly continuous, \(g(t)\) is uniformly continuous. Hence \(\dot{\widehat m}(t)\) is uniformly continuous. We now use the following standard consequence of Barbalat's lemma: if \(y(t)\) has a finite limit as \(t\to\infty\) and \(\dot y(t)\) is uniformly continuous on \([0,\infty)\), then \(\dot y(t)\to0\). Applying this result componentwise to \(y(t)=\widehat m(t)\), and using that \(\widehat m(t)\to0\) and \(\dot{\widehat m}(t)\) is uniformly continuous,
we obtain
\[
    \dot{\widehat m}(t)\to0.
\]
Using again $g(t+\alpha) = \widehat m(t)+\frac1{\lambda_1}\dot{\widehat m}(t)$, we conclude that
\[
    g(t+\alpha)\to0.
\]
Equivalently,
\[
    g(t)\to0.
\]
Finally, strong convexity implies
\[
    m\|\theta(t)-\theta^*\|
    \leq
    \|g(t)\|,
\]
and therefore
\[
    \theta(t)\to\theta^*.
\]
This completes the proof.
\end{proof}

Let us emphasize the content of this theorem. Under \(m\)-strong convexity and \(L\)-smoothness, together with the scalar Lyapunov condition and the one-sided shift-defect estimate, the Adam flow admits the Lyapunov functional
\[
V(t)=\Phi(t)+\frac{\widehat a(t)}{2\lambda_1}\|\widehat m(t)\|^2.
\]
The key point is that the dissipation of \(V\) is governed by the logarithmic growth of the effective Adam gain,
\[
h(t)=\frac{d}{dt}\log \widehat a(t).
\]
More precisely, the scalar condition \(h(t)<2\lambda_1\) is what makes the Lyapunov argument close. The presence of the shift \(t+\alpha\) produces an additional term involving \(g(t)-g(t+\alpha)\). This is why the theorem also requires the one-sided shift-defect estimate. Under these hypotheses, \(V\) is nonincreasing and the dissipation inequality implies
\[
\widehat m(t)\to0,\qquad g(t)\to0,\qquad \theta(t)\to\theta^*
\]
as \(t\to\infty\).

The shift-defect condition used in Theorem~\ref{thm:adam_shifted} may look somewhat abstract at first sight. We now provide a simple sufficient condition, based only on the \(L\)-smoothness of \(f\) and a local control of the Adam update over intervals of length \(\alpha\).

\begin{proposition}[Sufficient condition for the shift-defect estimate]
Assume that \(f\) is \(L\)-smooth. Let the Adam dynamics be given by
\[
    \dot\theta(t)=-\widehat a(t)\widehat m(t).
\]
Suppose that there exists a constant \(C_{\mathrm{loc}}>0\) such that, for
all \(t\geq0\) with \(\widehat m(t)\neq0\),
\[
    \sup_{s\in[t,t+\alpha]}
    \widehat a(s)\|\widehat m(s)\|
    \leq
    C_{\mathrm{loc}}\|\widehat m(t)\|.
\]
Then the shift-defect estimate
\[
    \bigl\langle g(t)-g(t+\alpha),\widehat m(t)\bigr\rangle
    \geq
    -\sigma_\alpha\|\widehat m(t)\|^2
\]
holds with
\[
    \sigma_\alpha:=L\alpha C_{\mathrm{loc}}.
\]
In particular, if
\[
    L\alpha C_{\mathrm{loc}}<c_\beta,
\]
then the shift-defect condition required in Theorem~\ref{thm:adam_shifted} is satisfied.
\end{proposition}

\begin{proof}
If \(\widehat m(t)=0\), the estimate is trivial. Assume therefore that \(\widehat m(t)\neq0\). By the Cauchy-Schwarz inequality and the \(L\)-smoothness of \(f\), we have
\[
\begin{aligned}
    \bigl\langle g(t)-g(t+\alpha),\widehat m(t)\bigr\rangle
    &\geq
    -\|g(t)-g(t+\alpha)\|\,\|\widehat m(t)\| \\
    &\geq
    -L\|\theta(t+\alpha)-\theta(t)\|\,\|\widehat m(t)\|.
\end{aligned}
\]
Using the dynamics,
\[
    \theta(t+\alpha)-\theta(t)
    =
    -\int_t^{t+\alpha}
    \widehat a(s)\widehat m(s)\,\mathrm ds .
\]
Hence
\[
\begin{aligned}
    \|\theta(t+\alpha)-\theta(t)\|
    &\leq
    \int_t^{t+\alpha}
    \widehat a(s)\|\widehat m(s)\|\,\mathrm ds \\
    &\leq
    \alpha
    \sup_{s\in[t,t+\alpha]}
    \widehat a(s)\|\widehat m(s)\| \\
    &\leq
    \alpha C_{\mathrm{loc}}\|\widehat m(t)\|.
\end{aligned}
\]
Combining the previous estimates gives
\[
    \bigl\langle g(t)-g(t+\alpha),\widehat m(t)\bigr\rangle
    \geq
    -L\alpha C_{\mathrm{loc}}\|\widehat m(t)\|^2.
\]
Thus the shift-defect estimate holds with
\[
    \sigma_\alpha=L\alpha C_{\mathrm{loc}}.
\]
The condition \(L\alpha C_{\mathrm{loc}}<c_\beta\) then guarantees \(\sigma_\alpha<c_\beta\), as required in Theorem~\ref{thm:adam_shifted}.
\end{proof}

\subsection{Case II: NonConvex Objective Functions}
In this subsection, we work under the regularity assumption---Assumption~\ref{ass:regularidad}---and allow functions that are not necessarily convex \cite{Nesterov2004}. The following assumption for the nonconvex analysis is:
\begin{assumption}[Lower bound and precompact level set]\label{ass:nonconvexidad}
Let \(f:\mathbb{R}^n\to\mathbb{R}\) be of class \(C^1\). $f$ is bounded below, and the sublevel set 
    \[
    \mathcal{S}_0 := \{\theta \in \mathbb{R}^n : f(\theta) \leq f(\theta(0))\}
    \]
is compact.
\end{assumption}
We also make use of two standard conditions (\cite{PL,KL}) for functions that need not be convex:
\begin{itemize}
    \item \textbf{Polyak–Łojasiewicz (PL) inequality.} Let \(f:\mathbb{R}^n\to\mathbb{R}\) be of class \(C^1\). There exists $C_{PL}>0$ such that
    \[
    \|\nabla f(\theta)\|^2 \;\geq\; C_{PL}\,[f(\theta)-f^*],
    \qquad \text{where}\qquad f^* := \inf_{\theta\in\mathbb{R}^n} f(\theta)
    \]
    \item \textbf{Kurdyka–Łojasiewicz (KL) inequality.} Let \(f:\mathbb{R}^n\to\mathbb{R}\) be of class \(C^1\). For every critical point $\theta^*$ of $f$ on $\mathcal{S}_0$ there exist $\sigma\in[0,1)$ and $C_{KL}>0$ such that
    \[
    \|\nabla f(\theta)\| \;\geq\; C_{KL}\,[f(\theta)-f(\theta^*)]^\sigma.
    \]
\end{itemize}
The PL inequality is a gradient-dominance condition: on $\mathcal S_0$, the objective gap $f(\theta)-f^*$ is controlled by $\|\nabla f(\theta)\|^2$. Consequently, if $\bar\theta\in\mathcal S_0$ is a critical point, then $\nabla f(\bar\theta)=0$, and the PL inequality implies $f(\bar\theta)=f^*$. Hence every critical point in $\mathcal S_0$, and in particular every local minimizer in $\mathcal S_0$, is globally optimal on $\mathcal S_0$. Note, however, that PL alone does not ensure existence of a critical point or attainment of $f^*$; compact sublevel sets (Assumption~\ref{ass:nonconvexidad}) provide that. In contrast, the KL inequality is local around a given critical point: combined with a descent property and bounded sublevel sets, it guarantees convergence to some critical point, with a rate dictated by the exponent $\sigma$. As with PL, KL alone does not ensure that the limit is a minimizer; that conclusion requires convexity or PL in a neighborhood of the limit.

As in Section~\ref{Section:Convex_math}, before stating the main theorems we first introduce several preparatory lemmas. 


\begin{lemma}[Bounded memory and $L^2$-integrability of the gradient]
\label{lem:nonconvex_L2_gradient_memory}
Assume Assumption~\ref{ass:nonconvexidad}. Let $g(t):=\nabla f(\theta(t))$, and consider a global classical solution of
\begin{equation}
\label{eq:nonconvex_adagrad_rmsprop_shifted}
    \dot\theta(t)
    =
    -
    \frac{g(t)}
    {\sqrt{M_\nu[g^2](t+\alpha)}+\epsilon},
    \qquad t\geq0,
\end{equation}
where \(\alpha>0\), \(\epsilon>0\), and the memory variable satisfies
\begin{equation}
\label{eq:memory_ode_nonconvex}
    \dot M_\nu[g^2](t)=\lambda\|g(t)\|^2-\xi M_\nu[g^2](t),
    \qquad
    \lambda>0,
    \qquad
    \xi\in\{0,\lambda\}.
\end{equation}
Here \(\xi=0\) corresponds to the AdaGrad memory and \(\xi=\lambda\) to the RMSProp memory. Then
\[
    \theta(t)\in \mathcal{S}_0
    \qquad\text{for all }t\geq0,
\]
the gradient is bounded on the trajectory, and
\[
    \int_0^\infty \|g(t)\|^2\,dt<\infty.
\]
Moreover, \(M_\nu[g^2](t)\) is uniformly bounded on \([0,\infty)\). Consequently, there exists \(\mu_0>0\) such that
\[
    \mu(t)
    :=
    \frac{1}{\sqrt{M_\nu[g^2](t+\alpha)}+\epsilon}
    \geq \mu_0,
    \qquad t\geq0.
\]
\end{lemma}

\begin{proof}
Define $\Phi(t):=f(\theta(t))-f_*$. Along the flow and using equation~(\ref{eq:nonconvex_adagrad_rmsprop_shifted}), we have
\[
    \dot\Phi(t)
    =
    \langle g(t),\dot\theta(t)\rangle
    =
    -
    \frac{\|g(t)\|^2}{\sqrt{M_\nu[g^2](t+\alpha)}+\epsilon}
    \leq0.
\]
Hence \(f(\theta(t))\leq f(\theta(0))\) for all \(t\geq0\), and therefore
\[
    \theta(t)\in S_0.
\]
Since \(S_0\) is compact and \(f\in C^1\), there exists $G_0:=\sup_{\theta\in S_0}\|\nabla f(\theta)\|<\infty$ such that
\[
    \|g(t)\|\leq G_0,
    \qquad t\geq0.
\]
Moreover, integrating the identity for \(\dot\Phi\), we obtain
\begin{equation}
\label{eq:weighted_L2_nonconvex}
    \int_0^\infty
    \frac{\|g(t)\|^2}{\sqrt{M_\nu[g^2](t+\alpha)}+\epsilon}\,dt
    \leq
    \Phi(0)
    =
    f(\theta(0))-f_*.
\end{equation}

We now distinguish the two cases.

\medskip

\noindent\textbf{RMSProp case: \(\xi=\lambda\).}
By solving the differential equation~(\ref{eq:memory_ode_nonconvex}), we get
\[
    M_\nu[g^2](t)
    =
    e^{-\lambda t}M_\nu[g^2](0)
    +
    \lambda\int_0^t e^{-\lambda(t-s)}\|g(s)\|^2\,ds.
\]
Using \(\|g(s)\|\leq G_0\), we get
\[
    M_\nu[g^2](t)
    \leq
    e^{-\lambda t}M_\nu[g^2](0)
    +
    G_0^2(1-e^{-\lambda t})
    \leq
    \overline M_\nu,
\]
where
\[
    \overline M_\nu:=\max\{M_\nu[g^2](0),G_0^2\}.
\]
Thus
\[
    \sqrt{M_\nu[g^2](t+\alpha)}+\epsilon
    \leq
    \sqrt{\overline M_\nu}+\epsilon.
\]
From equation~(\ref{eq:weighted_L2_nonconvex}), it follows that
\[
    \int_0^\infty \|g(t)\|^2\,dt
    \leq
    \bigl(\sqrt{\overline M_\nu}+\epsilon\bigr)\Phi(0)
    <\infty.
\]
The boundedness of \(M_\nu[g^2]\) has already been proved.

\medskip

\noindent\textbf{AdaGrad case: \(\xi=0\).}
In this case,
\[
    \dot M_\nu[g^2](t)=\lambda\|g(t)\|^2,
\]
so \(M\) is nondecreasing. Moreover, since \(\|g(t)\|\leq G_0\),
\[
    M_\nu[g^2](t+\alpha)
    =
    M_\nu[g^2](t)+\lambda\int_t^{t+\alpha}\|g(s)\|^2\,ds
    \leq
    M_\nu[g^2](t)+\lambda\alpha G_0^2.
\]
Set
\[
    C_\alpha:=\lambda\alpha G_0^2.
\]
Then
\[
    \sqrt{M_\nu[g^2](t+\alpha)}+\epsilon
    \leq
    \sqrt{M_\nu[g^2](t)+C_\alpha}+\epsilon.
\]
Using this in equation~(\ref{eq:weighted_L2_nonconvex}), we obtain, for every \(T>0\),
\[
    \int_0^T
    \frac{\|g(t)\|^2}{\sqrt{M_\nu[g^2](t)+C_\alpha}+\epsilon}\,dt
    \leq
    \Phi(0).
\]
Since \(\dot M_\nu[g^2](t)=\lambda\|g(t)\|^2\), this becomes
\[
    \frac1\lambda
    \int_0^T
    \frac{\dot M_\nu[g^2](t)}
    {\sqrt{M_\nu[g^2](t)+C_\alpha}+\epsilon}\,dt
    \leq
    \Phi(0).
\]
Equivalently,
\[
    \int_{M(0)}^{M(T)}
    \frac{du}{\sqrt{u+C_\alpha}+\epsilon}
    \leq
    \lambda\Phi(0).
\]
Define
\[
    F_\alpha(r)
    :=
    \int_0^r
    \frac{du}{\sqrt{u+C_\alpha}+\epsilon}.
\]
An explicit primitive is
\[
    F_\alpha(r)
    =
    2\sqrt{r+C_\alpha}
    -
    2\epsilon\log\bigl(\sqrt{r+C_\alpha}+\epsilon\bigr)
    -
    F_\alpha(0),
\]
up to the harmless normalization \(F_\alpha(0)=0\). Since $F_\alpha(r)\to\infty$ as $r\to\infty$, the estimate
\[
    F_\alpha(M_\nu[g^2](T))-F_\alpha(M_\nu[g^2](0))
    \leq
    \lambda\Phi(0)
\]
implies that \(M_\nu[g^2](T)\) is uniformly bounded in \(T\). Hence there exists \(\overline M_A<\infty\) such that
\[
    M_\nu[g^2](t)\leq \overline M_A,
    \qquad t\geq0.
\]
Since $M_\nu[g^2](t)=M_\nu[g^2](0)+\lambda\int_0^t\|g(s)\|^2\,ds$, we conclude that
\[
    \int_0^\infty\|g(s)\|^2\,ds
    =
    \frac{1}{\lambda}\bigl(\lim_{t\to\infty}M_\nu[g^2](t)-M_\nu[g^2](0)\bigr)
    <\infty.
\]

Combining both cases, there exists \(\overline M<\infty\) such that
\[
    M_\nu[g^2](t)\leq \overline M,
    \qquad t\geq0.
\]
Therefore,
\[
    \sqrt{M_\nu[g^2](t+\alpha)}+\epsilon
    \leq
    \sqrt{\overline M}+\epsilon,
\]
and hence
\[
    \mu(t)
    =
    \frac{1}{\sqrt{M_\nu[g^2](t+\alpha)}+\epsilon}
    \geq
    \frac{1}{\sqrt{\overline M}+\epsilon}
    =:\mu_0>0.
\]
This completes the proof.
\end{proof}

\begin{corollary}[Vanishing gradient]
\label{cor:nonconvex_vanishing_gradient}
Under the assumptions of Lemma~\ref{lem:nonconvex_L2_gradient_memory}, one has
\[
    \lim_{t\to\infty}\|\nabla f(\theta(t))\|=0.
\]
Consequently, every accumulation point of \(\theta(t)\) belongs to the critical set of \(f\), namely,
\[
    \omega(\theta_0)\subseteq \operatorname{Crit}(f)\cap S_0.
\]
\end{corollary}

\begin{proof}
By Lemma~\ref{lem:nonconvex_L2_gradient_memory}, we know that
\[
    \theta(t)\in S_0,
    \qquad
    \|g(t)\|\leq G_0,
    \qquad
    \int_0^\infty \|g(t)\|^2\,dt<\infty.
\]
Moreover,
\[
    \dot\theta(t)
    =
    -
    \frac{g(t)}
    {\sqrt{M_\nu[g^2](t+\alpha)}+\epsilon}.
\]
Since \(\|g(t)\|\leq G_0\) and the denominator is bounded below by \(\epsilon>0\), we have
\[
    \|\dot\theta(t)\|
    \leq
    \frac{G_0}{\epsilon},
    \qquad t\geq0.
\]
Thus \(\theta(t)\) is uniformly continuous on \([0,\infty)\). Since \(S_0\) is compact and \(f\in C^1\), the map \(\nabla f\) is uniformly continuous on \(S_0\) \cite[Theorem 4.19]{Rudin1976}. Therefore $g(t)=\nabla f(\theta(t))$ is uniformly continuous, and hence \(\|g(t)\|^2\) is uniformly continuous. Since $\|g(t)\|^2\in L^1(0,\infty)$, Barbalat's lemma \cite{Barbalat1959} implies
\[
    \|g(t)\|^2\to0,
    \qquad t\to\infty.
\]
Therefore
\[
    \|g(t)\|\to0.
\]

Finally, since \(\theta(t)\in S_0\) and \(S_0\) is compact, the trajectory has accumulation points  \cite[Theorem 3.6(a)]{Rudin1976}. If \(\bar\theta\) belongs to this accumulation point set \(\omega(\theta_0)\) \cite[Section~6.3, p.~193]{Teschl2012}, then there exists \(t_k\to\infty\) such that
\[
    \theta(t_k)\to\bar\theta.
\]
By continuity of \(\nabla f\),
\[
    \nabla f(\bar\theta)
    =
    \lim_{k\to\infty}\nabla f(\theta(t_k))
    =
    0.
\]
Thus \(\bar\theta\in \operatorname{Crit}(f)\cap S_0\), which proves the claim.
\end{proof}

This corollary is worth emphasizing. It shows that, although no convergence to a single critical point is asserted at this stage, the trajectory becomes asymptotically stationary in the sense that
\[
    \|\nabla f(\theta(t))\|\to0
    \qquad\text{as }t\to\infty.
\]
Consequently, all possible limit points of the trajectory are critical points of $f$. More precisely, since the trajectory remains in the compact sublevel set \(S_0\), its \(\omega\)-limit set is nonempty and satisfies
\[
    \omega(\theta_0)\subseteq \operatorname{Crit}(f)\cap S_0.
\]
Thus, in the non-convex setting, the result guarantees convergence toward the critical set in the sense of accumulation points, but not necessarily convergence of the whole trajectory to a unique critical point.

\begin{theorem}[Nonconvex convergence under PL/KL for AdaGrad/RMSProp]
\label{thm:nonconvex_PL_KL_adagrad_rmsprop}
Assume Assumption~\ref{ass:nonconvexidad}. Let $g(t):=\nabla f(\theta(t))$, and consider a global classical solution of
\begin{equation}
\label{eq:nonconvex_adagrad_rmsprop_theorem4}
    \dot\theta(t)
    =
    -
    \frac{g(t)}
    {\sqrt{M_\nu[g^2](t+\alpha)}+\epsilon},
    \qquad t\geq0,
\end{equation}
where \(\alpha>0\), \(\epsilon>0\), and
\[
    \dot M_\nu[g^2](t)
    =
    \lambda\|g(t)\|^2-\xi M_\nu[g^2](t),
    \qquad
    \lambda>0,
    \qquad
    \xi\in\{0,\lambda\}.
\]
Here \(\xi=0\) corresponds to AdaGrad and \(\xi=\lambda\) to RMSProp. Define
\[
    \mu(t)
    :=
    \frac{1}
    {\sqrt{M_\nu[g^2](t+\alpha)}+\epsilon}.
\]
Then there exists \(\mu_0>0\) such that $\mu(t)\geq\mu_0$ for all $t\geq0$. Moreover,
\[
    \lim_{t\to\infty}\|g(t)\|=0,
    \qquad
    \omega(\theta_0)\subseteq \operatorname{Crit}(f)\cap S_0.
\]

\medskip

\noindent\textbf{(PL case).}
Assume that \(f\) satisfies the Polyak--{\L}ojasiewicz inequality on \(S_0\). Then
\[
    f(\theta(t))-f_*
    \leq
    \bigl(f(\theta(0))-f_*\bigr)
    e^{-\mu_0 C_{\mathrm{PL}}t}.
\]
Furthermore, \(\theta(t)\) has finite length and converges to some point
\[
    \theta^*\in\operatorname{argmin}_{S_0}f.
\]

\medskip

\noindent\textbf{(KL case).}
Assume that the \(\omega\)-limit set satisfies a uniform Kurdyka--{\L}ojasiewicz inequality. Then \(\theta(t)\) has finite length and converges to a single critical point
\[
    \theta^*\in\operatorname{Crit}(f)\cap S_0,
\]
with $f_\infty=f(\theta^*).$ Setting $\Phi_\infty(t):=f(\theta(t))-f(\theta^*)$, the following rates hold for \(t\geq T\):

\begin{enumerate}
    \item If \(\sigma=\frac12\), then
    \[
        \Phi_\infty(t)
        \leq
        \Phi_\infty(T)
        e^{-\mu_0 C_{\mathrm{KL}}^2(t-T)}.
    \]

    \item If \(\frac12<\sigma<1\), then
    \[
        \Phi_\infty(t)
        \leq
        \left[
            \Phi_\infty(T)^{1-2\sigma}
            +
            \mu_0 C_{\mathrm{KL}}^2(2\sigma-1)(t-T)
        \right]^{-\frac1{2\sigma-1}}.
    \]

    \item If \(0\leq\sigma<\frac12\), then \(\Phi_\infty(t)\) reaches zero in finite time. More precisely,
    \[
        \Phi_\infty(t)=0
    \]
    for every
    \[
        t
        \geq
        T+
        \frac{\Phi_\infty(T)^{1-2\sigma}}
        {\mu_0 C_{\mathrm{KL}}^2(1-2\sigma)}.
    \]
\end{enumerate}
\end{theorem}

\begin{proof}
By Lemma~\ref{lem:nonconvex_L2_gradient_memory}, the memory $M_\nu$ is uniformly bounded and there exists \(\mu_0>0\) such that $\mu(t)\geq\mu_0$ for all $t\geq0$. Moreover, the assumptions of Corollary~\ref{cor:nonconvex_vanishing_gradient} are satisfied. Hence
\[
    \lim_{t\to\infty}\|g(t)\|=0,
    \qquad
    \omega(\theta_0)\subseteq \operatorname{Crit}(f)\cap S_0.
\]

We first prove the PL statement. Define $\Phi(t):=f(\theta(t))-f_*$ and $f_*:=\inf_{\theta\in S_0}f(\theta)$. Along the flow, 
\[
    \dot\Phi(t)
    =
    \langle g(t),\dot\theta(t)\rangle
    =
    -\mu(t)\|g(t)\|^2.
\]
Define
\[
    \tau:=\inf\{t\geq0:\Phi(t)=0\}.
\]
If \(\tau=0\), then the trajectory starts at a global minimizer and the claim is immediate. Hence we assume \(\tau>0\). On the interval \([0,\tau)\), we have \(\Phi(t)>0\). Set
\[
    \varphi_{\mathrm{PL}}(s)
    :=
    \frac{2}{\sqrt{C_{\mathrm{PL}}}}\sqrt{s},
    \qquad s>0.
\]
Then, for \(t<\tau\), the PL inequality gives
\[
    \varphi_{\mathrm{PL}}'(\Phi(t))\|g(t)\|
    =
    \frac{1}{\sqrt{C_{\mathrm{PL}}\Phi(t)}}\|g(t)\|
    \geq1.
\]
Therefore, for \(t<\tau\),
\[
\begin{aligned}
    -\frac{\mathrm d}{\mathrm dt}
    \varphi_{\mathrm{PL}}(\Phi(t))
    &=
    \varphi_{\mathrm{PL}}'(\Phi(t))\mu(t)\|g(t)\|^2 \\
    &\geq
    \mu(t)\|g(t)\|
    =
    \|\dot\theta(t)\|.
\end{aligned}
\]
Consequently, for every \(T<\tau\),
\[
    \int_0^T \|\dot\theta(t)\|\,\mathrm dt
    \leq
    \varphi_{\mathrm{PL}}(\Phi(0))
    -
    \varphi_{\mathrm{PL}}(\Phi(T))
    \leq
    \varphi_{\mathrm{PL}}(\Phi(0)).
\]
Letting \(T\uparrow\tau\), we obtain
\[
    \int_0^\tau \|\dot\theta(t)\|\,\mathrm dt
    <\infty.
\]

If \(\tau=\infty\), this gives
\[
    \int_0^\infty \|\dot\theta(t)\|\,\mathrm dt<\infty.
\]
If $\tau<\infty$, then $\Phi(t)=0$ for all $t\ge\tau$, since $\Phi$ is nonincreasing and nonnegative. Hence $f(\theta(t))=f_*$ for all $t\ge\tau$. Moreover, since $\theta(t)\in S_0$ and $f_*=\inf_{\theta\in\mathbb R^n} f(\theta) =\min_{\theta\in S_0} f(\theta)$, it follows that
\[
    \theta(t)\in\operatorname{argmin}_{S_0} f
    \qquad \forall t\ge\tau .
\]
Since $f\in C^1$, $g(t)=\nabla f(\theta(t))=0$ for all $t\ge\tau$. Therefore $\dot\theta(t)=0$ for all $t\ge\tau$. Thus, in both cases,
\[
    \int_0^\infty \|\dot\theta(t)\|\,\mathrm dt<\infty .
\]
Consequently, the trajectory has finite length \cite[Theorem~6.27]{Rudin1976}.

We now show that the trajectory converges to a single point. For any $s>t\ge T$,
\[
    \|\theta(s)-\theta(t)\|
    \le
    \int_t^s \|\dot\theta(r)\|\,\mathrm dr
    \le
    \int_T^\infty \|\dot\theta(r)\|\,\mathrm dr .
\]
Since the last term tends to zero as $T\to\infty$, the trajectory $\theta(t)$ is
Cauchy as $t\to\infty$. Since $\mathbb R^n$ is complete, there exists
$\theta^*\in\mathbb R^n$ such that \cite[Theorem~3.11]{Rudin1976}
\[
    \theta(t)\to\theta^*
    \qquad \text{as } t\to\infty \,,
\]
Since $S_0$ is closed and $\theta(t)\in S_0$ for all $t\ge0$, we have $\theta^*\in S_0$. Therefore
\[
    \omega(\theta_0)=\{\theta^*\}.
\]
By Corollary~\ref{cor:nonconvex_vanishing_gradient}, every accumulation point of
the trajectory is critical. Hence
\[
    \theta^*\in\operatorname{Crit}(f)\cap S_0 .
\]

Finally, the PL inequality implies that every critical point in \(S_0\) has value \(f_*\). Indeed, if \(\nabla f(\theta^*)=0\), then
\[
    0
    =
    \|\nabla f(\theta^*)\|^2
    \geq
    C_{\mathrm{PL}}\bigl(f(\theta^*)-f_*\bigr),
\]
and therefore \(f(\theta^*)=f_*\) since, by definition of $f_*$, $f(\theta^*)\geq f_*$. Thus
\[
    \theta^*\in\operatorname{argmin}_{S_0}f.
\]

We now prove the KL statement. Since \(f(\theta(t))\) is nonincreasing and bounded below, the limit
\[
    f_\infty:=\lim_{t\to\infty}f(\theta(t))
\]
exists. Since the trajectory is contained in the compact set \(S_0\), the \(\omega\)-limit set is nonempty. By Corollary~\ref{cor:nonconvex_vanishing_gradient}, $\omega(\theta_0)\subseteq \operatorname{Crit}(f)\cap S_0.$ Moreover, \(f\) is constant on \(\omega(\theta_0)\), with value \(f_\infty\).

For \(t\geq T\), define
\[
    \Phi_\infty(t):=f(\theta(t))-f_\infty.
\]
The uniform KL inequality gives $\|g(t)\|\geq C_{\mathrm{KL}}\Phi_\infty(t)^\sigma$. Since $\dot\Phi_\infty(t) = -\mu(t)\|g(t)\|^2,$ we obtain
\[
    \dot\Phi_\infty(t)
    \leq
    -\kappa_{\mathrm{KL}}\Phi_\infty(t)^{2\sigma}.
\]
where $\kappa_{\mathrm{KL}}:=\mu_0 C_{\mathrm{KL}}^2$. We now distinguish three cases according to the value of \(2\sigma\). If \(\sigma=\frac12\), then the inequality is linear:
\[
    \dot\Phi_\infty(t)
    \leq
    -\kappa_{\mathrm{KL}}\Phi_\infty(t),
\]
and Gronwall's inequality gives the exponential decay
\[
    \Phi_\infty(t)
    \leq
    \Phi_\infty(T)e^{-\kappa_{\mathrm{KL}}(t-T)}.
\]

If \(\frac12<\sigma<1\), then \(2\sigma>1\). Dividing by
\(\Phi_\infty(t)^{2\sigma}\) and integrating from \(T\) to \(t\), one obtains
\[
    \Phi_\infty(t)
    \leq
    \left[
        \Phi_\infty(T)^{1-2\sigma}
        +
        \kappa_{\mathrm{KL}}(2\sigma-1)(t-T)
    \right]^{-\frac1{2\sigma-1}}.
\]
Thus the convergence of the objective gap is polynomial:
\[
    \Phi_\infty(t)
    =
    O\!\left(t^{-\frac1{2\sigma-1}}\right).
\]

Finally, if \(0\leq\sigma<\frac12\), then \(2\sigma<1\). In this case, integrating the same differential inequality shows that \(\Phi_\infty(t)\) reaches zero in finite time. More precisely, $\Phi_\infty(t)=0$ for every
\[
    t
    \geq
    T+
    \frac{\Phi_\infty(T)^{1-2\sigma}}
    {\kappa_{\mathrm{KL}}(1-2\sigma)}.
\]
This proves the three stated rates.

Finally, to prove convergence of the trajectory, introduce the function
\[
    \varphi(s)
    :=
    \frac{1}{C_{\mathrm{KL}}(1-\sigma)}s^{1-\sigma}.
\]
Then
\[
    \varphi'(s)
    =
    \frac{1}{C_{\mathrm{KL}}}s^{-\sigma}.
\]
Using the KL inequality,
\[
    \varphi'(\Phi_\infty(t))\|g(t)\|
    \geq1.
\]
Therefore,
\[
    -\frac{\mathrm d}{\mathrm dt}\varphi(\Phi_\infty(t))
    =
    \varphi'(\Phi_\infty(t))\mu(t)\|g(t)\|^2
    \geq
    \mu(t)\|g(t)\|
    =
    \|\dot\theta(t)\|.
\]
Thus
\[
    \int_T^\infty\|\dot\theta(t)\|\,\mathrm dt
    \leq
    \varphi(\Phi_\infty(T))<\infty.
\]
Hence
\[
    \int_0^\infty \|\dot\theta(t)\|\,\mathrm dt<\infty .
\]
By the finite-length argument established above, the trajectory converges to a single point $\theta^*\in S_0$, and $\omega(\theta_0)=\{\theta^*\}$. Using again Corollary~\ref{cor:nonconvex_vanishing_gradient}, every accumulation point of the trajectory is critical. Hence $\theta^*\in\operatorname{Crit}(f)\cap S_0$. Moreover, since $f(\theta(t))\to f_\infty$ and $\theta(t)\to\theta^*$, the continuity of $f$ gives
\[
    f_\infty=f(\theta^*) .
\]
Therefore
\[
    \Phi_\infty(t)=f(\theta(t))-f(\theta^*) ,
\]
which completes the proof.
\end{proof}
Under the PL inequality on the sublevel set \(S_0\), together with the uniform lower bound on the effective step size \(\mu(t)\geq \mu_0>0\), one obtains global exponential decay of the objective gap \(f(\theta(t))-f_*\), where \(f_*=\min_{\theta\in S_0}f(\theta)\). Moreover, since the trajectory has finite length, \(\theta(t)\) converges to a single point \(\theta^*\in S_0\). By Corollary~\ref{cor:nonconvex_vanishing_gradient}, every accumulation point is critical; hence \(\theta^*\in\operatorname{Crit}(f)\cap S_0\). The PL inequality then implies that every critical point in \(S_0\) has value \(f_*\), and consequently \(\theta^*\in\operatorname{argmin}_{S_0}f\).

By contrast, the KL inequality is local around the limiting critical set. Once the trajectory lies in a neighborhood where the uniform KL inequality holds, the convergence rates are dictated by the exponent \(\sigma\). More precisely, if \(\theta^*\) denotes the limiting critical point and
\[
    \Phi_\infty(t):=f(\theta(t))-f(\theta^*),
\]
then \(\Phi_\infty(t)\) decays exponentially when \(\sigma=\tfrac12\), polynomially when \(\tfrac12<\sigma<1\), and reaches zero in finite time when \(0\leq\sigma<\tfrac12\). In particular, KL alone guarantees convergence to a critical point, but it does not imply that this point is a minimizer; that conclusion requires an additional condition such as PL in a neighborhood of the limit.

\begin{theorem}[Nonconvex extended memory functional for AdaGrad/RMSProp]
\label{thm:nonconvex_extended_memory_adagrad_rmsprop}
Assume Assumption~\ref{ass:nonconvexidad}. Let $g(t):=\nabla f(\theta(t))$, and consider a global classical solution of
\begin{equation}
\label{eq:nonconvex_extended_memory_dyn}
    \dot\theta(t)
    =
    -
    \frac{g(t)}
    {\sqrt{M_\nu[g^2](t+\alpha)}+\epsilon},
    \qquad t\geq0,
\end{equation}
where \(\alpha>0\), \(\epsilon>0\), and the memory variable satisfies
\begin{equation}
\label{eq:nonconvex_extended_memory_ode}
    \dot M_\nu[g^2](t)
    =
    \lambda\|g(t)\|^2-\xi M_\nu[g^2](t),
    \qquad
    \lambda>0,
    \qquad
    \xi\in\{0,\lambda\}.
\end{equation}
Here \(\xi=0\) corresponds to AdaGrad and \(\xi=\lambda\) to RMSProp. Define
\[
    \mu(t)
    :=
    \frac{1}
    {\sqrt{M_\nu[g^2](t+\alpha)}+\epsilon}.
\]
Then there exists \(\mu_0>0\) such that $\mu(t)\geq\mu_0$ for all $t\geq0$. Let
\[
    f_*:=\min_{\theta\in S_0}f(\theta),
    \qquad
    \Phi_*(t):=f(\theta(t))-f_*,
\]
and define the extended memory functional
\[
    E_\rho^*(t)
    :=
    \Phi_*(t)+\rho M_\nu[g^2](t).
\]
Choose
\[
    \rho:=\frac{\mu_0}{2\lambda}.
\]
Then, for every \(t\geq0\),
\begin{equation}
\label{eq:nonconvex_Erho_dissipation}
    \dot E_\rho^*(t)
    \leq
    -
    \frac{\mu_0}{2}\|g(t)\|^2
    -
    \rho\xi M_\nu[g^2](t)
    \leq0.
\end{equation}

Consequently:

\begin{enumerate}
\item \textbf{AdaGrad case \((\xi=0)\).}
In this case,
\[
    \dot E_\rho^*(t)
    \leq
    -
    \frac{\mu_0}{2}\|g(t)\|^2
    \leq0.
\]
Thus \(E_\rho^*\) is nonincreasing. Moreover, $M_\nu[g^2](t)$ is nondecreasing and converges to a finite limit
\[
    M_\infty\geq0.
\]
In general, \(M_\infty\neq0\).

\item \textbf{RMSProp case \((\xi=\lambda)\).}
In this case,
\[
    \dot E_\rho^*(t)
    \leq
    -
    \frac{\mu_0}{2}\|g(t)\|^2
    -
    \rho\lambda M_\nu[g^2](t).
\]
Moreover,
\[
    M_\nu[g^2](t)\to0,
    \qquad t\to\infty.
\]

\item \textbf{PL refinement.}
If \(f\) satisfies the Polyak--{\L}ojasiewicz inequality on \(S_0\),
\[
    \|g(t)\|^2
    \geq
    C_{\mathrm{PL}}\Phi_*(t),
\]
then Theorem~\ref{thm:nonconvex_PL_KL_adagrad_rmsprop} gives
\[
    \Phi_*(t)
    \leq
    \Phi_*(0)e^{-\mu_0C_{\mathrm{PL}}t}.
\]
In addition, in the RMSProp case \((\xi=\lambda)\), the extended functional satisfies the closed exponential estimate
\[
    E_\rho^*(t)
    \leq
    E_\rho^*(0)e^{-\kappa_{\mathrm{PL}}t},
\]
where
\[
    \kappa_{\mathrm{PL}}
    :=
    \min\left\{
        \frac{\mu_0 C_{\mathrm{PL}}}{2},
        \lambda
    \right\}.
\]
Consequently,
\[
    \Phi_*(t)\leq E_\rho^*(0)e^{-\kappa_{\mathrm{PL}}t},
    \qquad
    M_\nu[g^2](t)
    \leq
    \frac{1}{\rho}E_\rho^*(0)e^{-\kappa_{\mathrm{PL}}t}.
\]
In the AdaGrad case \((\xi=0)\), the PL condition still gives exponential decay of \(\Phi_*(t)\), but \(E_\rho^*(t)\) generally converges to \(\rho M_\infty\), not to zero.

\item \textbf{KL refinement.}
If the uniform KL assumptions of Theorem~\ref{thm:nonconvex_PL_KL_adagrad_rmsprop} hold, then the trajectory converges to a single critical point
\[
    \theta^*\in\operatorname{Crit}(f)\cap S_0.
\]
Setting
\[
    \Phi_\infty(t):=f(\theta(t))-f(\theta^*),
\]
the objective gap \(\Phi_\infty(t)\) satisfies the KL rates stated in Theorem~\ref{thm:nonconvex_PL_KL_adagrad_rmsprop}. The memory behaves as follows:
\[
    M_\nu[g^2](t)\to M_\infty
    \quad\text{for AdaGrad},
    \qquad
    M_\nu[g^2](t)\to0
    \quad\text{for RMSProp}.
\]
\end{enumerate}
\end{theorem}

\begin{proof}
The dissipation estimate follows from the same computation as in Theorem~\ref{thm:extended_memory_adagrad_rmsprop}. Indeed, the derivation of the identity for \(\dot E_\rho^*(t)\) does not use convexity, but only the evolution equations
\[
    \dot\theta(t)=-\mu(t)g(t),
    \qquad
    \dot M_\nu[g^2](t)
    =
    \lambda\|g(t)\|^2-\xi M_\nu[g^2](t).
\]
By Lemma~\ref{lem:nonconvex_L2_gradient_memory}, the memory is uniformly bounded, and therefore there exists \(\mu_0>0\) such that $\mu(t)\geq \mu_0$ for all $t\geq0$. Defining
\[
    f_*:=\min_{\theta\in S_0}f(\theta),
    \qquad
    \Phi_*(t):=f(\theta(t))-f_*,
    \qquad
    E_\rho^*(t):=\Phi_*(t)+\rho M_\nu[g^2](t),
\]
and choosing $\rho:=\frac{\mu_0}{2\lambda}$, the same calculation as in Theorem~\ref{thm:extended_memory_adagrad_rmsprop} yields
\[
    \dot E_\rho^*(t)
    \leq
    -
    \frac{\mu_0}{2}\|g(t)\|^2
    -
    \rho\xi M_\nu[g^2](t)
    \leq0.
\]
This proves the basic dissipation estimate.

We now distinguish the two memory regimes. If \(\xi=0\), corresponding to AdaGrad, then
\[
    \dot M_\nu[g^2](t)=\lambda\|g(t)\|^2\geq0.
\]
Hence \(M_\nu[g^2](t)\) is nondecreasing. Since it is uniformly bounded by Lemma~\ref{lem:nonconvex_L2_gradient_memory}, it converges to a finite limit
\[
    M_\infty:=\lim_{t\to\infty}M_\nu[g^2](t).
\]
In general \(M_\infty\neq0\), reflecting the cumulative nature of AdaGrad.

If \(\xi=\lambda\), corresponding to RMSProp, then
\[
    \dot M_\nu[g^2](t)+\lambda M_\nu[g^2](t)
    =
    \lambda\|g(t)\|^2.
\]
By Corollary~\ref{cor:nonconvex_vanishing_gradient}, \(\|g(t)\|\to0\). Since the above equation is a stable linear equation forced by a term vanishing at infinity, it follows that
\[
    M_\nu[g^2](t)\to0.
\]

Assume now that the PL inequality holds on \(S_0\). The exponential decay of \(\Phi_*(t)\) follows directly from Theorem~\ref{thm:nonconvex_PL_KL_adagrad_rmsprop}. In the RMSProp case, using \(\xi=\lambda\), the dissipation estimate gives
\[
    \dot E_\rho^*(t)
    \leq
    -
    \frac{\mu_0 C_{\mathrm{PL}}}{2}\Phi_*(t)
    -
    \rho\lambda M_\nu[g^2](t).
\]
Therefore
\[
    \dot E_\rho^*(t)
    \leq
    -
    \kappa_{\mathrm{PL}}
    \bigl(
        \Phi_*(t)+\rho M_\nu[g^2](t)
    \bigr)
    =
    -
    \kappa_{\mathrm{PL}}E_\rho^*(t),
\]
with
\[
    \kappa_{\mathrm{PL}}
    :=
    \min\left\{
        \frac{\mu_0 C_{\mathrm{PL}}}{2},
        \lambda
    \right\}.
\]
By Gronwall's inequality,
\[
    E_\rho^*(t)
    \leq
    E_\rho^*(0)e^{-\kappa_{\mathrm{PL}}t}.
\]
Since both summands of \(E_\rho^*\) are nonnegative, this also yields
\[
    \Phi_*(t)
    \leq
    E_\rho^*(0)e^{-\kappa_{\mathrm{PL}}t},
    \qquad
    M_\nu[g^2](t)
    \leq
    \frac1\rho E_\rho^*(0)e^{-\kappa_{\mathrm{PL}}t}.
\]

Finally, assume that the uniform KL assumptions of Theorem~\ref{thm:nonconvex_PL_KL_adagrad_rmsprop} hold. Then Theorem~\ref{thm:nonconvex_PL_KL_adagrad_rmsprop} implies that
\(\theta(t)\) converges to a single critical point $\theta^*\in\operatorname{Crit}(f)\cap S_0$ and, setting $\Phi_\infty(t):=f(\theta(t))-f(\theta^*)$, one has the KL rates stated there for \(\Phi_\infty(t)\). 

Define now
\[
    E_\rho^\infty(t)
    :=
    \Phi_\infty(t)+\rho M_\nu[g^2](t).
\]
Since \(f(\theta^*)\) is constant, the derivative of \(\Phi_\infty\) is the same as the derivative of the objective along the flow. Hence, exactly as before,
\[
\begin{aligned}
    \dot E_\rho^\infty(t)
    &=
    -\bigl(\mu(t)-\rho\lambda\bigr)\|g(t)\|^2
    -
    \rho\xi M_\nu[g^2](t)\\
    &\leq
    -
    \frac{\mu_0}{2}\|g(t)\|^2
    -
    \rho\xi M_\nu[g^2](t).
\end{aligned}
\]
Using the KL inequality, $\|g(t)\|\geq C_{\mathrm{KL}}\Phi_\infty(t)^\sigma$, we obtain
\begin{equation}
\label{eq:KL_energy_partial_dissipation}
    \dot E_\rho^\infty(t)
    \leq
    -
    \frac{\mu_0}{2}C_{\mathrm{KL}}^2
    \Phi_\infty(t)^{2\sigma}
    -
    \rho\xi M_\nu[g^2](t).
\end{equation}
This estimate shows that the objective part of the energy dissipates according to the KL geometry, while the memory part dissipates only when \(\xi>0\), i.e. in the RMSProp case.

In the AdaGrad case, \(\xi=0\), and therefore the memory term is not directly dissipated:
\[
    \dot E_\rho^\infty(t)
    \leq
    -
    \frac{\mu_0}{2}C_{\mathrm{KL}}^2
    \Phi_\infty(t)^{2\sigma}.
\]
Thus the KL rates apply to \(\Phi_\infty(t)\), but not to the full functional
\(E_\rho^\infty(t)\), since \(M_\nu[g^2](t)\) may converge to a nonzero finite limit.

In the RMSProp case, \(\xi=\lambda\), and equation~\ref{eq:KL_energy_partial_dissipation} becomes
\[
    \dot E_\rho^\infty(t)
    \leq
    -
    \frac{\mu_0}{2}C_{\mathrm{KL}}^2
    \Phi_\infty(t)^{2\sigma}
    -
    \rho\lambda M_\nu[g^2](t).
\]
Therefore the memory is genuinely dissipated. Moreover, since \(E_\rho^\infty=\Phi_\infty+\rho M_\nu[g^2]\), the full energy also converges to zero in the RMSProp case. However, one should not write directly
\[
    \dot E_\rho^\infty(t)
    \leq
    -c\bigl(E_\rho^\infty(t)\bigr)^{2\sigma}
\]
without an additional comparison argument, because \(E_\rho^\infty\geq\Phi_\infty\), and the KL inequality controls only the objective gap, not the memory term.

Consequently, the KL rates are assigned directly to the objective gap
\[
    \Phi_\infty(t)=f(\theta(t))-f(\theta^*),
\]
while the memory component is handled separately: it converges to a finite, possibly nonzero, limit in AdaGrad, and it converges to zero in RMSProp. This completes the proof.
\end{proof}

Theorem \ref{thm:nonconvex_PL_KL_adagrad_rmsprop} and \ref{thm:nonconvex_extended_memory_adagrad_rmsprop} rely on the same memory viewpoint but they emphasize different aspects of the dynamics through the extended functional
\[
    E_\rho^*(t)
    =
    \Phi_*(t)+\rho M_\nu[g^2](t),
    \qquad
    \Phi_*(t):=f(\theta(t))-f_*,
    \qquad
    f_*:=\min_{\theta\in S_0}f(\theta).
\]
Under the PL inequality on \(S_0\), together with the uniform lower bound on the effective step size $\mu(t)\geq \mu_0>0$, one obtains exponential decay of the objective gap. If, in addition, the memory has forgetting, namely \(\xi>0\), then the extended functional also satisfies a closed exponential estimate. More precisely, in the RMSProp case \(\xi=\lambda\), one obtains
\[
    E_\rho^*(t)\leq E_\rho^*(0)e^{-\kappa_{\mathrm{PL}}t},
    \qquad
    \kappa_{\mathrm{PL}}
    =
    \min\left\{
        \frac{\mu_0 C_{\mathrm{PL}}}{2},
        \xi
    \right\}.
\]
Consequently,
\[
    \Phi_*(t)\leq E_\rho^*(0)e^{-\kappa_{\mathrm{PL}}t},
    \qquad
    M_\nu[g^2](t)
    \leq
    \frac{1}{\rho}E_\rho^*(0)e^{-\kappa_{\mathrm{PL}}t}.
\]
Thus, in the forgetting case, both the optimization error and the memory term decay exponentially. Moreover, since the trajectory has finite length, \(\theta(t)\) converges to a single point \(\theta^*\in\operatorname{argmin}_{S_0}f\); in particular, if the minimizer in \(S_0\) is unique, then the whole trajectory converges to that minimizer. As in the convex case, the memory effect appears explicitly in the decay constant: stronger accumulated memory corresponds to a smaller lower bound \(\mu_0\), and therefore slows down the dynamics.

Under the KL condition, which is local around the limiting critical set, the standard KL rates are recovered for the objective gap
\[
    \Phi_\infty(t):=f(\theta(t))-f(\theta^*),
\]
where \(\theta^*\in\operatorname{Crit}(f)\cap S_0\) is the critical point to which the trajectory converges. Namely, \(\Phi_\infty(t)\) decays exponentially when \(\sigma=\tfrac12\), polynomially like
\[
    t^{-\frac{1}{2\sigma-1}}
\]
when \(\tfrac12<\sigma<1\), and reaches zero in finite time when \(0\leq\sigma<\tfrac12\). This finite-time extinction concerns the objective gap relative to the limiting critical value \(f(\theta^*)\). It does not arise in the strongly convex theorem, where the convergence rate is exponential.

For the case \(\xi=0\), corresponding to AdaGrad, one still has
\[
    \dot E_\rho^*(t)\leq0,
\]
but the inequality cannot be closed in the form
\[
    \dot E_\rho^*(t)\leq -\kappa E_\rho^*(t)
\]
for some \(\kappa>0\). This reflects the cumulative nature of AdaGrad: the memory has no forgetting mechanism and may converge to a nonzero finite limit. Nevertheless, if \(\mu(t)\geq\mu_0>0\) and the PL inequality holds on \(S_0\), then
\[
    \Phi_*(t)\lesssim e^{-\mu_0 C_{\mathrm{PL}}t},
\]
and the finite-length argument yields convergence of \(\theta(t)\) to a single point in \(\operatorname{argmin}_{S_0}f\). If this minimizer is unique, then \(\theta(t)\to\theta^*\).

Let us finally point out an observation that is easy to overlook. In the strongly convex case, the objective gap directly controls the distance to the minimizer through a global error bound, which allows one to obtain rates for \(\|\theta(t)-\theta^*\|\) directly from the decay of \(\Phi(t)\). In the nonconvex case, such a conclusion is not automatic. One can obtain rates for the objective gap under PL or KL assumptions, and convergence of the trajectory through finite length, but a quantitative rate for \(\|\theta(t)-\theta^*\|\) would require an additional error-bound type estimate relating the distance to the limiting critical point with the objective gap. Such an estimate is not guaranteed for a general nonconvex function.

We finally consider the nonconvex Adam dynamics. In this setting, the correct analogue of the strongly convex result is not convergence to a prescribed minimizer, but rather convergence toward the critical set. Indeed, without convexity or a global gradient-dominance condition, one cannot exclude convergence to non-minimizing critical points.

For this reason, we formulate the result in terms of a compact positively invariant set for the augmented dynamics and prove asymptotic stationarity. Additional assumptions, such as finiteness of the critical set or a Polyak--Łojasiewicz condition, are then used to upgrade this stationarity result to convergence to a single critical point or to a minimizer.

\begin{theorem}[Nonconvex Adam flow: Lyapunov dissipation and criticality]
\label{thm:nonconvex_adam_shifted}
Let \(f:\mathbb R^n\to\mathbb R\) be of class \(C^1\) and bounded from below. Let $g(t):=\nabla f(\theta(t))$, and consider a global classical solution of the continuous Adam dynamics
\[
    \dot\theta(t)
    =
    -
    \widehat a(t)\widehat m(t),
    \qquad t\geq0,
\]
where $\widehat m(t):=m(t+\alpha)$ and $\widehat v(t):=v(t+\alpha)$ and
\[
    \widehat a(t)
    :=
    \frac{\eta(t+\alpha)}
    {\sqrt{\widehat v(t)}+\varepsilon(t+\alpha)}.
\]
Assume that the shifted moments satisfy
\[
    \dot{\widehat m}(t)+\lambda_1\widehat m(t)
    =
    \lambda_1 g(t+\alpha),
    \qquad
    \dot{\widehat v}(t)+\lambda_2\widehat v(t)
    =
    \lambda_2\|g(t+\alpha)\|^2,
\]
where
\[
    \lambda_1:=\frac{1-\beta_1}{\alpha},
    \qquad
    \lambda_2:=\frac{1-\beta_2}{\alpha}.
\]
Assume also the scalar Lyapunov condition
\[
    \mathcal H_{\beta_1,\beta_2}<2(1-\beta_1),
\]
and set
\[
    c_\beta
    :=
    1-
    \frac{\mathcal H_{\beta_1,\beta_2}}
    {2(1-\beta_1)}
    >0.
\]
Finally, assume that the shift defect satisfies
\[
    \bigl\langle g(t)-g(t+\alpha),\widehat m(t)\bigr\rangle
    \geq
    -\sigma_\alpha\|\widehat m(t)\|^2,
    \qquad t\geq0,
\]
for some \(0\leq\sigma_\alpha<c_\beta\).

Let
\[
    f_{\inf}:=\inf_{\theta\in\mathbb R^n}f(\theta),
    \qquad
    \Phi(t):=f(\theta(t))-f_{\inf},
\]
and define
\[
    V(t)
    :=
    \Phi(t)
    +
    \frac{\widehat a(t)}{2\lambda_1}
    \|\widehat m(t)\|^2.
\]
Assume that the augmented sublevel set
\[
    \mathcal K_0
    :=
    \left\{
        \theta\in\mathbb R^n:
        f(\theta)\leq f_{\inf}+V(0)
    \right\}
\]
is compact. Then
\[
    \dot V(t)
    \leq
    -
    \widehat a(t)
    \bigl(c_\beta-\sigma_\alpha\bigr)
    \|\widehat m(t)\|^2
    \leq0.
\]
Consequently,
\[
    \theta(t)\in\mathcal K_0,
    \qquad t\geq0,
\]
and
\[
    \widehat m(t)\to0,
    \qquad
    \dot\theta(t)\to0,
    \qquad
    \nabla f(\theta(t))\to0,
    \qquad
    \widehat v(t)\to0.
\]
In particular, the \(\omega\)-limit set is nonempty, compact, and satisfies
\[
    \omega(\theta_0)
    \subseteq
    \operatorname{Crit}(f)\cap\mathcal K_0.
\]
Moreover, \(f(\theta(t))\) admits a finite limit as \(t\to\infty\).

If \(\operatorname{Crit}(f)\cap\mathcal K_0\) is finite, then \(\theta(t)\) converges to a single critical point
\[
    \theta^*\in\operatorname{Crit}(f)\cap\mathcal K_0.
\]
Finally, if \(f\) satisfies the Polyak--{\L}ojasiewicz inequality on \(\mathcal K_0\), then every accumulation point of \(\theta(t)\) is a minimizer of \(f\) on \(\mathcal K_0\). In particular, if this minimizer is unique, then
\[
    \theta(t)\to\theta^*,
    \qquad
    \theta^*\in\operatorname{argmin}_{\mathcal K_0}f.
\]
\end{theorem}

\begin{proof}
The Lyapunov computation is exactly the same as in the proof of Theorem~\ref{thm:adam_shifted}. Indeed, that computation only uses the Adam dynamics, the shifted first-moment equation, the estimate on
\(\frac{d}{dt}\log\widehat a(t)\), and the one-sided shift-defect condition; it does not use convexity. Hence, under the present assumptions, we obtain
\[
    \dot V(t)
    \leq
    -
    \widehat a(t)
    \bigl(c_\beta-\sigma_\alpha\bigr)
    \|\widehat m(t)\|^2
    \leq0.
\]

Since \(V(t)\leq V(0)\) and $\Phi(t)=f(\theta(t))-f_{\inf}\leq V(t)$, we have
\[
    f(\theta(t))\leq f_{\inf}+V(0),
    \qquad t\geq0.
\]
Therefore \(\theta(t)\in\mathcal K_0\) for all \(t\geq0\). By assumption, \(\mathcal K_0\) is compact, and since \(f\in C^1\), the gradient is bounded and uniformly continuous on \(\mathcal K_0\) \cite[Theorem 4.14 and Theorem 4.19]{Rudin1976}.

Moreover, exactly as in the proof of Theorem~\ref{thm:adam_shifted}, the boundedness of \(g(t)\), together with the shifted second-moment equation, implies that \(\widehat v(t)\) is uniformly bounded. Since \(t+\alpha\geq\alpha\),
the bias-correction factors stay away from the singular regime. Hence there exist constants $0<a_*\leq a^*<\infty$ such that
\[
    a_*\leq \widehat a(t)\leq a^*,
    \qquad t\geq0.
\]

Integrating the dissipation inequality gives
\[
    \int_0^\infty \|\widehat m(t)\|^2\,dt<\infty.
\]
The same compactness and uniform-continuity argument used in Corollary~\ref{cor:nonconvex_vanishing_gradient} then yields
\[
    \widehat m(t)\to0.
\]
Since $\dot\theta(t)=-\widehat a(t)\widehat m(t)$, and \(\widehat a(t)\) is bounded, we also get
\[
    \dot\theta(t)\to0.
\]

Recall that
\[
    \dot{\widehat m}(t)
    =
    \lambda_1\bigl(g(t+\alpha)-\widehat m(t)\bigr).
\]
Since the trajectory remains in the compact set \(\mathcal K_0\) and \(f\in C^1\), the map \(\nabla f\) is uniformly continuous on
\(\mathcal K_0\) \cite[Theorem 4.19]{Rudin1976}. Moreover, \(\dot\theta(t)=-\widehat a(t)\widehat m(t)\) is bounded, because both \(\widehat a(t)\) and \(\widehat m(t)\) are bounded. Hence \(\theta(t)\) is uniformly continuous on \([0,\infty)\) \cite[Theorem 5.10]{Rudin1976}. It follows that \(g(t+\alpha)=\nabla f(\theta(t+\alpha))\) is uniformly continuous.

On the other hand, \(\widehat m(t)\) is uniformly continuous, since
\[
    \dot{\widehat m}(t)
    =
    \lambda_1\bigl(g(t+\alpha)-\widehat m(t)\bigr)
\]
is bounded. Consequently, \(\dot{\widehat m}(t)\), being the difference of two uniformly continuous functions multiplied by the constant \(\lambda_1\), is uniformly continuous.

Since \(\widehat m(t)\to0\), the function \(\widehat m(t)\) has a finite limit. Therefore, by the Barbalat-type criterion stating that if a differentiable function has a finite limit and its derivative is uniformly continuous, then its derivative converges to zero, we obtain
\[
    \dot{\widehat m}(t)\to0.
\]

From the shifted first-moment equation,
\[
    g(t+\alpha)
    =
    \widehat m(t)
    +
    \frac1{\lambda_1}\dot{\widehat m}(t),
\]
we get $g(t+\alpha)\to0$, and therefore
\[
    g(t)=\nabla f(\theta(t))\to0.
\]

The shifted second-moment equation $\dot{\widehat v}(t)+\lambda_2\widehat v(t) = \lambda_2\|g(t+\alpha)\|^2$ then gives
\[
    \widehat v(t)\to0.
\]

We now identify the possible limit points of the trajectory. Since $\theta(t)$ remains in the compact set $\mathcal K_0$, the $\omega$-limit set $\omega(\theta_0)$ is nonempty, compact, and connected \cite[Lemma~6.6]{Teschl2012}. Let $\bar\theta\in\omega(\theta_0)$. By the definition of the $\omega$-limit set \cite[Section~6.3, p.~193]{Teschl2012}, there exists a sequence $t_k\to\infty$ such that
\[
    \theta(t_k)\to\bar\theta.
\]
Since $g(t)=\nabla f(\theta(t))\to0$, we have
\[
    \nabla f(\theta(t_k))=g(t_k)\to0.
\]
Passing to the limit and using the continuity of $\nabla f$, we obtain
\[
    \nabla f(\bar\theta)
    =
    \lim_{k\to\infty}\nabla f(\theta(t_k))
    =
    0.
\]
Therefore $\bar\theta\in\operatorname{Crit}(f)$. Since $\bar\theta$ was arbitrary,
\[
    \omega(\theta_0)
    \subseteq
    \operatorname{Crit}(f)\cap\mathcal K_0.
\]
Next, since $V(t)$ is nonincreasing and bounded below, it has a finite limit. Moreover, since $\widehat m(t)\to0$ and $\widehat a(t)$ is bounded, the memory term in $V(t)$ vanishes. Therefore $f(\theta(t))$ also admits a finite limit.

We now derive two complementary consequences of the inclusion
\[
    \omega(\theta_0)
    \subseteq
    \operatorname{Crit}(f)\cap\mathcal K_0.
\]

First, assume that $\operatorname{Crit}(f)\cap\mathcal K_0$ is finite. Then $\omega(\theta_0)$ is contained in a finite set. On the other hand, $\omega(\theta_0)$ is connected \cite[Lemma~6.6]{Teschl2012}. Hence $\omega(\theta_0)$ must be a singleton: indeed, a connected subset of a finite set consists of a single point \cite[Definition~2.45, p.~42]{Rudin1976}. Therefore, there exists
\[
    \theta^*\in\operatorname{Crit}(f)\cap\mathcal K_0
\]
such that
\[
    \omega(\theta_0)=\{\theta^*\}.
\]
We now show that this implies convergence of the whole trajectory. Since the trajectory is contained in the compact set $\mathcal K_0$, every sequence $\{\theta(t_k)\}$ with $t_k\to\infty$ admits a convergent subsequence \cite[Theorem~3.6(a)]{Rudin1976}. If $\theta(t)$ did not converge to $\theta^*$, then there would exist $\varepsilon>0$ and a sequence
$t_k\to\infty$ such that
\[
    \|\theta(t_k)-\theta^*\|\ge\varepsilon .
\]
Extracting a convergent subsequence, still denoted by $\theta(t_k)$, we obtain
\[
    \theta(t_k)\to\bar\theta
\]
for some $\bar\theta\in\omega(\theta_0)$. Since $\omega(\theta_0)=\{\theta^*\}$, we must have $\bar\theta=\theta^*$, which contradicts the inequality above. Therefore
\[
    \theta(t)\to\theta^*
    \qquad \text{as } t\to\infty .
\]

Second, assume that $f$ satisfies the PL inequality on $\mathcal K_0$, namely, there exists $C_{\mathrm{PL}}>0$ such that
\[
    \|\nabla f(\theta)\|^2
    \ge
    C_{\mathrm{PL}}\bigl(f(\theta)-f_*\bigr),
    \qquad \theta\in\mathcal K_0,
\]
where
\[
    f_*:=\min_{\theta\in\mathcal K_0} f(\theta).
\]
The minimum exists because $\mathcal K_0$ is compact and $f$ is continuous. Let $\bar\theta\in\omega(\theta_0)$ be arbitrary. By the inclusion proved above,
\[
    \bar\theta\in\operatorname{Crit}(f)\cap\mathcal K_0.
\]
Hence $\nabla f(\bar\theta)=0$. Applying the PL inequality at $\bar\theta$ gives
\[
    0
    =
    \|\nabla f(\bar\theta)\|^2
    \ge
    C_{\mathrm{PL}}\bigl(f(\bar\theta)-f_*\bigr).
\]
Since $C_{\mathrm{PL}}>0$, this implies
\[
    f(\bar\theta)\le f_*.
\]
On the other hand, since $\bar\theta\in\mathcal K_0$ and $f_*$ is the minimum of
$f$ on $\mathcal K_0$, we also have
\[
    f(\bar\theta)\ge f_*.
\]
Therefore
\[
    f(\bar\theta)=f_*.
\]
Thus every accumulation point of the trajectory is a minimizer of $f$ on
$\mathcal K_0$, namely
\[
    \omega(\theta_0)\subseteq \operatorname{argmin}_{\mathcal K_0} f.
\]

Consequently, if the minimizer of $f$ on $\mathcal K_0$ is unique, say
\[
    \operatorname{argmin}_{\mathcal K_0} f=\{\theta^*\},
\]
then
\[
    \omega(\theta_0)=\{\theta^*\}.
\]
By the same compactness argument used in the first case, this implies
\[
    \theta(t)\to\theta^*
    \qquad \text{as } t\to\infty .
\]
This completes the proof.
\end{proof}

In contrast with the extended memory functional \(E_\rho=\Phi+\rho M_\nu[g^2]\), used for the AdaGrad/RMSProp dynamics, the Adam flow is controlled, under the scalar Lyapunov condition and the one-sided shift-defect estimate, by the Lyapunov functional
\[
V(t)=\Phi(t)+\frac{\widehat a(t)}{2\lambda_1}\|\widehat m(t)\|^2,
\]
where \(\widehat m(t)=m(t+\alpha)\) and \(\widehat a(t)\) denotes the shifted Adam gain. The key point is that the Lyapunov computation is the same as in Theorem \ref{thm:adam_shifted}: it only requires an upper bound on $h(t):=\frac{d}{dt}\log \widehat a(t)$
together with the one-sided shift-defect estimate. Under these hypotheses, one obtains
\[
\dot V(t)\le -\widehat a(t)(c_\beta-\sigma_\alpha)\|\widehat m(t)\|^2\le0.
\]
Therefore \(V\) is nonincreasing and
\[
\int_0^\infty \widehat a(t)\|\widehat m(t)\|^2\,dt<\infty.
\]
Together with the uniform continuity of the shifted moment, Barbalat's lemma yields \(\widehat m(t)\to0\). The shifted first-moment equation then implies \(\nabla f(\theta(t))\to0\), and consequently every accumulation point of the trajectory is critical:
\[
\omega(\theta_0)\subset \operatorname{Crit}(f)\cap K_0.
\]
Thus, in the nonconvex Adam case, the theorem establishes asymptotic stationarity. Convergence to a single point requires an additional structural assumption, such as finiteness of \(\operatorname{Crit}(f)\cap K_0\), while convergence to a minimizer requires a condition such as the PL inequality, together with uniqueness if one wants convergence of the whole trajectory to a prescribed minimizer.

\section{Numerical Simulations}

In this section, we present numerical simulations\footnote{All simulations can be reproduced from the repository available at \href{https://github.com/carlosherediapimienta/numerical_simulations_nonlocal}{GitHub}.} that validate the approximations proposed in the preceding sections. As shown, the simulations closely mirror the behavior of the discrete algorithms, confirming that the first-order integro-differential equations serve as accurate and reliable models for analyzing the dynamics of these optimization methods, both for convex and for nonconvex functions. 

Before presenting the results, we provide an overview of the numerical method employed to solve the first-order integro-differential equations.

\subsection{Numerical Method}

To solve these equations, we employed the IDESolver method as described in \cite{GELMI2014}, which we implemented from scratch, tailoring it to our specific requirements rather than relying on the existing Python module. See the pseudocode in Algorithm~\ref{alg:IDESolver}. Moreover, this implementation was carried out within the JAX framework in order to fully exploit GPU-based numerical computation. The GPU used was an NVIDIA RTX 5070ti.

The numerical method operates as follows: It generates an initial guess for \( y(x) \) by disregarding the integral term (nonlocal part) in the equation and solving the resulting ordinary differential equation. This initial guess is then used to solve the original equation, and the new solution is compared to the initial guess. The similarity between the two is quantified by an error, defined as the sum of squared differences at the discretization points:
\[
\text{error} = \sum_{n=1}^{M} \left( y_{\text{current},n} - y_{\text{guess},n} \right)^2,
\]
where \( M \) denotes the total number of discretization points,  \( y_{\text{current},n} \) represents the current estimate of \( y \),  and \( y_{\text{guess},n} \) is the guess, both evaluated at the \( n \)-th step.

If the error exceeds a predefined tolerance, a new guess \( y_{\text{new}} \) is generated by a linear combination of the current \( y_{\text{current}} \) and the old guess $y_{\text{guess}}$:
\[
y_{\text{new}} = a y_{\text{current}} + (1 - a) y_{\text{guess}}.
\]
In our implementation, the value of \( a \) is adaptive, varying based on the error between iterations, and ranges from 0.5 to 0.9999. If the error increases between iterations, \( a \) is incremented by 0.0005. This adaptation allows for finer control of convergence. We set a global error tolerance of \( 1 \times 10^{-4} \) and a maximum number of iterations of \( 1 \times 10^{10} \).
    
For the integral calculations, we employed a Gaussian quadrature method adapted for JAX with $n=1000$, instead of the \texttt{quad} method used in the original paper. This leads to a significant increase in calculation speed. 

Additionally, we used the Euler method to solve the differential equation, rather than the RK45 method commonly employed in the original approach, to ensure consistency with Assumption \ref{Ass:Assumption}. The (future) time increment term \(\alpha\) of Propositions \ref{prop:AdaGrad}, \ref{prop:RMSProp} and \ref{prop:Adam} is obtained using the \texttt{interp1d} interpolation function from the Interpax library.

\begin{algorithm}[h!]
    \caption{Iterative Modified IDESolver Method}
    \label{alg:IDESolver}
    \begin{algorithmic}[1]
    \State Initialize the iteration counter $k \gets 0$
    \State Compute the initial solution $y_{\text{current}}$ using the original differential equation
    \State Compute the initial guess $y_{\text{guess}}$ including the integral part with $y_{\text{current}}$
    \State Calculate the initial global error $\text{error} \gets \| y_{\text{current}} - y_{\text{guess}} \|$
    \While{$\text{error} > \text{tolerance}$}
        \State Compute new solution $y_{\text{new}}$ using a smoothing factor with $y_{\text{current}}$ and $y_{\text{guess}}$
        \State Update $y_{\text{guess}}$ solving the ODE including the integral part with $y_{\text{new}}$
        \State Calculate the current global error $\text{new\_error} \gets \| y_{\text{new}} - y_{\text{guess}} \|$
        \If{$\text{new\_error} > \text{error}$}
            \If{maximum smoothing factor reached}
                \State Exit the loop without achieving the desired tolerance
            \Else
                \State Update the smoothing factor to the next value
            \EndIf
        \EndIf
        \State Update $y_{\text{current}} \gets y_{\text{new}}$
        \State Increment the iteration counter $k \gets k + 1$
        \If{$k > k_{\text{max}}$}
            \State Exit the loop
        \EndIf
        \State Update $\text{error} \gets \text{new\_error}$
    \EndWhile
    \State Set the final solution $y \gets y_{\text{guess}}$
    \State \Return time values and the corresponding solution $y$
    \end{algorithmic}
    \end{algorithm}

This process continues until the error falls below the predefined tolerance. For further details on the implementation, please refer to the \href{https://github.com/carlosherediapimienta/numerical_simulations_nonlocal}{GitHub repository}.

\subsection{Numerical Results}\label{subsec:Numerical}
In this section, we present the numerical results obtained using the numerical method described previously and compare them with the outcomes derived from the original algorithms. All the figures are provided in Appendix~\ref{App:SF}.

It is important to clarify that throughout this section, when we refer to ``iterations," we specifically mean the value of \( k \) as defined by the relation: \( k = t/\alpha \). This should not be confused with the iterations of the numerical method used to compute the solution. The latter refers to the internal steps required to obtain the overall solution, rather than the temporal evolution of \(\theta\). To prevent confusion, we will use the term ``k-iterations."

\subsubsection{Convex Function Simulations}

We now begin with the simulations for the convex function $f(\theta) = (\theta-4)^2$, which has a unique and well-defined minimum at $\theta^*=4\,.$ We begin with the nonlocal model of AdaGrad, continue with RMSProp, and conclude with Adam.

\begin{figure}[h!]
\centering
\includegraphics[width=0.9\textwidth]{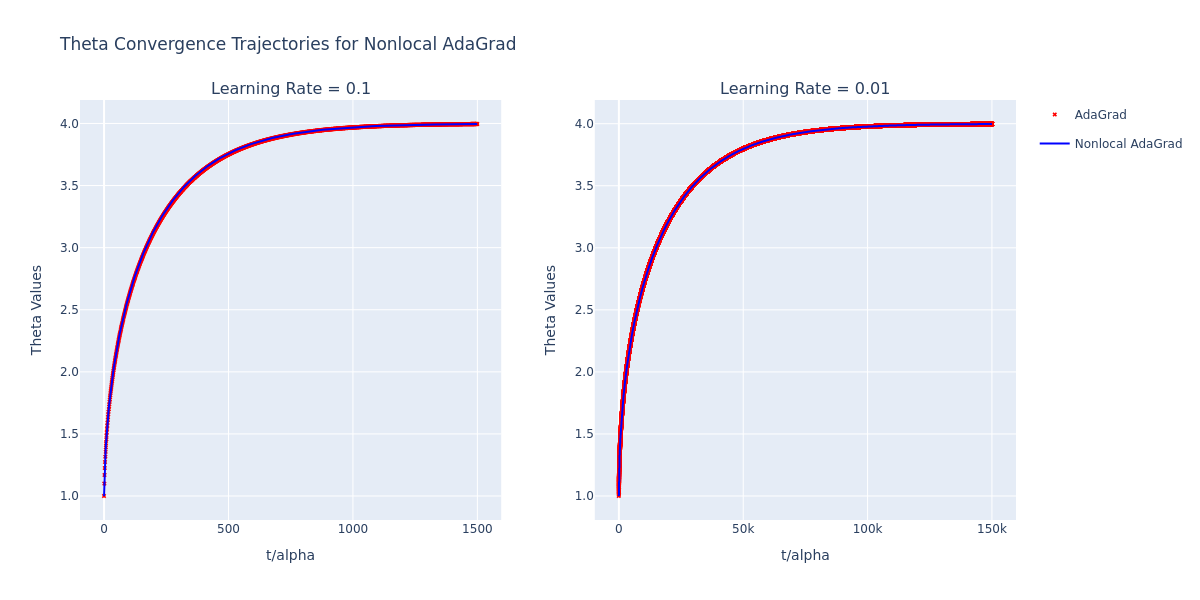}
\caption{\textbf{Convergence of \(\theta(t)\) using the first-order nonlocal continuous AdaGrad method.} The plot illustrates the convergence trajectories for minimizing the function \((\theta - 4)^2\) using two different learning rates: \(0.1\) (left) and \(0.01\) (right). At a higher learning rate (\(0.1\)), the algorithm rapidly converges to the target value \(\theta = 4\), stabilizing in under 1,500 k-iterations. With a lower learning rate (\(0.01\)), the convergence is more gradual, reaching the target around 100,000 k-iterations.}
\label{fig:AdaGrad_Nonlocal_theta}
\end{figure}

\paragraph{Nonlocal AdaGrad:} Figure \ref{fig:AdaGrad_Nonlocal_theta} shows the convergence of \(\theta\) values using the nonlocal continuous AdaGrad method for minimizing the function \((\theta - 4)^2\), evaluated at two learning rates: \(0.1\) and \(0.01\). This figure demonstrates that the nonlocal continuous AdaGrad method effectively replicates the optimization dynamics of the conventional AdaGrad method.

\begin{figure}[h!]
\centering
\includegraphics[width=0.9\textwidth]{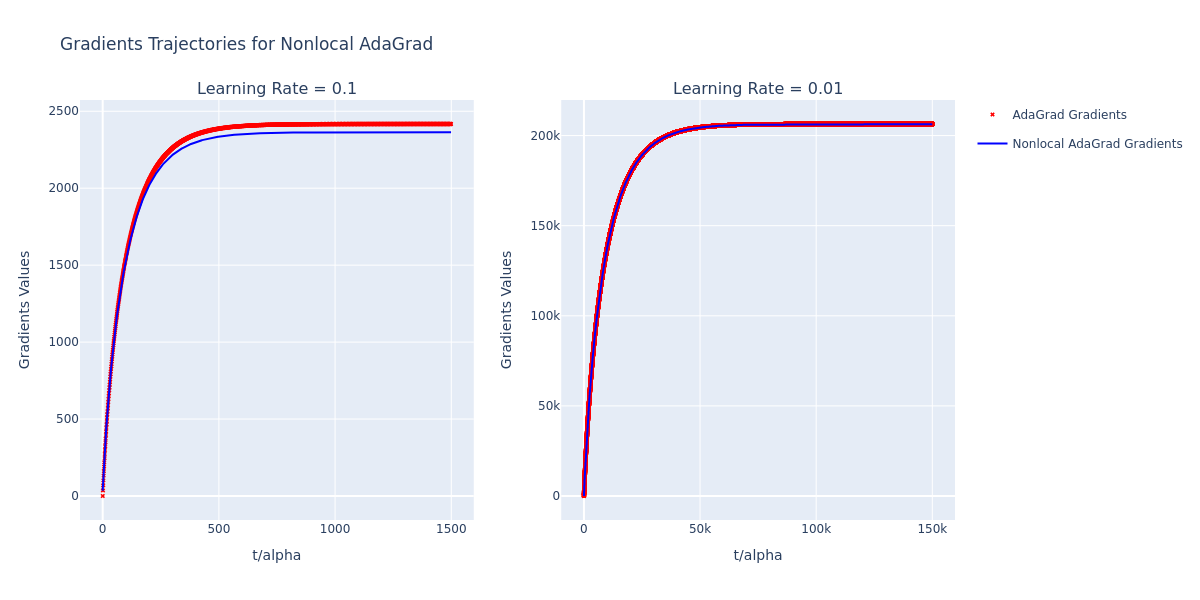}
\caption{\textbf{Accumulated gradients \(G(t)\) convergence trajectories using the first-order nonlocal continuous AdaGrad method.} The plot shows that the nonlocal continuous AdaGrad method exhibits a nearly identical gradient accumulation behavior to the conventional AdaGrad method, with rapid convergence and stabilization at both learning rates.}
\label{fig:AdaGrad_Nonlocal_G}
\end{figure}

Figure \ref{fig:AdaGrad_Nonlocal_G} depicts the evolution of the accumulated gradients \(G(t)\) under the nonlocal continuous AdaGrad method. For a learning rate of \(0.1\), the method converges to the target value \(\theta = 4\), with \(G(t)\) stabilizing around 1,000 k-iterations. While this behavior closely matches the results observed with the conventional AdaGrad, a slight difference can be noted: the nonlocal continuous method reaches saturation slightly earlier than the discrete version.

When the learning rate is reduced to \(0.01\), the nonlocal continuous AdaGrad method still converges toward \(\theta = 4\), with \(G(t)\) reaching slightly above 100,000 k-iterations. As shown in Figure \ref{fig:AdaGrad_Nonlocal_G}, as the learning rate decreases, the nonlocal continuous method and the discrete AdaGrad model become nearly identical, both requiring approximately 150,000 k-iterations to achieve overall stabilization.

In these simulations, AdaGrad corresponds to the cumulative case $\xi=0$ in Theorems~\ref{thm:adagrad_rmsprop_memory} and~\ref{thm:extended_memory_adagrad_rmsprop}. Theorem~\ref{thm:adagrad_rmsprop_memory} should be interpreted as a preliminary a priori estimate: it yields the robust subexponential bound $\|\theta(t)-\theta^*\|\leq C e^{-c\sqrt{t}}$, which is sufficient to control the accumulated memory term. The sharper interpretation of the AdaGrad 
dynamics is then provided by Theorem~\ref{thm:extended_memory_adagrad_rmsprop}. Since $\xi=0$, there is no forgetting term, and the extended functional satisfies
\[
    \dot E_\rho(t)\leq -m\mu_0\Phi(t)\leq 0.
\]
Thus $E_\rho$ is nonincreasing, but it is not forced to decay exponentially. Nevertheless, once the accumulated memory has been shown to be bounded, the effective factor
\[
    \mu(t)=\frac{1}{\sqrt{G(t+\alpha)}+\varepsilon}
\]
is bounded from below by a positive constant, which yields exponential convergence of $\theta(t)$ to the minimizer. At the same time, the accumulated squared-gradient memory $G(t)$ is monotone increasing and converges to a finite, 
generally nonzero, limit $G_\infty$, as shown in Fig.~\ref{fig:AdaGrad_Nonlocal_G}. Consequently, $\mu(t)$ decreases and stabilizes at
\[
    \frac{1}{\sqrt{G_\infty}+\varepsilon},
\]
which explains the gradual flattening of the $\theta$-trajectories in Fig.~\ref{fig:AdaGrad_Nonlocal_theta}.

\begin{figure}[h!]
\centering
\includegraphics[width=0.9\textwidth]{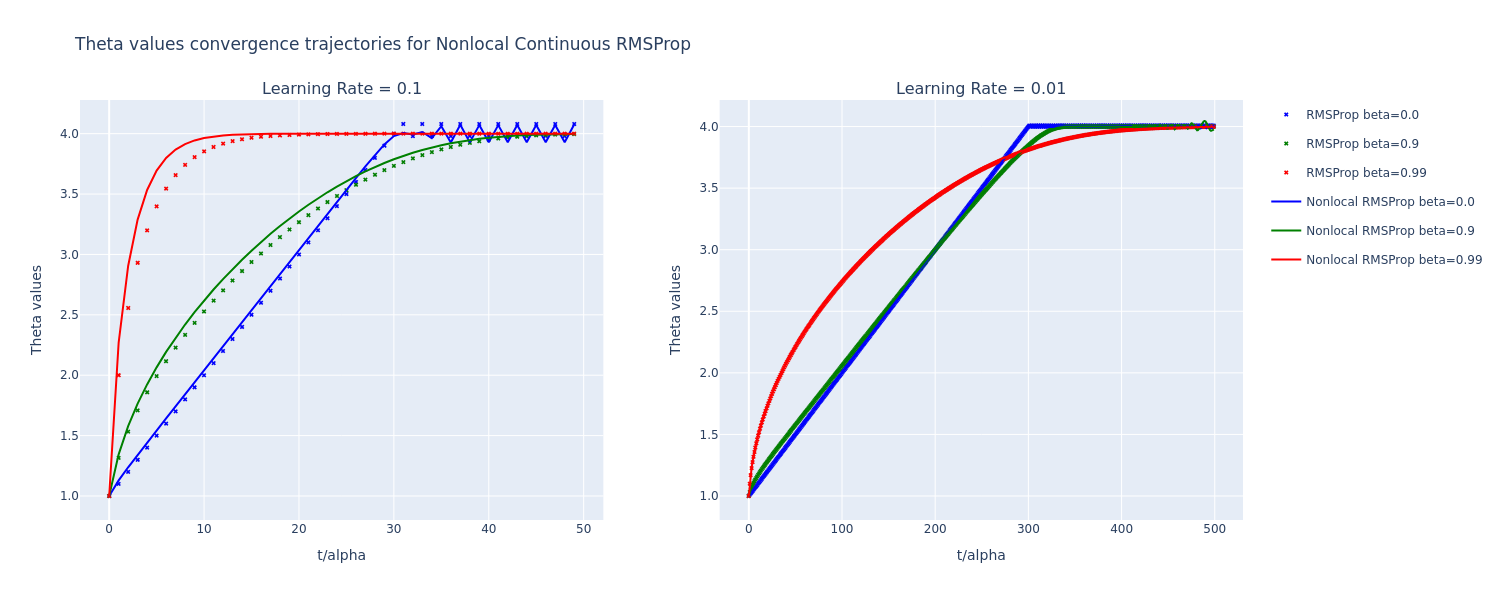}
\caption{\textbf{Convergence of \(\theta(t)\) using the first-order nonlocal continuous RMSProp method.} The plot shows the convergence to the minimum value of \(\theta = 4\) for the convex function \((\theta - 4)^2\). The left subplot corresponds to a learning rate of 0.1, while the right subplot uses a learning rate of 0.01. With a learning rate of 0.1, \(\beta = 0.0\) exhibits more noticeable oscillations, whereas \(\beta = 0.9\) and \(0.99\) converge more smoothly. For a learning rate of 0.01, a slight destabilization occurs for \(\beta = 0.9\) as the solution approaches the final result, caused by the numerical method. At higher learning rates, slight differences can be observed between the models: for \(\beta = 0.0\), the oscillations in the discrete case start immediately upon reaching the minimum, whereas in the continuous case, they take a few k-iterations to begin. On the other hand, for \(\beta = 0.99\), the curve is less pronounced in the discrete model compared to the continuous one.
}
\label{fig:RMSProp_Nonlocal_theta}
\end{figure}

\paragraph{Nonlocal RMSProp:} Figure \ref{fig:RMSProp_Nonlocal_theta} presents the convergence trajectories for \(\theta\) values using the first-order nonlocal continuous RMSProp optimizer. The figure includes two subplots corresponding to different learning rates (0.1 on the left and 0.01 on the right) and varying \(\beta\) parameters (\(0.0\), \(0.9\), \(0.99\)).

For \(\theta\), with a learning rate of 0.1, \(\beta = 0.99\) leads to faster convergence, while \(\beta = 0.0\) and \(\beta = 0.9\) result in slower convergence, with the latter exhibiting more linear behavior but showing noticeable oscillations. When the learning rate is reduced to 0.01, the overall convergence slows, and the oscillatory behavior for \(\beta = 0.0\) becomes less pronounced\footnote{A slight oscillation is observed for \(\beta = 0.9\) and a learning rate of 0.01, induced by the numerical method, as the convergence error does not decrease below \(1 \times 10^{-4}\).}.

\begin{figure}[h!]
\centering
\includegraphics[width=0.9\textwidth]{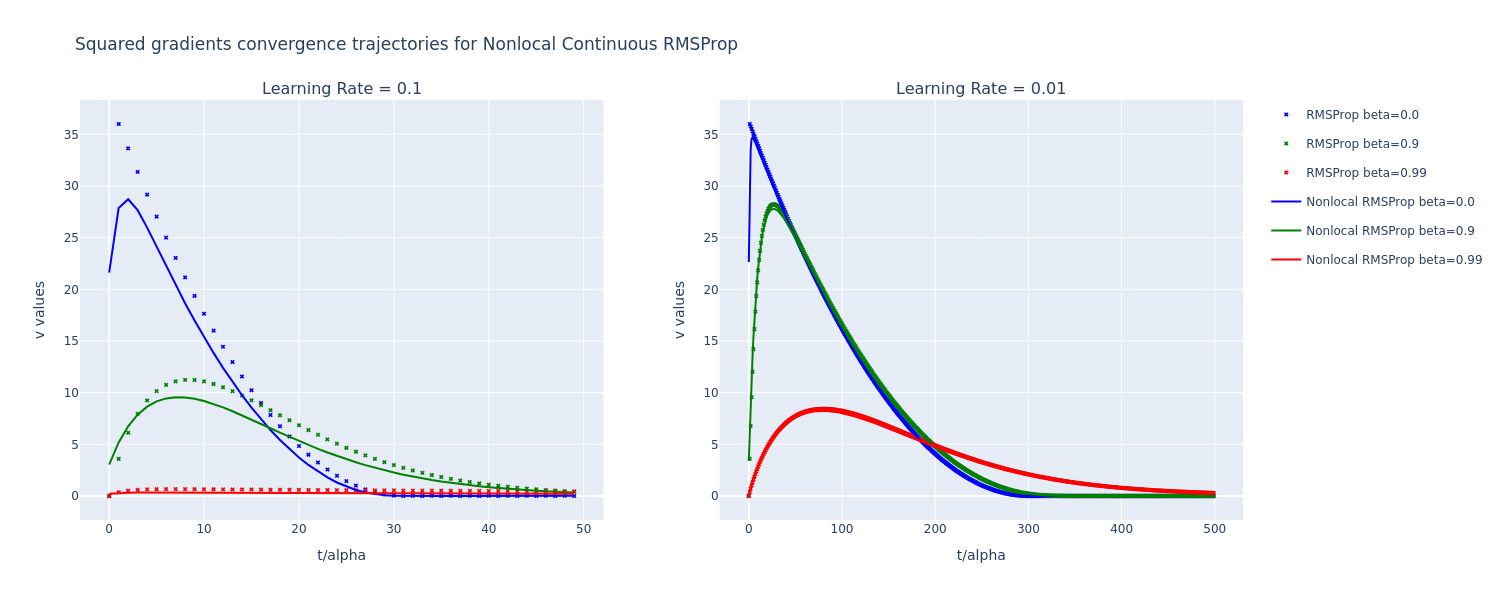}
\caption{\textbf{Convergence trajectories of \(v(t)\) for the first-order nonlocal continuous RMSProp method.} This plot illustrates the convergence of the squared gradient moving average \(v(t)\). The left subplot represents a learning rate of 0.1, while the right subplot corresponds to a learning rate of 0.01. For \(\beta\) values of 0.0 and 0.9, a slight initial bump is noticeable, but both eventually decay towards zero. The main differences between the continuous and discrete models appear at a learning rate of \(0.1\), where the values are slightly higher for \(\beta = 0.9\) and \(\beta = 0.99\), and for \(\beta = 0.0\) a small bump is observed instead of a direct descent.}
\label{fig:RMSProp_Nonlocal_G}
\end{figure}

Figure \ref{fig:RMSProp_Nonlocal_G} shows the behavior of the squared gradients \(v(t)\) under the nonlocal continuous RMSProp method. With a learning rate of 0.1, lower \(\beta\) values initially produce a bump followed by a decay, while higher \(\beta\) values exhibit a less pronounced bump. When the learning rate is decreased, this bump becomes more noticeable.

When compared to the standard RMSProp optimizer, the nonlocal continuous variant exhibits a broadly similar convergence pattern, albeit with subtle differences. For instance, for \(\beta = 0.0\) and \(\beta = 0.9\), the nonlocal continuous model displays a slightly different dynamic than the discrete case, attaining marginally higher values. However, as the learning rate decreases, these differences gradually vanish, and the two models become nearly indistinguishable.

Finally, these simulations correspond to the forgetting case $\xi=\lambda$ in Theorems~\ref{thm:adagrad_rmsprop_memory} and~\ref{thm:extended_memory_adagrad_rmsprop}. In RMSProp, the memory variable $v(t)$ satisfies a dissipative equation of the form
\[
    \dot v(t)=\lambda\|g(t)\|^2-\lambda v(t),
    \qquad 
    \lambda=\frac{1-\beta}{\alpha},
\]
so that the term $-\lambda v(t)$ acts as exponential forgetting. Therefore, the extended functional
\[
    E_\rho(t)=\Phi(t)+\rho v(t)
\]
satisfies, by Theorem~\ref{thm:extended_memory_adagrad_rmsprop},
\[
    \dot E_\rho(t)\leq -\kappa_0 E_\rho(t),
    \qquad 
    \kappa_0=\min\{m\mu_0,\lambda\}>0.
\]
Consequently, $E_\rho(t)$, the objective gap $\Phi(t)$, and the memory variable $v(t)$ decay exponentially. This is consistent with the behavior observed in Fig.~\ref{fig:RMSProp_Nonlocal_G}: after the initial transient, $v(t)$ decays towards zero. Moreover, the decay becomes slower as $\beta$ increases, since $\lambda=(1-\beta)/\alpha$ becomes smaller and the forgetting mechanism is weaker. Finally, the convergence of the $\theta$-trajectories in Fig.~\ref{fig:RMSProp_Nonlocal_theta} is guaranteed first by the preliminary a priori estimate of Theorem~\ref{thm:adagrad_rmsprop_memory}, while the sharper exponential interpretation in the RMSProp case follows from the closed dissipation estimate for $E_\rho$.

\begin{figure}[h!]
\centering
\includegraphics[width=0.9\textwidth]{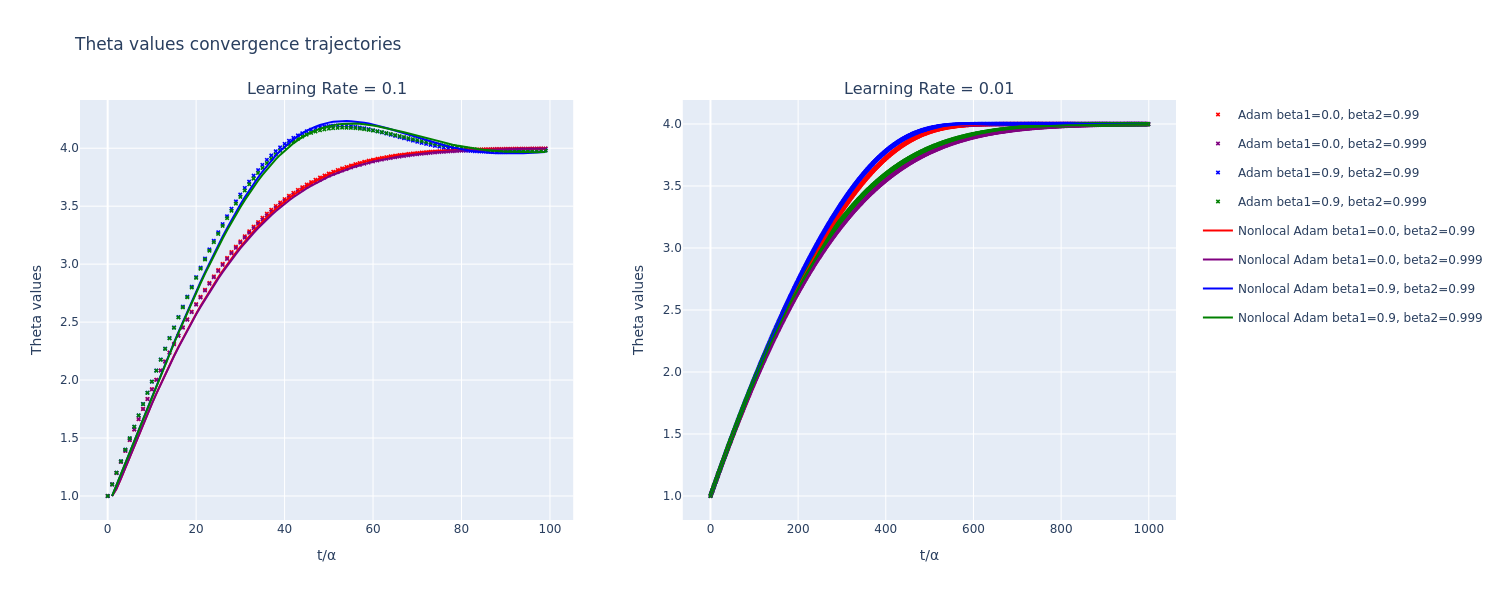}
\caption{\textbf{Convergence of \(\theta(t)\) using the first-order nonlocal continuous Adam method.} The plot illustrates the convergence trajectories of \(\theta\)-values for the first-order nonlocal continuous Adam model under different parameter settings of \(\beta_1\) and \(\beta_2\) with two distinct learning rates (\(0.1\) and \(0.01\)). For \(\beta_1 = 0.9\) and a learning rate of 0.1, a noticeable oscillation is observed, requiring a longer time to stabilize at the minimum value. This behavior improves when the learning rate is reduced to 0.01. 
}
\label{fig:Adam_Nonlocal_theta}
\end{figure}

\begin{figure}[h!]
\centering
\includegraphics[width=0.9\textwidth]{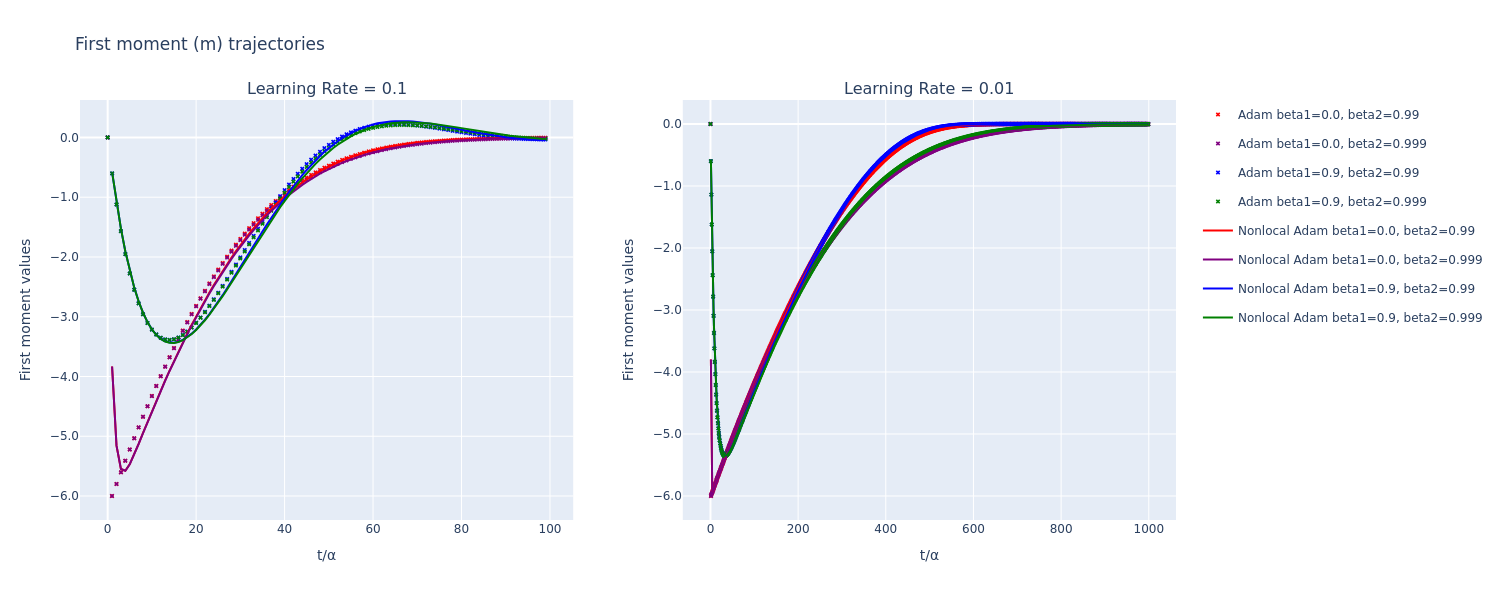}
\caption{\textbf{Convergence trajectories of \(m(t)\) for the first-order nonlocal continuous Adam optimizer.} The plot shows the evolution of the first moment \(m(t)\) over time, scaled by the learning rate. With a higher learning rate (\(\alpha = 0.1\)), the values of \(m(t)\) exhibit certain oscillations, whereas a lower learning rate (\(\alpha = 0.01\)) results in a smoother and slower convergence.
}
\label{fig:Adam_Nonlocal_M}
\end{figure}

\begin{figure}[h!]
\centering
\includegraphics[width=0.9\textwidth]{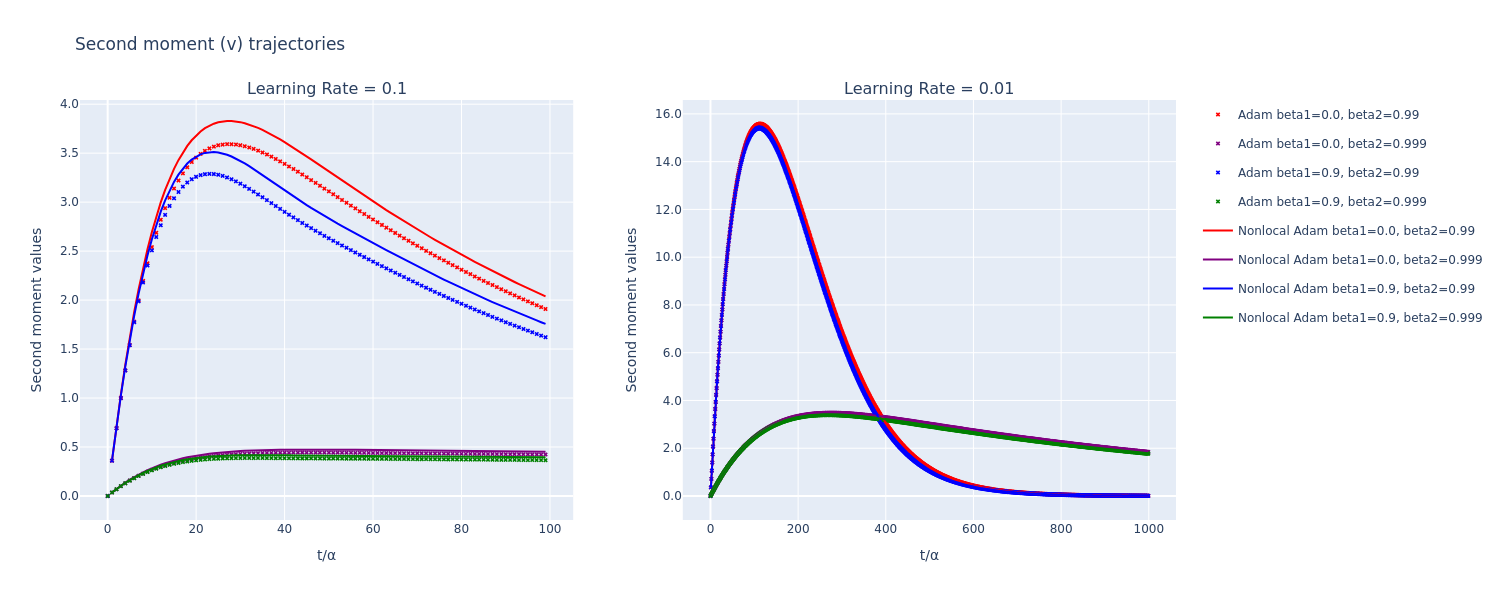}
\caption{\textbf{Convergence trajectories of \(v(t)\) for the first-order nonlocal continuous Adam optimizer.} The plot illustrates the convergence of the squared gradient moving average \(v(t)\). At a higher learning rate (\(\alpha = 0.1\)), the peaks of \(v(t)\) vary according to the parameter settings, with \(\beta_1 = 0.0\) and \(\beta_2 = 0.99\) producing the most significant peak and the slowest rate of decay. When the learning rate is reduced to \(0.01\), all curves display a more pronounced initial peak followed by a gradual decline.
}
\label{fig:Adam_Nonlocal_G}
\end{figure}

\paragraph{Nonlocal Adam:} Figures \ref{fig:Adam_Nonlocal_theta}, \ref{fig:Adam_Nonlocal_M}, and \ref{fig:Adam_Nonlocal_G} illustrate the behavior of the first-order nonlocal continuous Adam optimization model by showing the convergence trajectories of \(\theta(t)\), \(m(t)\), and \(v(t)\) over time for different parameter settings (\(\beta_1\) and \(\beta_2\)) and learning rates (\(\alpha = 0.1\) and \(\alpha = 0.01\)).

Figure \ref{fig:Adam_Nonlocal_theta} shows the convergence of \(\theta\). For a higher learning rate (\(\alpha = 0.1\)), convergence is rapid, with different combinations of \(\beta_1\) and \(\beta_2\) reaching a stable value in fewer steps. However, for configurations with \(\beta_1 = 0.9\), although convergence occurs more quickly, there are small oscillations, making the process slightly more unstable before reaching the minimum value. These oscillations are reduced when the learning rate is lowered to \(\alpha = 0.01\). When the learning rate is lower, convergence is slower and more stable, with subtle differences in behavior between various parameter settings. 

Figure \ref{fig:Adam_Nonlocal_M} highlights the evolution of the first moment \(m(t)\). For a learning rate of \(\alpha = 0.1\), \(m(t)\) demonstrates a more pronounced oscillatory behavior before stabilizing around zero. When compared to the standard Adam optimizer's convergence of the first moment \(m_k\), both the nonlocal and standard Adam models show identical initial oscillatory patterns. With a reduced learning rate (\(\alpha = 0.01\)), both models achieve smoother and more gradual convergence.

Figure \ref{fig:Adam_Nonlocal_G} shows the evolution of the second moment \(v(t)\). For a higher learning rate (\(\alpha = 0.1\)), the nonlocal continuous Adam model displays varying peak values depending on the parameter settings. Configurations such as \(\beta_1 = 0.0\) and \(\beta_2 = 0.99\) produce the highest peaks and the slowest decay. When compared to the standard Adam optimizer's second moment \(v_k\), both models follow similar trends, with distinct peak behaviors based on the parameter settings. Notably, the peaks for \(\beta_1 = 0.0,\,\,\beta_2 = 0.99\,\) and \(\beta_1 = 0.9,\,\,\beta_2 = 0.99\,\) are generally higher in the nonlocal model.

For a smaller learning rate (\(\alpha = 0.01\)), the model shows an increase in the size of the peaks across all parameter settings, with a more pronounced bump for \(\beta_1 = 0.0\) and \(\beta_2 = 0.99\) or \(\beta_1 = 0.9\) and \(\beta_2 = 0.99\) , identical to the behavior observed in the standard case.

To conclude this section, let us analyze these plots in light of Theorem~\ref{thm:adam_shifted}. In the Adam formulation, the Lyapunov analysis is written in terms of the shifted quantities $\widehat m(t):=m(t+\alpha)$ and $\widehat v(t):=v(t+\alpha)$ and of the effective Adam gain
\[
    \widehat a(t)
    :=
    \frac{\eta(t+\alpha)}
    {\sqrt{\widehat v(t)}+\varepsilon(t+\alpha)}.
\]
Under the scalar Lyapunov condition $\mathcal H_{\beta_1,\beta_2}<2(1-\beta_1)$, together with the one-sided shift-defect estimate, the Lyapunov functional
\[
    V(t)
    :=
    \Phi(t)
    +
    \frac{\widehat a(t)}{2\lambda_1}\|\widehat m(t)\|^2
\]
satisfies
\[
    \dot V(t)
    \leq
    -\widehat a(t)(c_\beta-\sigma_\alpha)\|\widehat m(t)\|^2
    \leq 0.
\]
Consequently, Theorem~\ref{thm:adam_shifted} yields
\[
    \widehat m(t)\to0,
    \qquad
    \nabla f(\theta(t))\to0,
    \qquad
    \theta(t)\to\theta^*
    \quad\text{as }t\to\infty.
\]
The second moment does not need to appear as a separate term in the Lyapunov functional, since its effect is encoded in the gain $\widehat a(t)$. Therefore, the decay observed in the $m(t)$ and $v(t)$ plots, together with the convergence of the $\theta$-trajectories, is consistent with the corrected stability mechanism for the shifted continuous Adam dynamics.

\subsubsection{Nonconvex Function Simulations}

Finally, we present the simulations for the non-convex function $f(\theta)=\frac{1}{4}(\theta^2-1)^2$, which has different local minima at $\theta=\pm1$ and an unstable critical point (a local maximum) at $0$. In this subsection, we restrict our discussion to the case $\theta=\pm1$, since it is of particular interest how a change in the initial conditions may lead the system to one minimum or the other. For illustrations of the case $\theta=0$, please refer to Appendix~\ref{Appendix}. All the figures are placed at Appendix~\ref{App:SF}.

Before proceeding, we would like to make a brief observation regarding the position of the minima $\pm 1$ of the nonconvex function under study. Note that they are located as mirror images of each other, so it is to be expected that the simulations will exhibit mirror symmetry when comparing the negative and positive cases, as we shall see for AdaGrad, RMSProp, and Adam.

\begin{figure}[htbp]
  \centering
  \begin{subfigure}{\linewidth}
    \centering
    \includegraphics[width=0.9\linewidth]{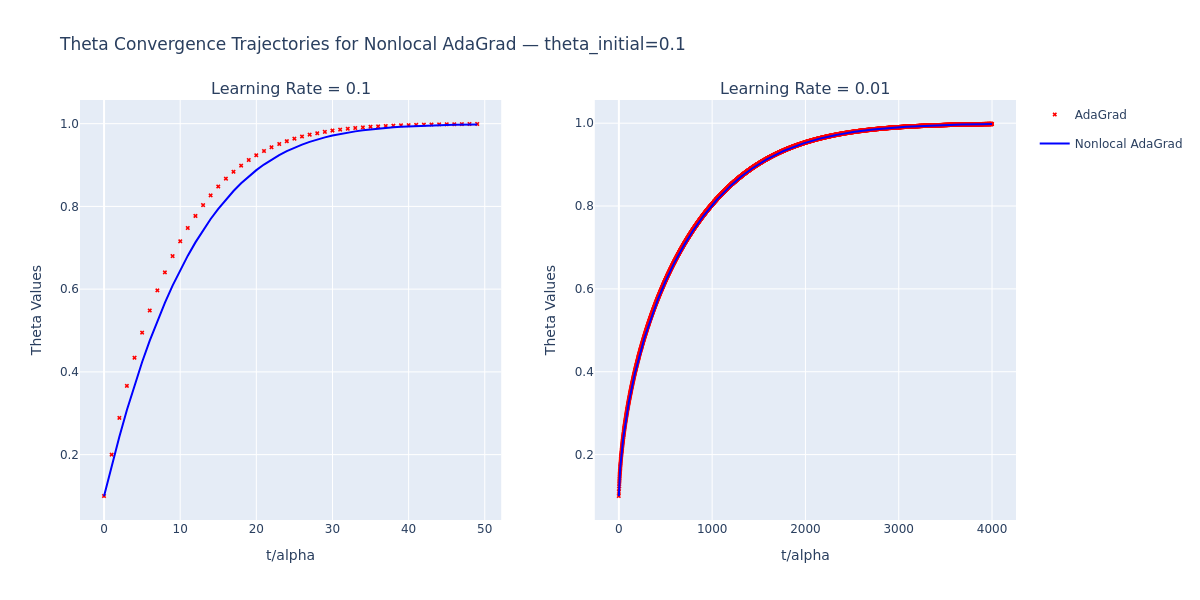}
    \caption{\textbf{Convergence of \(\theta(t)\) with \(\theta_0 = 0.1\) for AdaGrad}. The nonlocal method and discrete AdaGrad converge monotonically towards \(\theta^*\approx 1\). With \(\alpha=0.1\), the nonlocal approximation is slightly smoother; with \(\alpha=0.01\), both trajectories overlap almost completely, although they require many more iterations.}
    \label{fig:adagrad-nonlocal-t_ncx_p}
  \end{subfigure}

  \vspace{0.75em}

  \begin{subfigure}{\linewidth}
    \centering
    \includegraphics[width=0.9\linewidth]{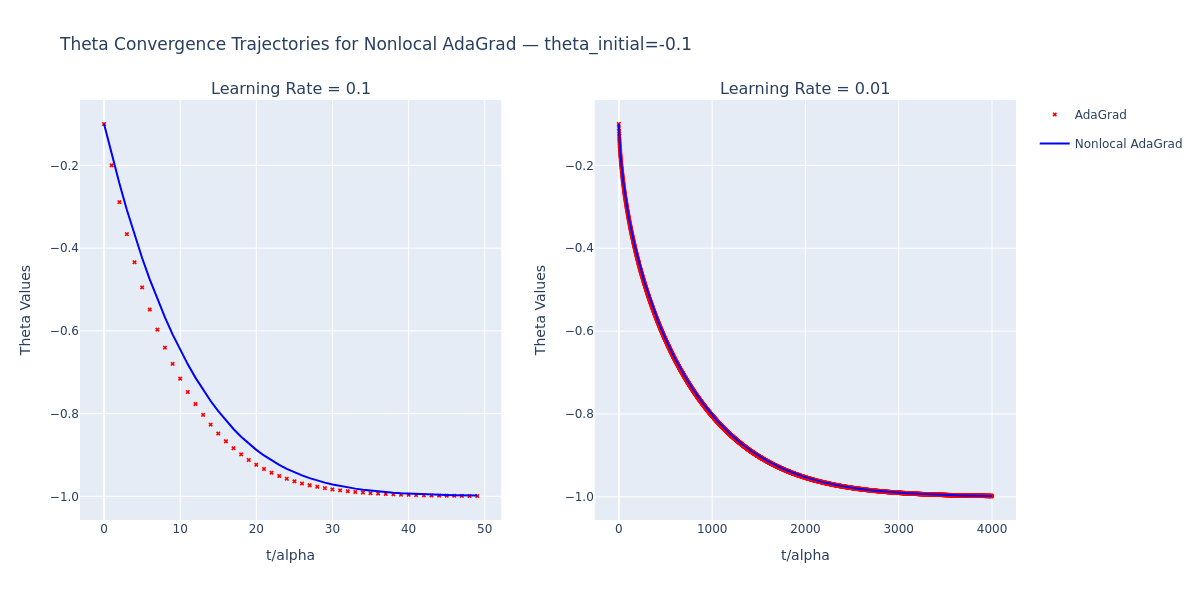}
    \caption{\textbf{Convergence of \(\theta(t)\) with \(\theta_0 =-0.1\) for AdaGrad}. Symmetric behavior with respect to the positive case: monotonic convergence to \(\theta^*\approx - 1\). At \(\alpha=0.1\), the nonlocal model is slightly slower at the beginning; at \(\alpha=0.01\), both curves essentially coincide throughout the entire interval.}
    \label{fig:adagrad-nonlocal-t_ncx_n}
  \end{subfigure}

  \caption{\textbf{AdaGrad numerical simulations}. These two plots show the evolution of the trajectory of \(\theta(t)\) for the nonconvex function \(\tfrac{1}{4}(\theta^2-1)^2\) with two initial conditions $\theta_0=\pm0.1$ for AdaGrad. }
  \label{fig:adagrad-nonlocal-ncx_0}
\end{figure}

\begin{figure}[htbp]
  \centering
  \begin{subfigure}{\linewidth}
    \centering
    \includegraphics[width=0.9\linewidth]{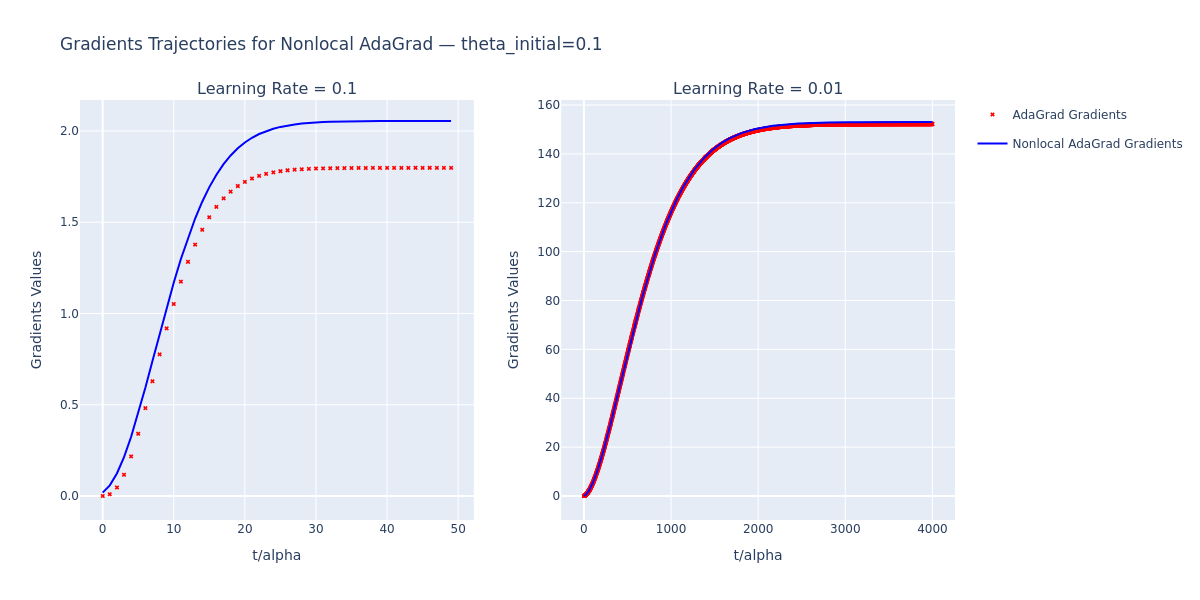}
    \caption{\textbf{Gradient trajectories for nonlocal and discrete AdaGrad dynamics with initial value \(\theta_0 = 0.1\).} Gradient trajectories with \(\theta_0=0.1\). For \(\alpha=0.1\), the nonlocal model reaches a slightly higher plateau and shows a smoother transition than discrete AdaGrad; for \(\alpha=0.01\), both trajectories overlap almost completely throughout \(t/\alpha\).}
    \label{fig:adagrad-nonlocal-G_ncx_p}
  \end{subfigure}

  \vspace{0.75em}

  \begin{subfigure}{\linewidth}
    \centering
    \includegraphics[width=0.9\linewidth]{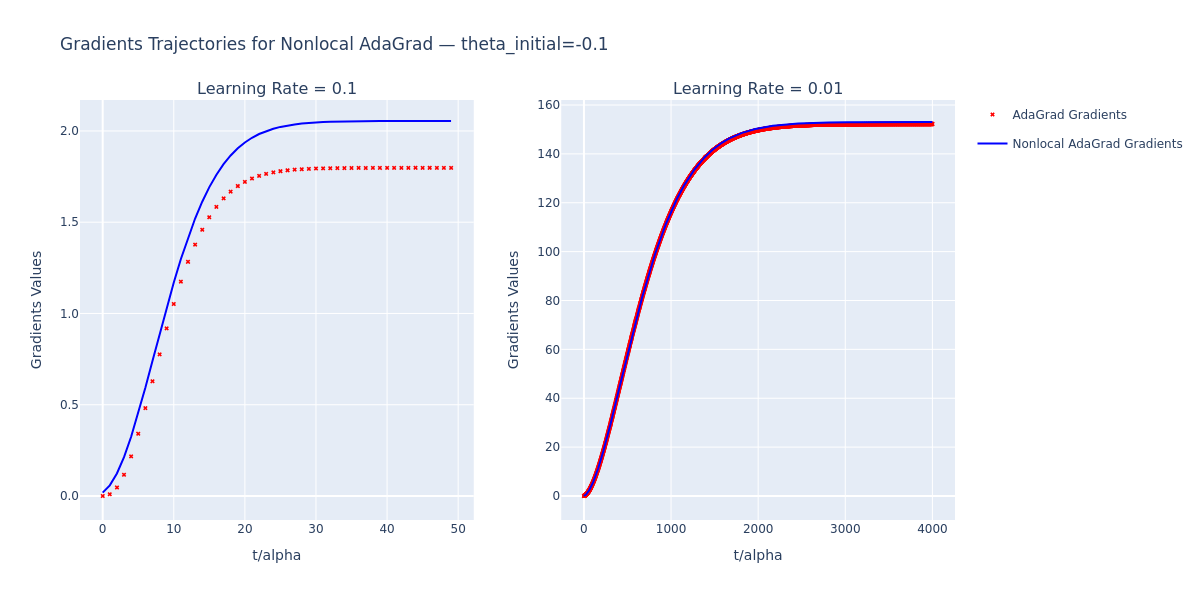}
    \caption{\textbf{Gradient trajectories for nonlocal and discrete AdaGrad dynamics with initial value \(\theta_0 =-0.1\).}  Gradient trajectories with \(\theta_0=-0.1\). Results analogous to the case \(\theta_0=0.1\): at \(\alpha=0.1\) the nonlocal term is smoother and reaches a slightly higher plateau; at \(\alpha=0.01\) the agreement with the discrete scheme is practically exact.}
    \label{fig:adagrad-nonlocal-G_ncx_n}
  \end{subfigure}

  \caption{\textbf{Gradient trajectories of AdaGrad}. These two plots show the evolution of the accumulated squared-gradient memory \(G(t)\) for the nonconvex function \(\tfrac{1}{4}(\theta^2-1)^2\) with two initial conditions $\theta_0=\pm0.1$ for AdaGrad.}
  \label{fig:adagrad-nonlocal-G_ncx}
\end{figure}

\paragraph{Nonlocal AdaGrad:} Figures~\ref{fig:adagrad-nonlocal-t_ncx_p}, \ref{fig:adagrad-nonlocal-t_ncx_n}, \ref{fig:adagrad-nonlocal-G_ncx_p}, and \ref{fig:adagrad-nonlocal-G_ncx_n} illustrate the behavior of the nonlocal continuous AdaGrad model compared with its discrete counterpart in a nonconvex setting, for two initial conditions of \(\theta\) (\(\theta_0=-0.1\) and \(\theta_0=0.1\)) and two learning rates (\(\alpha=0.1\) and \(\alpha=0.01\)). 

The trajectories in Figures~\ref{fig:adagrad-nonlocal-t_ncx_p} and \ref{fig:adagrad-nonlocal-t_ncx_n} converge monotonically toward the stable minima (\(\theta^*\approx -1\) and \(\theta^*\approx 1\)). For \(\alpha=0.1\), the nonlocal model approaches the minimum a bit more slowly than the discrete version, particularly in the early stages, although both coincide in the long run. When the learning rate is reduced to \(\alpha=0.01\), the two trajectories almost overlap, confirming the accuracy of the continuous model for smaller values of \(\alpha\) and highlighting the system’s symmetry: starting at \(\pm 0.1\) produces mirror profiles toward \(\pm 1\), as expected.

The gradient plots in Figures~\ref{fig:adagrad-nonlocal-G_ncx_p} and \ref{fig:adagrad-nonlocal-G_ncx_n} show a similar picture. With \(\alpha=0.1\), the nonlocal model reaches a slightly higher plateau but follows a smoother transition, while for \(\alpha=0.01\) the two curves are nearly indistinguishable across the whole time scale. 

Let us discuss these figures in light of Lemma~\ref{lem:nonconvex_L2_gradient_memory}, Theorem~\ref{thm:nonconvex_PL_KL_adagrad_rmsprop}, and Theorem~\ref{thm:nonconvex_extended_memory_adagrad_rmsprop}. In the figures with $\theta_0<0$ (Fig.~\ref{fig:adagrad-nonlocal-t_ncx_n}) and $\theta_0>0$ (Fig.~\ref{fig:adagrad-nonlocal-t_ncx_p}), the trajectories remain in their corresponding attraction basins and settle at the stable critical points $\theta^*\simeq -1$ and $\theta^*\simeq 1$, respectively. This behavior  is consistent with Lemma~\ref{lem:nonconvex_L2_gradient_memory}, which gives boundedness of the trajectory and of the memory, together with the asymptotic stationarity property
\[
    \|\nabla f(\theta(t))\|\to 0.
\]
Thus, accumulation points of the trajectory belong to the critical set inside $\mathcal S_0$. In the present one-dimensional toy model, the two initial conditions $\theta_0=\pm0.1$ lie in the basins of the stable minima $\pm1$, which explains the observed convergence to those values.

Moreover, the near-exponential shape observed close to the limiting points is consistent with the PL regime, or equivalently with the KL case with exponent $\sigma=1/2$, as described in Theorem~\ref{thm:nonconvex_PL_KL_adagrad_rmsprop}. For AdaGrad, namely the cumulative case $\xi=0$ in Theorem~\ref{thm:nonconvex_extended_memory_adagrad_rmsprop}, the extended functional
\[
    E_\rho(t)=\Phi(t)+\rho\,G(t)
\]
is nonincreasing, but the dissipation estimate cannot be closed in the form $\dot E_\rho\leq -\kappa E_\rho$ for some $\kappa>0$. This reflects the absence of a forgetting term in AdaGrad. Consequently, the accumulated squared-gradient memory $G(t)$ is nondecreasing and converges to a finite, generally nonzero, limit. Therefore, no exponential decay rate for $G(t)$ should be claimed in this case, as illustrated in Figs.~\ref{fig:adagrad-nonlocal-G_ncx_p} and~\ref{fig:adagrad-nonlocal-G_ncx_n}.

\begin{figure}[htbp]
  \centering
  \begin{subfigure}{\linewidth}
    \centering
    \includegraphics[width=0.9\linewidth]{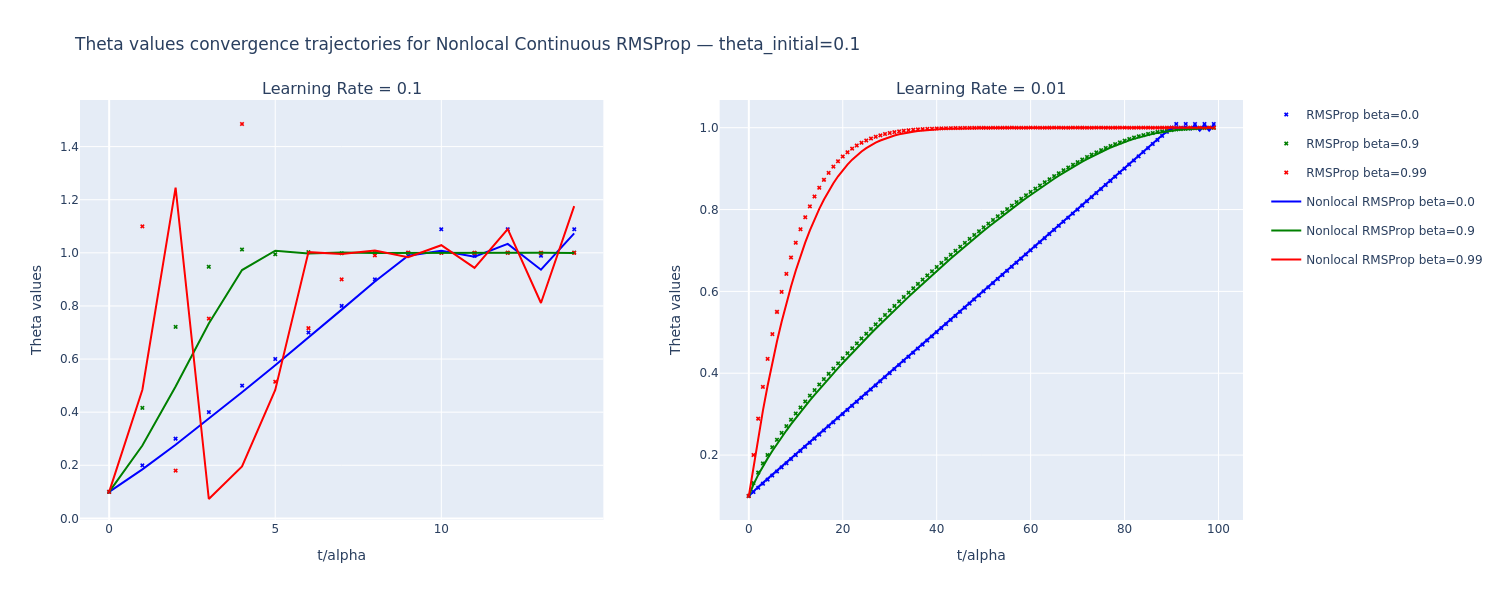}
    \caption{\textbf{Convergence trajectories of \(\theta(t)\) with \(\theta_0=0.1\) for the nonlocal and discrete models.} At \(\alpha=0.1\), high momentum (\(\beta=0.99\)) leads to pronounced overshoot/oscillations; \(\beta=0\) and \(0.9\) are smoother. At \(\alpha=0.01\), all settings converge smoothly with ordering of speeds \(\beta=0.99 > 0.9 > 0\). Nonlocal and discrete trajectories closely agree, especially for the smaller learning rate.}
    \label{fig:rmsprop-nonlocal-t_ncx_p}
  \end{subfigure}

  \vspace{0.75em}

  \begin{subfigure}{\linewidth}
    \centering
    \includegraphics[width=0.9\linewidth]{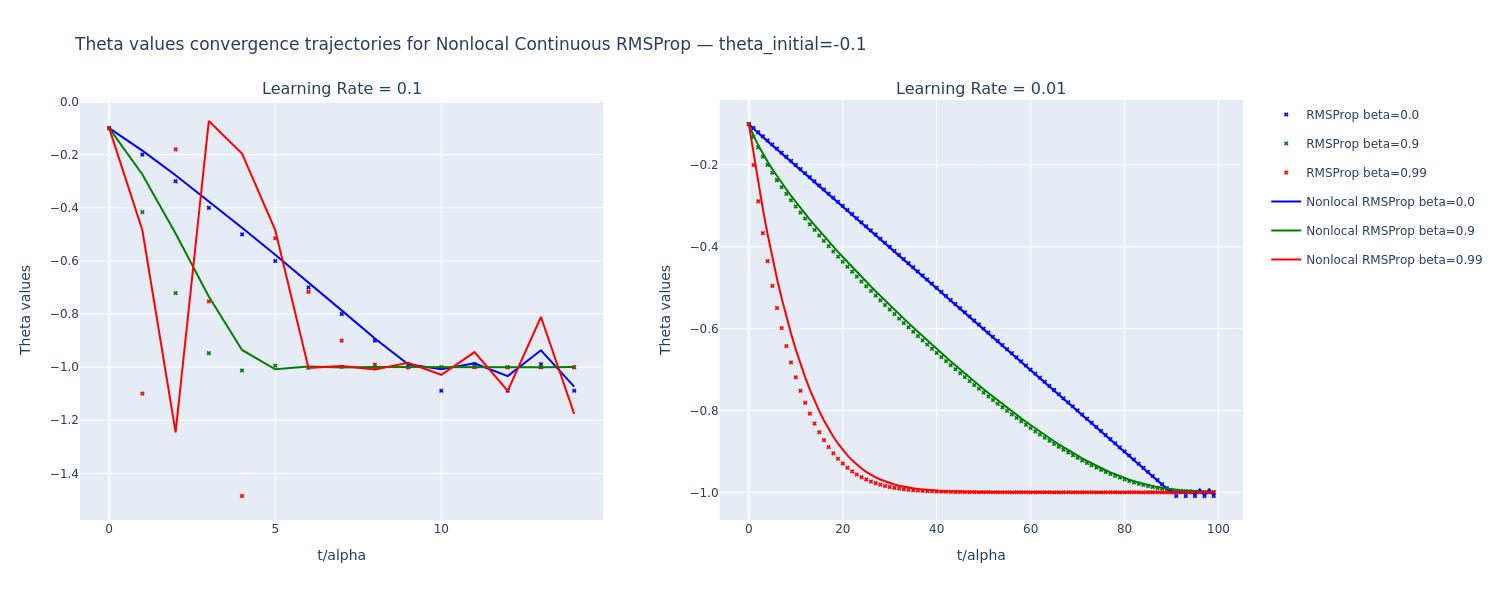}
    \caption{\textbf{Convergence trajectories of \(\theta(t)\) with \(\theta_0=-0.1\) for the nonlocal and discrete models.}  Mirror behavior of the positive case: convergence to \(\theta^*\approx- 1\) with oscillations for \(\beta=0.99\) when \(\alpha=0.1\), and smooth, rate-ordered convergence for \(\alpha=0.01\).}
    \label{fig:rmsprop-nonlocal-t_ncx_n}
  \end{subfigure}

  \caption{\textbf{Convergence trajectories of \(\theta(t)\) for the nonlocal and discrete RMSProp models.} These two plots show the evolution of the trajectory of \(\theta(t)\) for the nonconvex function \(\tfrac{1}{4}(\theta^2-1)^2\) with two initial conditions $\theta_0=\pm0.1$ for RMSProp. }
  \label{fig:rmsprop-nonlocal-t_ncx}
\end{figure}

\begin{figure}[htbp]
  \centering
  \begin{subfigure}{\linewidth}
    \centering
    \includegraphics[width=0.9\linewidth]{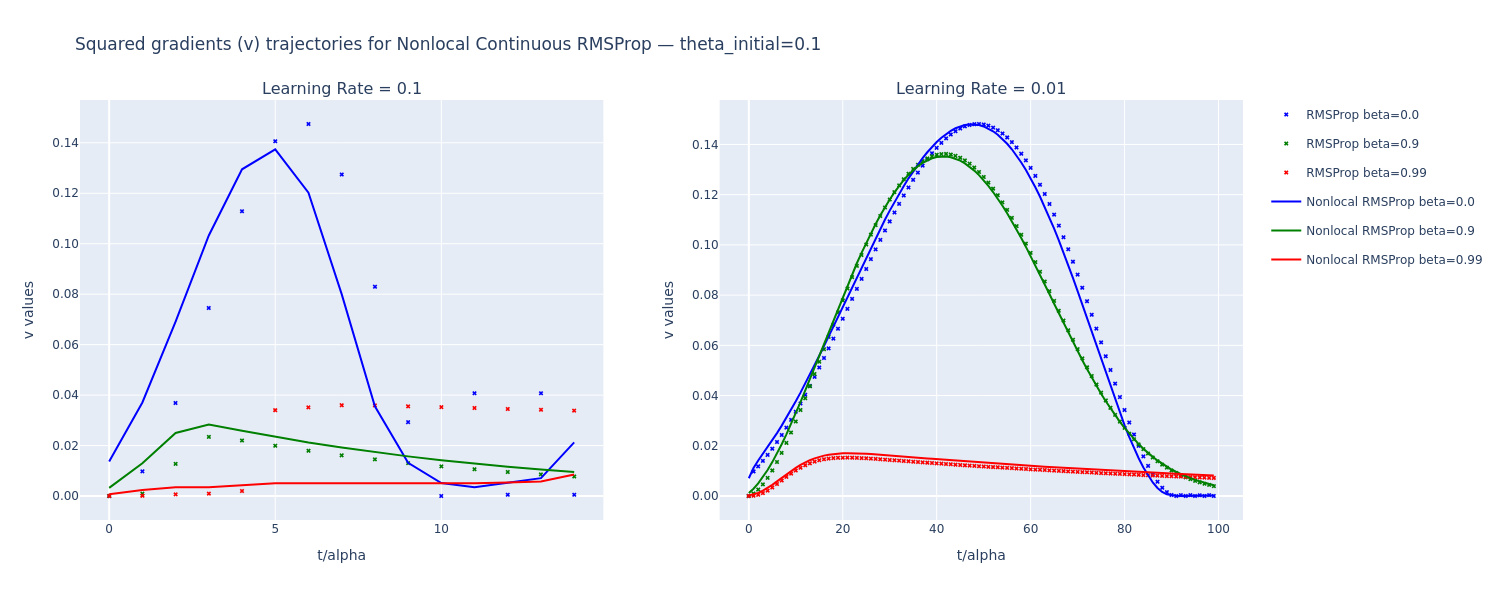}
    \caption{\textbf{Squared-gradient trajectories of \(v(t)\) for the nonlocal and discrete momentum models with positive initialization.} Squared-gradient trajectories \(v(t)\) with \(\theta_0=0.1\). For \(\alpha=0.1\), nonlocal curves are smoother than the discrete ones; \(\beta=0,0.9\) display a transient peak then decay, while \(\beta=0.99\) stays small and nearly flat. For \(\alpha=0.01\), discrete and nonlocal curves almost overlap: \(\beta\in\{0,0.9\}\) show a bell-shaped peak (larger for \(\beta=0\)) followed by decay, and \(\beta=0.99\) remains low throughout.}
    \label{fig:rmsprop-nonlocal-v_ncx_p}
  \end{subfigure}

  \vspace{0.75em}

  \begin{subfigure}{\linewidth}
    \centering
    \includegraphics[width=0.9\linewidth]{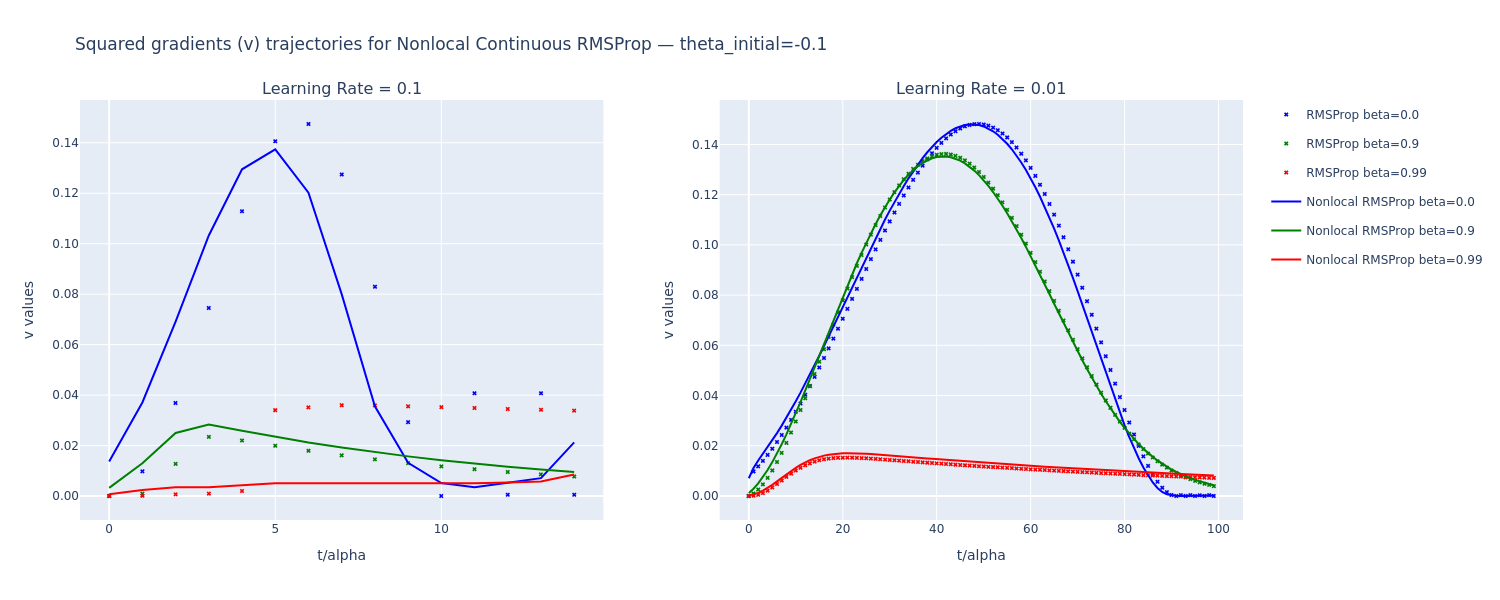}
    \caption{\textbf{Squared-gradient trajectories of \(v(t)\) for the nonlocal and discrete momentum models with negative initialization.} Squared-gradient trajectories \(v(t)\) with \(\theta_0=-0.1\). Same qualitative pattern as for \(\theta_0=0.1\): nonlocal curves are smoother at \(\alpha=0.1\); with \(\alpha=0.01\) both models align closely, with the largest peaks for \(\beta=0\), smaller for \(\beta=0.9\), and consistently low values for \(\beta=0.99\).}
    \label{fig:rmsprop-nonlocal-v_ncx_n}
  \end{subfigure}

  \caption{\textbf{Squared-gradient trajectories of \(v(t)\) for the nonlocal and discrete RMSProp models.} These two plots show the evolution of the trajectory of the gradient \(v(t)\) for the nonconvex function \(\tfrac{1}{4}(\theta^2-1)^2\) with two initial conditions $\theta_0=\pm0.1$ for RMSProp.}
  \label{fig:rmsprop-nonlocal-v_ncx}
\end{figure}

\paragraph{Nonlocal RMSProp:} Figures~\ref{fig:rmsprop-nonlocal-t_ncx_p}, \ref{fig:rmsprop-nonlocal-t_ncx_n}, \ref{fig:rmsprop-nonlocal-v_ncx_p}, and \ref{fig:rmsprop-nonlocal-v_ncx_n} compare the first–order nonlocal continuous RMSProp model with its discrete counterpart in the nonconvex case, for two initializations \(\theta_0\in\{-0.1,\,0.1\}\).

Figures~\ref{fig:rmsprop-nonlocal-t_ncx_p} and \ref{fig:rmsprop-nonlocal-t_ncx_n} show monotone convergence to the stable minima \(\theta^*\approx -1\) and \(\theta^*\approx 1\), respectively. With the larger learning rate \(\alpha=0.1\), high momentum (\(\beta=0.99\)) produces pronounced overshoots and oscillations around the target, whereas \(\beta=0\) and \(\beta=0.9\) yield smoother trajectories that stabilize more quickly. Reducing the learning rate to \(\alpha=0.01\) eliminates oscillations altogether: all settings converge smoothly, with \(\beta=0.99\) converging the fastest, followed by \(\beta=0.9\), and finally \(\beta=0\). The dynamics are symmetric with respect to the sign of the initial condition, as we discussed at the beginning of this section.

On the other hand, figures~\ref{fig:rmsprop-nonlocal-v_ncx_p} and \ref{fig:rmsprop-nonlocal-v_ncx_p} report the trajectories of \(v(t)\). For \(\alpha=0.1\), the nonlocal model produces smoother curves than the discrete scheme, particularly for \(\beta=0\) and \(\beta=0.99\). When \(\beta=0\) or \(0.9\), \(v(t)\) exhibits a transient rise followed by decay, whereas for \(\beta=0.99\) the values remain small and relatively flat—a behavior that does not occur in the discrete case, where at \(t/\alpha \approx 5\) the curve rises noticeably. With \(\alpha=0.01\), discrete and nonlocal curves nearly coincide for all \(\beta\): for \(\beta\in\{0,0.9\}\), they exhibit the characteristic bell-shaped peak (largest for \(\beta=0\)) before decaying to zero, while \(\beta=0.99\) keeps \(v(t)\) uniformly low with a slow decrease. As in the case of \(\theta(t)\), the behavior is mirrored between \(\theta_0=-0.1\) and \(\theta_0=0.1\).

Once again, let us read the plots in light of Lemma~\ref{lem:nonconvex_L2_gradient_memory}, Theorem~\ref{thm:nonconvex_PL_KL_adagrad_rmsprop}, and Theorem~\ref{thm:nonconvex_extended_memory_adagrad_rmsprop}. We restrict the discussion to statements that follow directly from these results for RMSProp, that is, the forgetting case $\xi=\lambda$. 

In the $v$-figures (Figs.~\ref{fig:rmsprop-nonlocal-v_ncx_p}--\ref{fig:rmsprop-nonlocal-v_ncx_n}), the memory variable $v(t)$ decays towards zero, in agreement with Theorem~\ref{thm:nonconvex_extended_memory_adagrad_rmsprop}. This decay is a consequence of the forgetting term $-\lambda v(t)$ in the RMSProp memory equation. Moreover, the relative speed observed for different values of $\beta$ is consistent with the dependence on 
\[
    \lambda=\frac{1-\beta}{\alpha}.
\]
As $\beta$ increases, $\lambda$ becomes smaller, the forgetting mechanism is weaker, and the decay of $v(t)$ becomes slower.

In the $\theta$-panels (Figs.~\ref{fig:rmsprop-nonlocal-t_ncx_p}--\ref{fig:rmsprop-nonlocal-t_ncx_n}), the trajectories remain in the compact sublevel set $\mathcal S_0$ and converge towards limiting values. Lemma~\ref{lem:nonconvex_L2_gradient_memory} provides the boundedness of the trajectory and of the memory, as well as the $L^2$-integrability of the gradient. Together with the subsequent vanishing-gradient argument, this yields 
\[
    \|\nabla f(\theta(t))\|\to0,
\]
so that accumulation points belong to the critical set. In the present one-dimensional toy model, the initial conditions lie in the attraction basins of the stable minima $\theta=\pm1$, which explains the observed convergence. This behavior is also consistent with the PL/KL conclusions of Theorem~\ref{thm:nonconvex_PL_KL_adagrad_rmsprop}: under PL one obtains exponential decay of the objective gap, while under KL the local rate is determined by the exponent $\sigma$.

\begin{figure}[htbp]
  \centering
  \begin{subfigure}{\linewidth}
    \centering
    \includegraphics[width=0.9\linewidth]{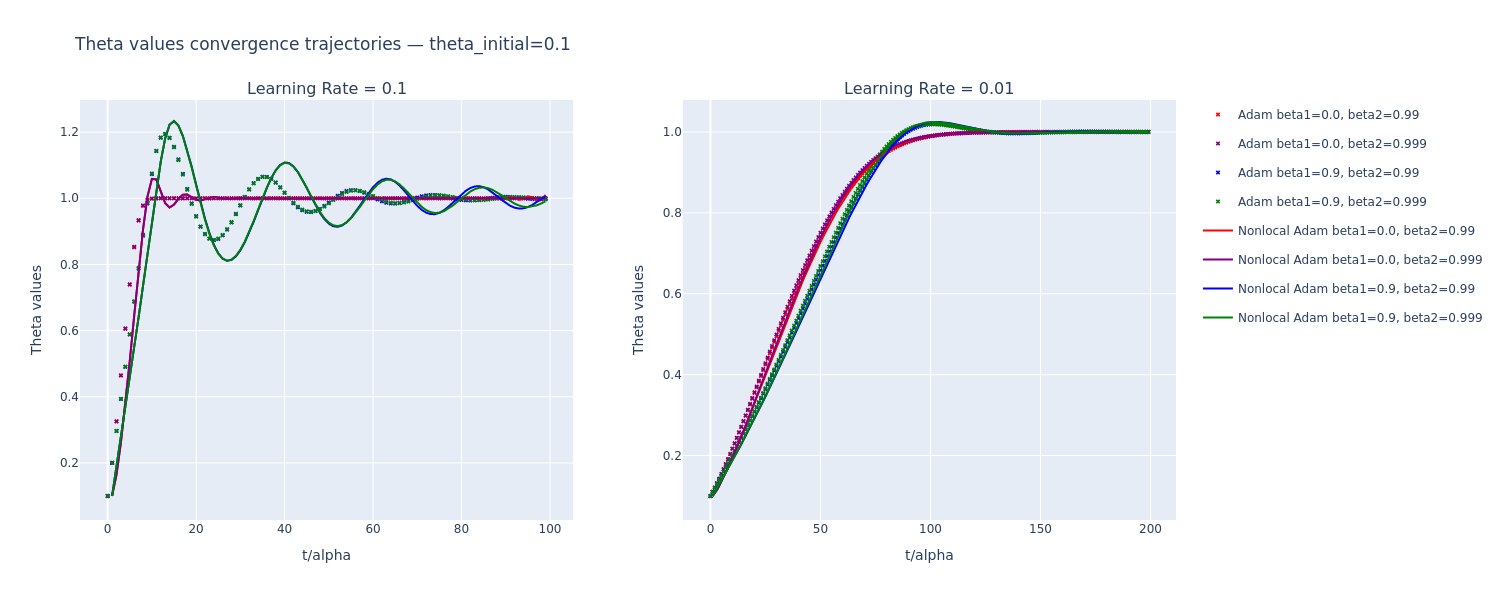}
    \caption{\textbf{Convergence trajectories of \(\theta(t)\) for the nonlocal and discrete Adam models with positive initialization.} Convergence of \(\theta(t)\) with \(\theta_0=0.1\). Solid lines: first-order nonlocal continuous Adam; markers: discrete Adam. Two learning rates (\(\alpha = 0.1,\, 0.01\)) and parameters \(\beta_1 \in \{0, 0.9\}\), \(\beta_2 \in \{0.99, 0.999\}\) are considered. With \(\alpha = 0.1\), the \(\beta_1 = 0.9\) configurations exhibit overshooting and lightly damped oscillations, whereas with \(\alpha = 0.01\) all curves converge smoothly and nearly overlap, approaching \(\theta^* \approx 1\).}
    \label{fig:adam-nonlocal-t_ncx_p}
  \end{subfigure}

  \vspace{0.75em}

  \begin{subfigure}{\linewidth}
    \centering
    \includegraphics[width=0.9\linewidth]{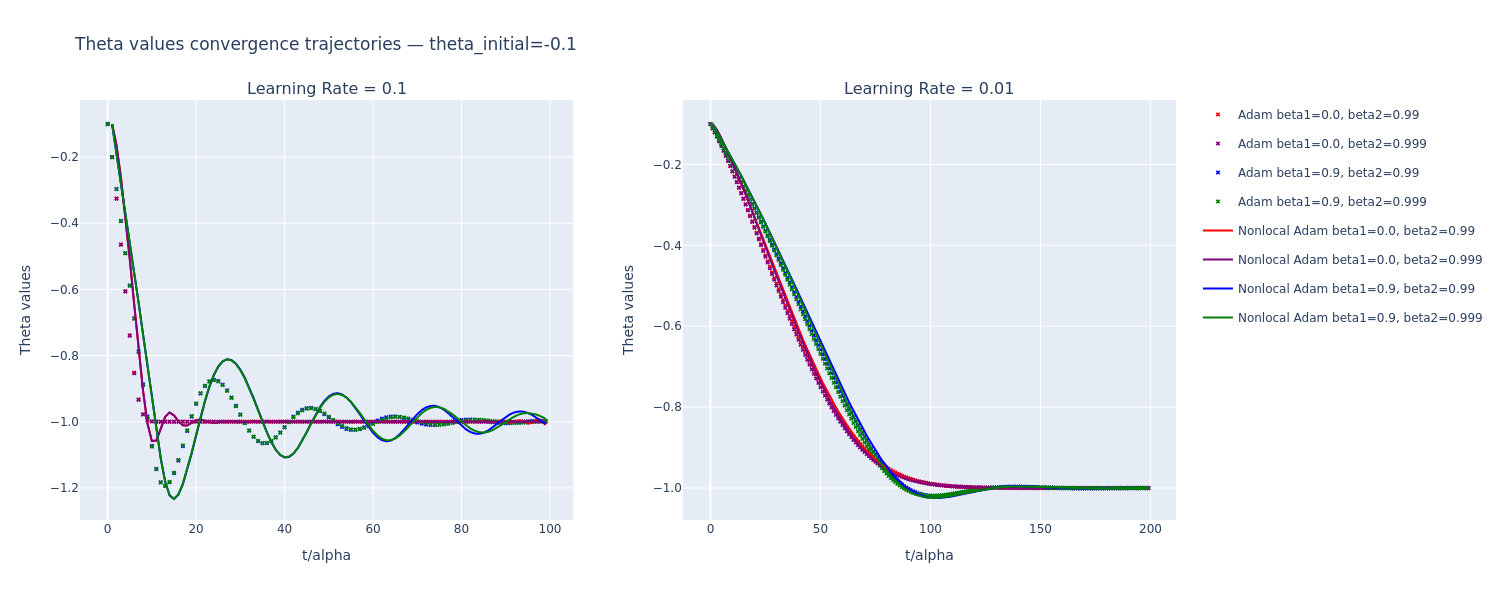}
    \caption{\textbf{Convergence trajectories of \(\theta(t)\) for the nonlocal and discrete Adam models with negative initialization.} Convergence of \(\theta(t)\) with \(\theta_0=-0.1\). Same setup as Fig.~\ref{fig:adam-nonlocal-t_ncx_p}. For \(\alpha=0.1\), \(\beta_1=0.9\) produces mild oscillations; for \(\alpha=0.01\) the trajectories are smooth and almost identical between the nonlocal and discrete models, converging to \(\theta^*\!\approx -1\).}
    \label{fig:adam-nonlocal-t_ncx_n}
  \end{subfigure}

  \caption{\textbf{Convergence trajectories of \(\theta(t)\) for the nonlocal and discrete Adam models.} These two plots show the evolution of the trajectory of \(\theta(t)\) for the nonconvex function \(\tfrac{1}{4}(\theta^2-1)^2\) with two initial conditions $\theta_0=\pm0.1$ for Adam.}
  \label{fig:rmsprop-nonlocal-t_ncx}
\end{figure}

\begin{figure}[htbp]
  \centering
  \begin{subfigure}{\linewidth}
    \centering
    \includegraphics[width=0.9\linewidth]{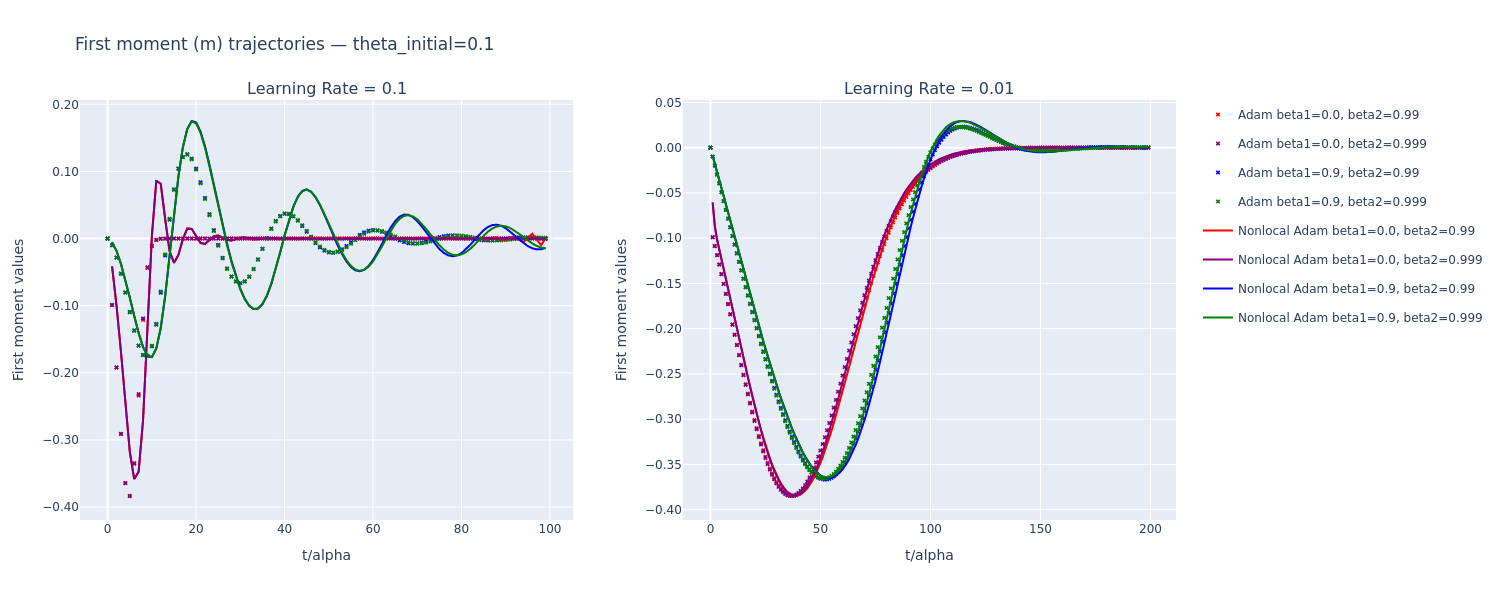}
    \caption{\textbf{First-moment trajectories of \(m(t)\) for the nonlocal and discrete Adam models with positive initialization.} First-moment trajectories \(m(t)\) with \(\theta_0=0.1\). At the larger learning rate \(\alpha=0.1\), \(m(t)\) exhibits damped oscillations around zero—most pronounced for \(\beta_1=0.9\) and \(\beta_2=0.999\). With \(\alpha=0.01\), all curves show a single hump followed by decay to zero. Nonlocal and discrete results closely match across regimes.}
    \label{fig:adam-nonlocal-m_ncx_p}
  \end{subfigure}

  \vspace{0.75em}

  \begin{subfigure}{\linewidth}
    \centering
    \includegraphics[width=0.9\linewidth]{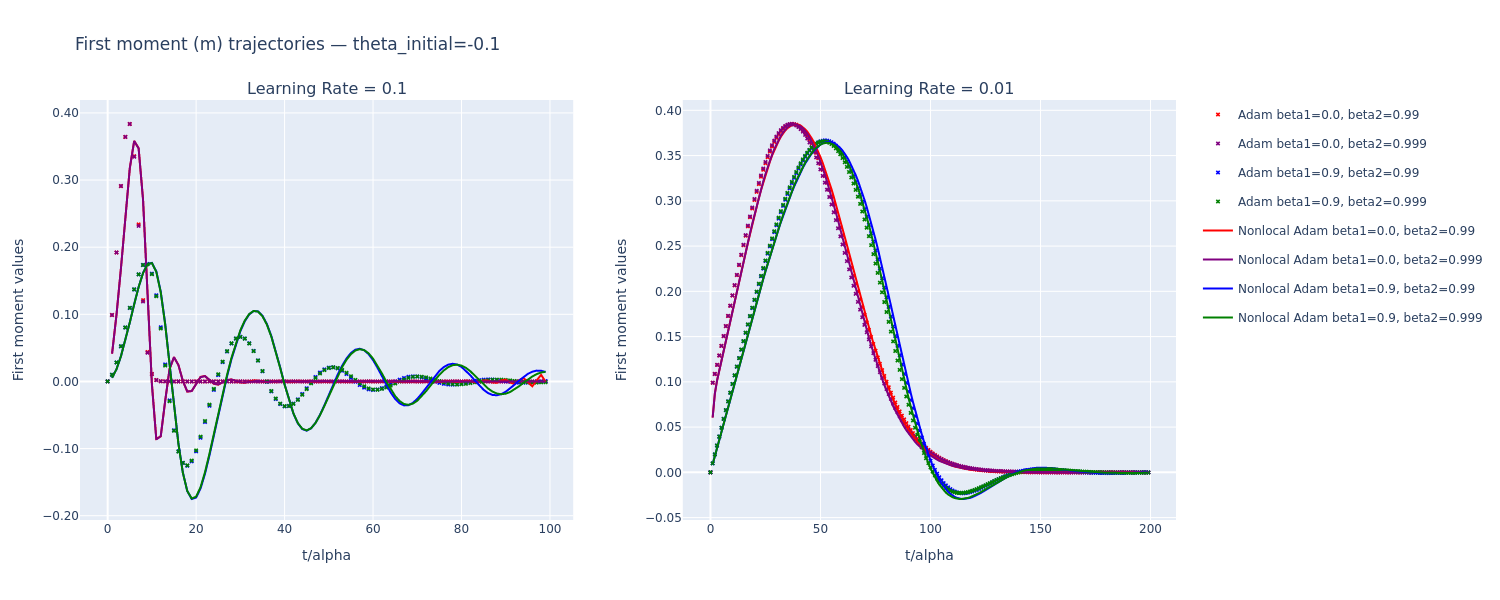}
    \caption{\textbf{First-moment trajectories of \(m(t)\) for the nonlocal and discrete Adam models with negative initialization.} First-moment trajectories \(m(t)\) with \(\theta_0=-0.1\). Mirror behavior of the positive case. For \(\alpha=0.1\) the oscillations are stronger for \(\beta_1=0.9\); for \(\alpha=0.01\) the profiles are smooth and the nonlocal curves overlap the discrete ones throughout.}
    \label{fig:adam-nonlocal-m_ncx_n}
  \end{subfigure}

  \caption{\textbf{Convergence trajectories of \(m(t)\) for the nonlocal and discrete Adam models.} These two plots show the evolution of the trajectory of \(m(t)\) for the nonconvex function \(\tfrac{1}{4}(\theta^2-1)^2\) with two initial conditions $\theta_0=\pm0.1$ for Adam. }
  \label{fig:adam-nonlocal-m_ncx}
\end{figure}

\begin{figure}[htbp]
  \centering
  \begin{subfigure}{\linewidth}
    \centering
    \includegraphics[width=0.9\linewidth]{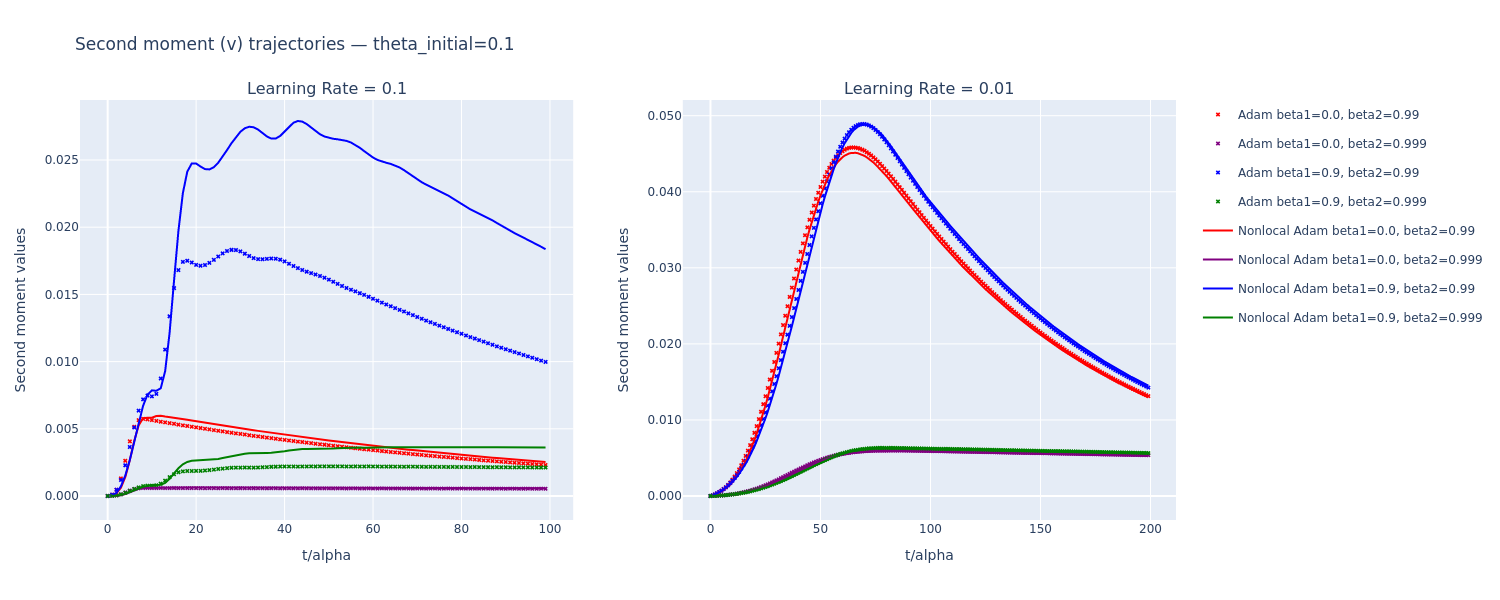}
    \caption{\textbf{Second-moment trajectories of \(v(t)\) for the nonlocal and discrete Adam models with positive initialization.} Second-moment trajectories \(v(t)\) with \(\theta_0=0.1\). Peak height and decay depend on the parameters: with \(\alpha=0.1\), \(\beta_1=0.9,\beta_2=0.99\) yields the largest transient peaks, whereas \(\beta_2=0.999\) produces smaller, flatter curves. For \(\alpha=0.01\), all settings display the characteristic bell-shaped peak followed by slow decay, and nonlocal and discrete curves nearly coincide.}
    \label{fig:adam-nonlocal-v_ncx_p}
  \end{subfigure}

  \vspace{0.75em}

  \begin{subfigure}{\linewidth}
    \centering
    \includegraphics[width=0.9\linewidth]{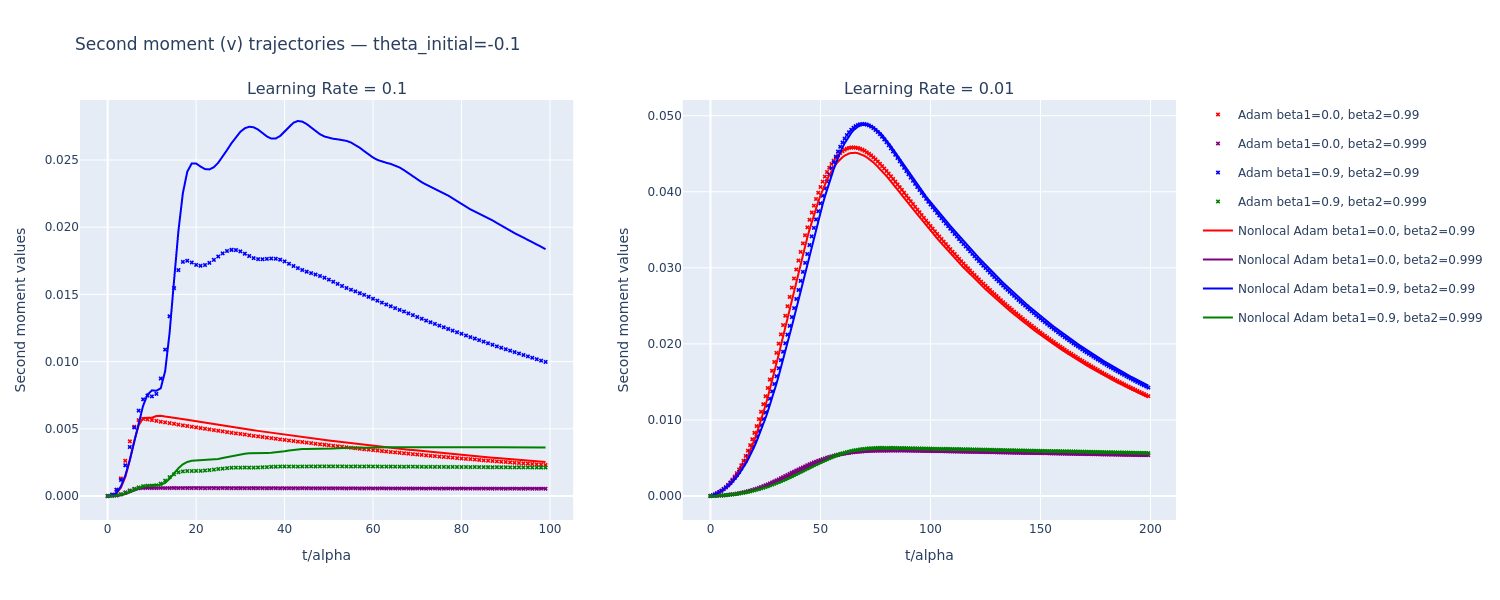}
    \caption{\textbf{Second-moment trajectories of \(v(t)\) for the nonlocal and discrete Adam models with negative initialization.} Second-moment trajectories \(v(t)\) with \(\theta_0=-0.1\). Same qualitative pattern as in Fig.~\ref{fig:adam-nonlocal-v_ncx_p}: larger peaks for \(\beta_2=0.99\) and much lower values for \(\beta_2=0.999\). Agreement between nonlocal and discrete models is very close, especially for the smaller learning rate.}
    \label{fig:adam-nonlocal-v_ncx_n}
  \end{subfigure}

  \caption{\textbf{Convergence trajectories of \(v(t)\) for the nonlocal and discrete Adam models.} These two plots show the evolution of the trajectory of \(v(t)\) for the nonconvex function \(\tfrac{1}{4}(\theta^2-1)^2\) with two initial conditions $\theta_0=\pm0.1$ for Adam. }
  \label{fig:adam-nonlocal-v_ncx}
\end{figure}

\paragraph{Nonlocal Adam:} Figures~\ref{fig:adam-nonlocal-t_ncx_p}, \ref{fig:adam-nonlocal-t_ncx_n}, \ref{fig:adam-nonlocal-m_ncx_p}, and \ref{fig:adam-nonlocal-m_ncx_n}, as well as \ref{fig:adam-nonlocal-v_ncx_p} and \ref{fig:adam-nonlocal-v_ncx_n}, compare the first–order nonlocal continuous Adam model with its discrete counterpart for both signs of the initialization \(\theta_0=\pm 0.1\) in the nonconvex setting. 

Figures~\ref{fig:adam-nonlocal-t_ncx_p} and \ref{fig:adam-nonlocal-t_ncx_n} show a rapid approach to the stable minimizers (\(\theta^*\approx \pm 1\)). With the larger learning rate \(\alpha=0.1\), settings with \(\beta_1=0.9\) display overshoot and lightly damped oscillations around the limit, whereas \(\beta_1=0\) stabilizes more quickly. The nonlocal trajectories follow the same trends (reduced oscillation amplitude) as the discrete ones. Lowering the learning rate to \(\alpha=0.01\) removes instability: all parameter choices converge smoothly and almost perfectly overlap between the nonlocal and discrete models. The dynamics are symmetric with respect to the sign of \(\theta_0\), as discussed earlier.

In Figures~\ref{fig:adam-nonlocal-m_ncx_p} and \ref{fig:adam-nonlocal-m_ncx_n}, \(m(t)\) exhibits damped oscillations around zero for \(\alpha=0.1\), with the most pronounced swings occurring for \(\beta_1=0.9\) (especially with \(\beta_2=0.99\) and \(0.999\)). For \(\alpha=0.01\), all curves show a single valley followed by a gradual decay to zero; across the whole range the nonlocal and discrete trajectories are nearly indistinguishable. As expected, the profiles for positive and negative \(\theta_0\) are mirror images.

Figures~\ref{fig:adam-nonlocal-v_ncx_p} and \ref{fig:adam-nonlocal-v_ncx_n} show the evolution of the squared–gradient moving average $v(t)$. With \(\alpha=0.1\), the peak height and decay rate depend strongly on the parameters: \(\beta_1=0.9,\beta_2=0.99\) yields the largest transient peaks, while \(\beta_2=0.999\) (for either \(\beta_1\)) produces much smaller, flatter curves due to heavier averaging. For \(\alpha=0.01\), both models present the characteristic bell–shaped peak (largest for \(\beta_2=0.99\)) followed by a slow decay, and the nonlocal curves practically coincide with the discrete ones.

To conclude, let us analyze these figures via Theorem~\ref{thm:nonconvex_adam_shifted}. In the nonconvex Adam analysis, the Lyapunov functional is written in terms of the shifted quantities
\[
    \widehat m(t):=m(t+\alpha),
    \qquad
    \widehat v(t):=v(t+\alpha),
\]
and of the shifted Adam gain $\widehat a(t)$. Under the scalar Lyapunov condition and the one-sided shift-defect estimate, Theorem~\ref{thm:nonconvex_adam_shifted} yields
\[
    \dot V(t)
    \leq
    -\widehat a(t)(c_\beta-\sigma_\alpha)\|\widehat m(t)\|^2
    \leq 0.
\]
Consequently, the shifted first moment satisfies $\widehat m(t)\to0$. This is consistent with the damping observed in the first-moment plots (Figs.~\ref{fig:adam-nonlocal-m_ncx_p} and~\ref{fig:adam-nonlocal-m_ncx_n}). Since the plots display $m(t)$ rather than $\widehat m(t)=m(t+\alpha)$, the agreement should be understood up to this small time shift.

The second-moment plots (Figs.~\ref{fig:adam-nonlocal-v_ncx_p} and~\ref{fig:adam-nonlocal-v_ncx_n}) show an initial transient peak followed by decay. This is also consistent with the theorem: once $\widehat m(t)\to0$, the shifted first-moment equation implies $\nabla f(\theta(t))\to0$, and the shifted second-moment equation then gives $\widehat v(t)\to0$. Again, the plotted quantity is $v(t)$, so the statement is to be interpreted modulo the shift by $\alpha$.

Finally, in the $\theta$-trajectories (Figs.~\ref{fig:adam-nonlocal-t_ncx_p} and~\ref{fig:adam-nonlocal-t_ncx_n}), the solutions stabilize at limiting values inside the compact set $K_0$. The general conclusion certified by Theorem~\ref{thm:nonconvex_adam_shifted} is asymptotic stationarity,
\[
    \|\nabla f(\theta(t))\|\to0,
\]
and therefore every accumulation point belongs to $\operatorname{Crit}(f)\cap K_0$. In the present one-dimensional toy model, the initial conditions $\theta_0=\pm0.1$ lie in the attraction basins of the stable minima $\theta=\pm1$, which explains the observed convergence to those values.

\section{Conclusion}

In this paper, we have introduced a continuous-time formulation for the AdaGrad, RMSProp, and Adam optimization algorithms by modeling them as first-order integro-differential equations. This novel approach captures the cumulative and memory effects inherent in these adaptive methods, providing a unified and precise mathematical framework that bridges the gap between discrete algorithms and continuous dynamical systems.

Our continuous models establish a direct connection between the discrete implementations of these algorithms and their continuous counterparts. The nonlocal integral terms naturally encapsulate the algorithms’ memory and adaptive behaviors, allowing the explicit representation of the influence of past gradients on current updates, Propositions~\ref{prop:AdaGrad} to \ref{prop:Adam}. 

Moreover, we study the convergence and stability properties of these continuous models. For AdaGrad and RMSProp in the strongly convex setting, Theorem~\ref{thm:adagrad_rmsprop_memory} provides a preliminary a priori convergence estimate. This estimate is robust and applies to both memory kernels, but it should not be interpreted as the sharp final rate of the methods. More precisely, it yields a subexponential bound of the form
\[
    \|\theta(t)-\theta^\ast\|\leq C e^{-c\sqrt t},
\]
which is sufficient to control the accumulated memory term. Once this control is obtained, the effective factor
\[
    \mu(t)
    =
    \frac{1}{\sqrt{M_\nu[g^2](t+\alpha)}+\varepsilon}
\]
is bounded from below by a positive constant, and sharper stability conclusions follow from the extended memory functional. To separate accumulation from forgetting, Theorem~\ref{thm:extended_memory_adagrad_rmsprop} introduces the functional
\[
    E_\rho(t)=\Phi(t)+\rho\,M_\nu[g^2](t),
\]
where $\Phi(t)=f(\theta(t))-f(\theta^\ast)$. The memory equation has the form
\[
    \dot M_\nu[g^2](t)
    =
    \lambda\|g(t)\|^2-\xi M_\nu[g^2](t),
    \qquad
    \xi\in\{0,\lambda\}.
\]
Thus, $\lambda\|g(t)\|^2$ represents accumulation, while $-\xi M_\nu[g^2](t)$ represents forgetting. With the choice $\rho=\mu_0/(2\lambda)$, one obtains
\[
    \dot E_\rho(t)
    \leq
    -m\mu_0\Phi(t)-\rho\xi M_\nu[g^2](t).
\]
In the RMSProp case, $\xi=\lambda>0$, this estimate closes as
\[
    \dot E_\rho(t)\leq -\kappa_0E_\rho(t),
    \qquad
    \kappa_0=\min\{m\mu_0,\xi\}>0.
\]
Consequently, $E_\rho(t)$, the objective gap $\Phi(t)$, and the memory variable decay exponentially. In contrast, AdaGrad corresponds to the cumulative case $\xi=0$. Then $E_\rho$ remains nonincreasing, but the estimate cannot be closed as an exponential decay for $E_\rho$, because there is no forgetting term. Nevertheless, the distance to the minimizer still decays exponentially once the uniform lower bound $\mu(t)\geq \mu_0>0$ has been established, while the accumulated memory converges to a finite, generally nonzero, limit.

For Adam, Theorem~\ref{thm:adam_shifted} establishes a conditional Lyapunov stability result for the continuous Adam dynamics. In this case the relevant quantities are the shifted first moment $\widehat m(t):=m(t+\alpha)$, the shifted second moment $\widehat v(t):=v(t+\alpha)$,
and the effective shifted Adam gain $\widehat a(t)$. The Lyapunov functional is
\[
    V(t)
    =
    \Phi(t)
    +
    \frac{\widehat a(t)}{2\lambda_1}
    \|\widehat m(t)\|^2.
\]
The dissipation estimate closes under the scalar Lyapunov condition controlling the logarithmic growth of $\widehat a(t)$, together with the one-sided shift-defect estimate. Under these assumptions,
\[
    \dot V(t)
    \leq
    -\widehat a(t)(c_\beta-\sigma_\alpha)\|\widehat m(t)\|^2
    \leq 0.
\]
Hence the shifted first moment vanishes asymptotically, $\widehat m(t)\to0$, and the strongly convexity of the objective yields
\[
    \nabla f(\theta(t))\to0,
    \qquad
    \theta(t)\to\theta^\ast .
\]
The second moment does not need to appear as a separate term in the Lyapunov functional, since its effect is encoded in the effective gain $\widehat a(t)$.

The nonconvex analysis follows the same memory-based viewpoint, but the conclusions must be stated more carefully. Lemma~\ref{lem:nonconvex_L2_gradient_memory} shows that, for AdaGrad and RMSProp, the trajectory remains in the compact sublevel set $\mathcal S_0$, the memory remains bounded, and the gradient is square-integrable in time. Together with the subsequent vanishing-gradient argument, this gives
\[
    \|\nabla f(\theta(t))\|\to0.
\]
Thus, every accumulation point of the trajectory belongs to the critical set $\operatorname{Crit}(f)\cap\mathcal S_0$. This is an asymptotic stationarity result; by itself, it does not imply convergence to a prescribed minimizer.

Building on this compactness and stationarity structure, Theorem~\ref{thm:nonconvex_PL_KL_adagrad_rmsprop} gives sharper conclusions under additional geometric assumptions. Under the Polyak--{\L}ojasiewicz (PL) condition on $\mathcal S_0$, the objective gap decays exponentially and the trajectory has finite length, hence it converges to a minimizer in $\mathcal S_0$. Under a uniform Kurdyka--{\L}ojasiewicz (KL) inequality near the limiting critical set, the trajectory converges to a single critical point, and the rate is dictated by the KL exponent $\sigma$: finite-time convergence when $0\leq\sigma<1/2$, exponential convergence when $\sigma=1/2$, and polynomial decay of order
\[
    O\!\left(t^{-\frac{1}{2\sigma-1}}\right)
\]
when $1/2<\sigma<1$.

Theorem~\ref{thm:nonconvex_extended_memory_adagrad_rmsprop} extends the functional $E_\rho$ to the nonconvex AdaGrad/RMSProp setting. In the RMSProp case, the forgetting term implies that the memory variable decays to zero, and under PL the extended functional satisfies a closed exponential estimate. In the AdaGrad case, the functional is still nonincreasing, but the accumulated memory is monotone and converges to a finite, generally nonzero, limit; therefore one should not claim exponential decay of $E_\rho$ in the cumulative case.

Finally, Theorem~\ref{thm:nonconvex_adam_shifted} treats the nonconvex Adam flow using the Lyapunov functional built from $\widehat m(t)$, $\widehat v(t)$, and the effective gain $\widehat a(t)$. Under the scalar Lyapunov condition and the one-sided shift-defect estimate, the theorem yields asymptotic stationarity,
\[
    \|\nabla f(\theta(t))\|\to0,
\]
and convergence toward the critical set inside the relevant compact sublevel set. Convergence to a single critical point requires an additional structural assumption, such as finiteness of the critical set in the compact region, while convergence to a minimizer requires a PL-type condition or an analogous minimizing structure. Thus, in the nonconvex setting, the general conclusion is not unconditional convergence to a prescribed minimizer, but rather asymptotic stationarity, with stronger convergence and rates obtained under PL or KL assumptions.

Through extensive numerical simulations, we have demonstrated that these continuous formulations accurately reproduce the convergence dynamics and behaviors of the original algorithms across a wide range of settings, including different learning rates and parameter configurations. Furthermore, as the learning rate decreases, the similarity between the continuous and discrete models increases, confirming the effectiveness of the continuous representation in capturing the behavior of the discrete algorithms. The simulations also align closely with the theoretical predictions: in AdaGrad, trajectories approach minimizers in the convex case (critical points in the nonconvex case) while the accumulated memory saturates and the effective step shrinks, flattening the curves; in RMSProp, the memory decay exponentially, with slower rates as the memory becomes stronger ($\beta \to 1$); and in Adam, both moments damp and the parameters stabilize, with longer transients under stronger memory.

However, while our continuous-time models provide valuable insights, they are based on deterministic settings and simplified objective functions. Extending these models to handle stochasticity, high-dimensional parameter spaces, and the complex architectures of modern neural networks is a natural next step. Furthermore, addressing the computational complexity associated with solving integro-differential equations is essential for enhancing their practical applicability, highlighting the need for efficient computational strategies.

In conclusion, our findings suggest that the integro-differential approach not only provides a valid method for representing the dynamics of AdaGrad, RMSProp, and Adam in a continuous-time framework but also serves as a powerful tool for advancing the theoretical understanding of these widely used optimization algorithms. By bridging discrete and continuous optimization methods, this approach has the potential to inspire novel methodologies and applications in AI research that involve memory (nonlocal) effects. Such applications include nonlocal Lagrangian and Hamiltonian formalisms, as well as extensions of Noether’s theorem to nonlocal Lagrangians, which may be further explored in future work \cite{heredia2025nonlocalmechanics}. In this direction, the second-order extension of the present framework for Adam, together with the associated Lagrangian-like formulation proposed for this optimizer, provides a natural continuation of the ideas developed here \cite{heredia2026adamadamlikelagrangianssecondorder}.

\section*{Acknowledgments}
I would like to express sincere gratitude to Josep Llosa (University of Barcelona), Rhys Gould (University of Cambridge), Hidenori Tanaka (Harvard University), and Javier Cristin (Autonomous University of Barcelona) for their insightful discussions on nonlocality and valuable feedback on this manuscript. I also thank the Digital Transformation Team and the Nennisiwok Team at DAMM for their ongoing support. Finally, we would like to thank the anonymous referee for their comments and observations on the theorems, which helped us improve them.

\section*{Author Contributions}
Carlos Heredia was solely responsible for the conception, development, implementation, and writing of this manuscript.

\section*{Funding}
The author declares that no funding was received for this work.

\appendix

\section*{Appendix}

\section{Nonconvex Simulation. Case $\theta^*=0$.}\label{Appendix}
This section illustrates the simulations for the \(\theta^*=0\) case of the nonconvex function \(\tfrac{1}{4}(\theta^2-1)^2\). Note that this case is trivial: since the initial condition is \(\theta_0 = 0\), the system starts at the critical (maximum) point and therefore remains there, as shown in Fig.~\ref{fig:adagrad-nonlocal-ncx_0}. As can be observed in the non-convex plots, any small perturbation causes the system to converge to one of the local minima.

On the other hand, since all the plots for AdaGrad, RMSProp, and Adam are identical, with all trajectories remaining at the unstable critical point \(\theta_0=0\), we only display the first one: AdaGrad. The others can be found in the \texttt{simulations/figures} folder of the GitHub repository.

\section{Simulation Figures}\label{App:SF}

In this section, we present the simulation figures for both the convex and nonconvex cases.

\subsection{Convex Cases}


\FloatBarrier
\subsection{Nonconvex Cases}

\begin{figure}[htbp]
  \centering
  \begin{subfigure}{\linewidth}
    \centering
    \includegraphics[width=0.9\linewidth]{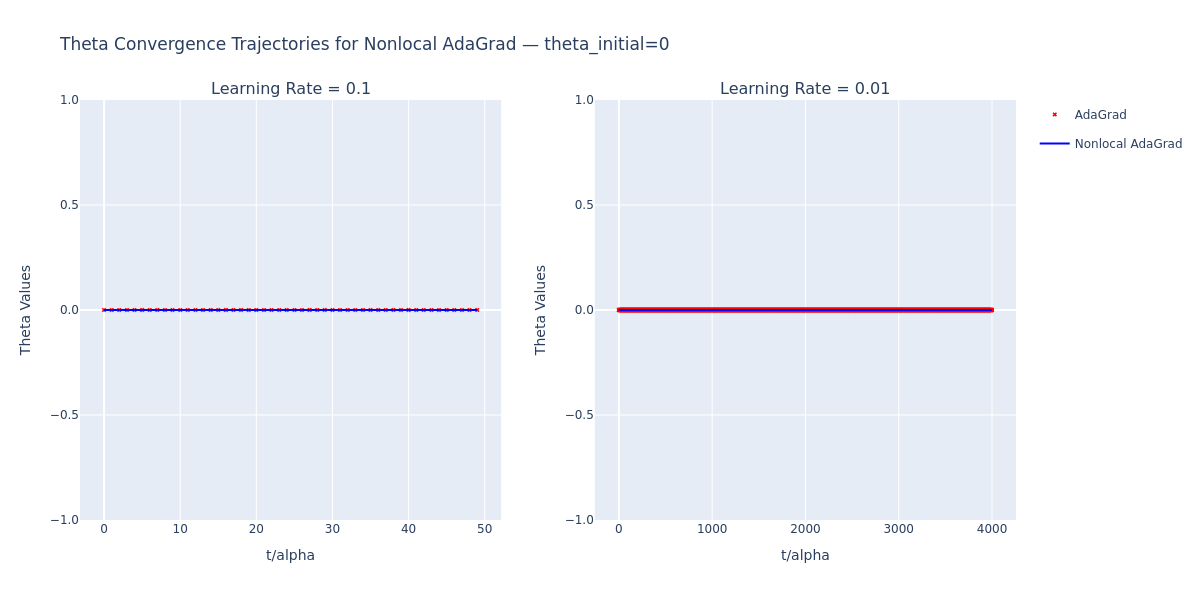}
    \caption{\textbf{Parameter trajectories of \(\theta(t)\) for the nonlocal and discrete AdaGrad models with zero initialization.} Theta trajectories (AdaGrad vs. Non-local AdaGrad) for the nonconvex function $\frac{1}{4}(\theta^2-1)^2$ for the initial value $\theta_0=0$.}
    \label{fig:ncx_theta_0}
  \end{subfigure}

  \vspace{0.75em}

  \begin{subfigure}{\linewidth}
    \centering
    \includegraphics[width=0.9\linewidth]{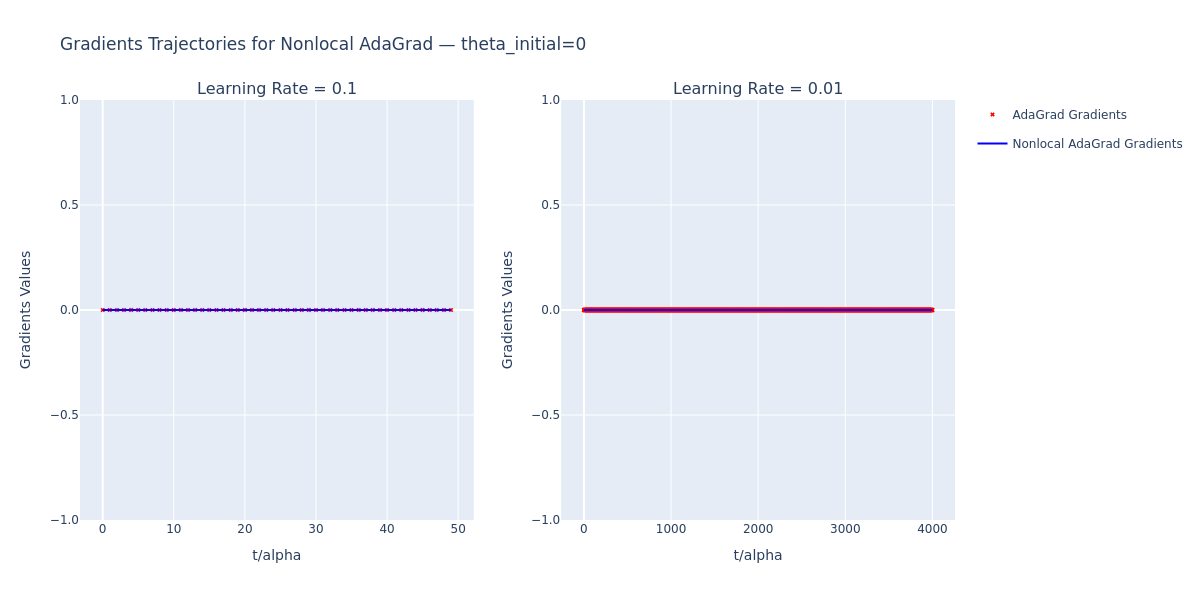}
    \caption{\textbf{Gradient trajectories of \(\theta(t)\) for the nonlocal and discrete AdaGrad models with zero initialization.} Gradient trajectories (AdaGrad vs. Non-local AdaGrad) for the nonconvex function $\frac{1}{4}(\theta^2-1)^2$ for the initial value $\theta_0=0$.}
    \label{fig:theta}
  \end{subfigure}

  \caption{\textbf{Equilibrium trajectories of \(\theta(t)\) and \(G(t)\) for the nonlocal and discrete AdaGrad models.} These two plots show the evolution of the trajectory of \(\theta(t)\) with the gradient \(G(t)\) for the nonconvex function \(\tfrac{1}{4}(\theta^2-1)^2\) under AdaGrad. As can be observed, since the system starts directly at the unstable local maximum \(\theta_0 = 0\), it remains unchanged. However, even an arbitrarily small perturbation would cause it to converge to one of the local minima. We also present the discrete case, and, as can be seen, the behavior of the continuous system is completely identical.}
  \label{fig:adagrad-nonlocal-ncx_0}
\end{figure}

\bibliography{iclr_conference}
\bibliographystyle{iclr_conference}

\end{document}